\documentclass[3p]{elsarticle}

\usepackage[hidelinks]{hyperref}
\usepackage{lineno}

\journal{International Journal of Approximate Reasoning}

\usepackage{import}
\usepackage{amsmath, amssymb, amsthm}
\usepackage{graphicx}
\usepackage{color}
\usepackage{algpseudocode}
\usepackage{algorithm}
\usepackage{subfigure}
\usepackage{pgf}
\usepackage[textsize=tiny,backgroundcolor=red!30]{todonotes}

\algnewcommand{\LineComment}[1]{\State \(\triangleright\) #1}
 
\newcommand{\revision}[1]{#1}  

\newtheorem{result}{Result}

\begin{document}

\begin{frontmatter}

\title{Degenerate Gaussian factors for probabilistic inference}

\date{October 12, 2021}

\author{J.C. Schoeman\corref{cor1}}
\ead{jcschoeman@sun.ac.za}
\cortext[cor1]{Corresponding author}
\author{C.E. van Daalen\corref{cor2}}
\ead{cvdaalen@sun.ac.za}
\author{J.A. du Preez\corref{cor2}}
\ead{dupreez@sun.ac.za}
\address{Department of Electrical and Electronic Engineering, Stellenbosch University, South Africa}

\begin{abstract}
	In this paper, we propose a parametrised factor that enables inference on Gaussian networks where linear dependencies exist among the random variables. Our factor representation is \revision{effectively} a generalisation of traditional Gaussian parametrisations where the positive-definite constraint \revision{of the covariance matrix} has been relaxed. For this purpose, we derive various statistical operations and results (such as marginalisation, multiplication and affine transformations of random variables) that extend the capabilities of Gaussian factors to these \emph{degenerate} settings. By using this principled factor definition, degeneracies can be accommodated accurately and automatically at little additional computational cost. As illustration, we apply our methodology to a representative example involving recursive state estimation of cooperative mobile robots.
\end{abstract}

\begin{keyword}
Gaussian network\sep inference\sep linear dependence\sep degenerate factor\sep Dirac delta function\sep recursive state estimation
\end{keyword}

\end{frontmatter}


\section{Introduction}

	Using probability theory to solve problems in engineering and computer science often follows the familiar process of modelling, observation and inference. \revision{A good example of this is the Kalman filter used in recursive state estimation, where noisy measurements are used to estimate time-dependent latent states \cite{kalman1960new}.} When dealing with continuous random variables \revision{such as these,} a popular choice for keeping the inference process tractable is to make the Gaussian assumption \revision{for all prior and conditional distributions \cite{shachter1989gaussian}.} Under this assumption, the representation is closed under typical statistical operations required for inference algorithms. \revision{In other words, the result of each operation is conveniently once again a Gaussian distribution \cite{koller2009}.}
	
	An important (and necessary) constraint for using Gaussian models is that the covariance matrix of any \revision{Gaussian distribution} must be positive definite \cite{peebles1987probability}. If the covariance matrix were instead only positive \emph{semi}-definite, the likelihood \revision{of the distribution} would not be defined, nor would the precision matrix (i.e., the inverse of the covariance matrix) exist. The latter would be problematic even for basic operations such as multiplication and conditioning on evidence. Such \revision{positive semi-definite settings correspond to cases of perfect correlation and} are commonly referred to as \emph{degenerate}, where it is understood that linear dependencies exist among certain random variables \cite{lauritzen2001stable}. This implies that the density in reality only has support (i.e., is not equal to zero) on a lower-dimensional manifold.
	
	\revision{There are a wide variety of applications where the need for degenerate Gaussian factors is apparent. Two examples include the work by Pawula et al. \cite{pawula1982distribution}, where they determine formulas for the symbol error rate in digital frequency modulation, and that by Cao and Shen \cite{cao2021fault} on the fault detection of photovoltaic cells using Gaussian mixture models (GMMs). In both of these cases, deterministic formulas lead to degeneracies when combined with standard probabilistic models. Two further studies by Quinonero-Candela and Rasmussen \cite{quinonero2005unifying} and Biernacki and Chr{\'e}tien \cite{biernacki2003degeneracy} show that degenerate cases arise even in general regression and estimation techniques such as Gaussian processes (GPs) and expectation maximisation (EM).}
	
	Although certain applications \revision{necessitate} the explicit \revision{modelling} of linear dependencies between random variables, another common manifestation \revision{thereof} is the result of machine precision limitations \cite{fitzgerald1971}. A simple example of the latter is the repeated noisy observation of a constant latent state. Although the uncertainty in the posterior distribution will theoretically only be zero in the limit, this will happen after finite time in practice. Since such characteristics are typically a function of model parameters, it can easily lead to numerical errors in certain cases if not handled appropriately.
	
	A straightforward solution to guard against degeneracies is to simply redefine the problem \revision{to remove redundant variables} as needed. For example, if a vehicle is moving on a constrained horizontal plane, it is unnecessary to estimate its position in the entire three-dimensional space. Similarly, should two random variables become perfectly correlated due to machine precision effects, one could manually reduce the dimensionality of the space until \revision{sufficient uncertainty once again allows} explicit separation. Although this solution might work, it involves repeated human engineering, \revision{which can be especially tedious when dealing with time-dependent models where degeneracies arise inconsistently and unpredictably.} An alternative approach that can automatically handle degenerate cases is therefore desirable.
	
%
	
	\paragraph{Related literature} \revision{In the majority of existing approaches, degeneracies are correctly identified, but no subsequent mechanisms are developed which would enable inference. For example,} in their work on Gaussian influence diagrams, Shachter and Kenley \cite{shachter1989gaussian} recognise that zero variance corresponds to linear dependencies, but do not provide a means to handle such cases automatically. Lauritzen and Jensen \cite{lauritzen2001stable} also acknowledge possible degeneracies in their definition of conditional Gaussian distributions -- a hybrid parametrisation where the joint distribution of continuous random variables given discrete random variables is assumed to be multivariate Gaussian. However, since they employ singular covariance matrices for this purpose, representing more general factors (with singular precision matrices) is not possible.
	
	\revision{A popular practical solution to deal with degeneracies} is to employ ridge regularisation, where a widely-used strategy is to add a small scalar value to the diagonal terms of the covariance matrix \cite{hoerl1970}. This not only ensures positive definiteness, but also that the condition number of the matrix remains manageable. Unfortunately, this added robustness often comes at the cost of a decrease in accuracy \cite{cortes2010impact}. \revision{As a result, there is always a trade-off between more numerically stable computations for larger regularisation values and less biased estimates otherwise. Although a number of studies have investigated optimal strategies for designing these regularisation values \cite{golub1979generalized,heckman2000penalized},} whether this approximation is tolerable or not \revision{ultimately} depends on the particular application \cite{marquardt1975ridge}.
	
	\revision{In contrast to ridge regularisation, Mikheev \cite{mikheev2006multidimensional} proposes an explicit representation for degenerate Gaussian distributions. For the special case where the mean is zero, they express the density function as the product of an exponential factor and a multi-dimensional Dirac delta function. This is then simplified to an expression for the density function that is only valid on the lower-dimensional manifold where the linear constraints are satisfied. However, their final result does not include a Dirac delta function and looks very similar to the standard Gaussian density where the matrix inverse is just replaced with a pseudo-inverse. By first making sure that the given vector satisfies the linear constraints imposed by the covariance matrix, this definition can then be used to calculate the relative likelihood. Mikheev \cite{mikheev2006multidimensional} does not, however, provide any means for performing inference with this representation.}
	
	The approach by Raphael \cite{raphael2003bayesian} is the only one that \revision{we are aware of that both represents parametrised degenerate factors explicitly and then also provides a means to use this representation to perform inference. In this case, the factors are represented} as the product of an exponential factor and a multivariate indicator function. For this purpose, Raphael defines three mutually orthogonal subspaces that form a decomposition of $\mathbb{R}^n$ according to (a) \revision{components} associated with finite variance, (b) those that need to satisfy the implied linear constraints, and (c) those that the factor does not depend on. Raphael then derives the necessary operations for inference algorithms and demonstrates their application to an example of automatic musical accompaniment.
	
	In addition to the three subspaces, which are represented by bases in the columns of matrices, Raphael's parametrisation includes an $n$-dimensional mean vector and two non-negative definite $n\times n$ matrices. \revision{The latter} are equivalent to the covariance and precision matrices in non-degenerate cases. Although this \revision{representation} makes the notation and results of some statistical operations more concise, it is over-parametrised. More specifically, \revision{it should not be necessary to store multiple parameters that can be obtained from one another by computing, for example, pseudo-inverses or column spaces.} In addition, \revision{Raphael mentions that this} definition cannot express unnormalised factors, which prohibits \revision{Bayesian} model comparison \cite{koller2009}. \revision{This limitation is ultimately due to their use of an indicator function, where the Dirac delta proposed by Mikheev \cite{mikheev2006multidimensional} provides a less-restrictive alternative. Finally, they do not} provide a method for approximating nonlinear models. \revision{This is especially relevant when using the unscented transform, where deterministic samples cannot be drawn naively from a singular covariance matrix.}
	
	\paragraph{Summary of contributions} In this paper, we present a generalised representation for Gaussian factors that can handle statistical operations in degenerate cases both automatically and accurately. \revision{This is in contrast to ridge regularisation \cite{hoerl1970} which introduces unnecessary approximations.} Our parametric form makes use of a canonical factor with a diagonal precision matrix to express the uncertainty in the (possibly) lower-dimensional manifold and a Dirac delta function to represent any linear dependencies. \revision{Unlike the work by Mikheev \cite{mikheev2006multidimensional}, we keep the Dirac delta functions as part of the representation to enable the derivation of inference operations. This is also in contrast to Raphael's \cite{raphael2003bayesian} use of the indicator function.}
	
	\revision{Unlike Raphael, we only keep track of two complementary subspaces. This is because canonical factors themselves can already express random variables with zero precision (or infinite variance).} The bases for the appropriate decomposition into \revision{these} two subspaces are then represented by the columns of two semi-orthogonal matrices. \revision{By using a \emph{diagonal} precision matrix, our representation achieves the same expressibility as that by Raphael with only $n^2+2n$ scalar parameters, as opposed to their $3n^2+n$ parameters. An added benefit is that deterministic samples can then also be drawn from a singular covariance matrix in a straightforward manner.}
	
	\revision{Importantly,} our representation explicitly keeps track of the normalisation constant, thereby allowing model comparison and maximum a posteriori (MAP) estimation. \revision{This is not possible in degenerate settings with any existing approaches.} We subsequently derive typical inference operations for our representation from first principles \revision{that are} applicable to both linear and nonlinear models. \revision{Lastly, this generalised representation also provides a method to represent the distribution over a rank-deficient transformation of non-degenerate Gaussian random variables. Collectively, these contributions have the implication that all the capabilities of Gaussian factors can be extended to degenerate settings without resorting to unnecessary approximations.}
	
	\paragraph{Structure of the paper} In Section \ref{sec:prelim} we briefly review typical operations required for inference on Gaussian networks. We also present the definition of both the one-dimensional and multidimensional Dirac delta functions and briefly review some relevant linear algebra concepts. In Section \ref{sec:degen_factors} we introduce our definition of the degenerate Gaussian factor. We also derive the first- and second-order moments of the generalised density function and cover the affine transformation of random variables. In Section \ref{sec:operations} we present the statistical operations (such as marginalisation, multiplication and reduction according to evidence) required for inference with degenerate factors. In Section \ref{sec:add_operations} we present further results that are necessary for performing inference on Bayesian networks. In Section \ref{sec:experiments} we present a recursive state estimation example that illustrate the advantage of explicitly representing and performing inference with degenerate Gaussian factors. Relevant identities related to canonical factors and Dirac delta functions are included in \ref{sec:canonical_identities} and the detailed derivations of the main results are documented in \ref{sec:proofs}.
	
	\newpage

\section{Preliminaries} \label{sec:prelim}

	\revision{To ensure that our development of degenerate factors is relevant, it is necessary to keep the context general enough.} Throughout this paper, we \revision{therefore} present the definition of \revision{and statistical operations on} degenerate Gaussian factors in the context of probabilistic graphical models \cite{koller2009,barber2012bayesian}. (This does not, however, limit the application of the subsequent development to this context alone.) For this purpose, we first provide an overview of Gaussian networks and specifically present the necessary operations for performing inference using canonical factors. These operations not only serve as the foundation for Section \ref{sec:operations}, but are themselves used in various proofs of the main results in \ref{sec:proofs}. We also present the definition of both the one-dimensional and multidimensional Dirac delta function, \revision{since} the latter forms part of our definition of degenerate factors in Section \ref{sec:degen_factors}. Finally, we briefly discuss concepts from the field of linear algebra that are relevant for our factor parametrisation and used throughout the derivations of the subsequent statistical operations.

	\subsection{Gaussian networks and canonical factors} \label{sec:can_factors}
	
	
	We use the term ``Gaussian network" to refer to \revision{any probabilistic model} where the joint distribution \revision{over all the random variables} is Gaussian. For representing parametrised factors in \revision{non-degenerate} Gaussian networks, Koller and Friedman \cite{koller2009} define the canonical factor over an $n$-dimensional vector $\mathbf{x}$ as
	\begin{linenomath*}\begin{equation} \label{eq:canon_def}
	\mathcal{C}\left(\mathbf{x};K,\mathbf{h},g\right) \triangleq \exp\left(-\frac{1}{2}\mathbf{x}^TK\mathbf{x}+\mathbf{h}^T\mathbf{x}+g\right),
	\end{equation}\end{linenomath*}
	where $K$ is a symmetric $n\times n$ precision matrix, $\mathbf{h}$ is an $n$-dimensional vector and $g$ is a scalar normalisation constant. The factor in \eqref{eq:canon_def} is a normalised probability density function if and only if the matrix $K$ is positive definite and
	\begin{linenomath*}\begin{equation} \label{eq:g_for_pdf}
	g=-\frac{1}{2}\mathbf{h}^TK^{-1}\mathbf{h}-\frac{1}{2}\log\left|2\pi K^{-1}\right|.
	\end{equation}\end{linenomath*}
	Any Gaussian density function with mean vector $\boldsymbol{\mu}$ and covariance matrix $\Sigma$ can be expressed as such a canonical factor \cite{koller2009}. A very useful property of the canonical factor in \eqref{eq:canon_def} is that it is closed under typical \revision{statistical} operations \revision{required for inference algorithms.}
	
	
	\revision{In the general case of probabilistic graphical models,} message-passing algorithms such as belief propagation \cite{shenoy1986propagating} or belief update \cite{lauritzen1988local} are a popular choice for performing inference. \revision{In the special case of canonical factors, Koller and Friedman \cite{koller2009} derive the statistical operations required for these algorithms as follows:} Firstly, for the partitioned canonical factor
	\begin{linenomath*}\begin{equation} \label{eq:multiNormal}
	\phi(\mathbf{x},\mathbf{y})=\mathcal{C}\left(\begin{bmatrix}
	\mathbf{x} \\
	\mathbf{y}
	\end{bmatrix};\begin{bmatrix}
	K_{\mathbf{xx}} & K_{\mathbf{xy}} \\
	K_{\mathbf{yx}} & K_{\mathbf{yy}}
	\end{bmatrix},\begin{bmatrix}
	\mathbf{h}_{\mathbf{x}} \\
	\mathbf{h}_{\mathbf{y}}
	\end{bmatrix},g\right),
	\end{equation}\end{linenomath*}
	\emph{marginalising} over $\mathbf{y}$ results in the marginal canonical factor
	\begin{linenomath*}\begin{equation} \label{eq:canonical_marg}
	\int \phi(\mathbf{x},\mathbf{y})\,\text{d}\mathbf{y}=\mathcal{C}\left(\mathbf{x};K_{\mathbf{xx}}^{}-K_{\mathbf{xy}}^{}K_{\mathbf{yy}}^{-1}K_{\mathbf{yx}}^{},\mathbf{h}_{\mathbf{x}}-K_{\mathbf{xy}}^{}K_{\mathbf{yy}}^{-1}\mathbf{h}_{\mathbf{y}}^{},g + \frac{1}{2}\mathbf{h}_{\mathbf{y}}^TK_{\mathbf{yy}}^{-1}\mathbf{h}_{\mathbf{y}}^{} + \frac{1}{2}\log\left|2\pi K_{\mathbf{yy}}^{-1}\right| \right).
	\end{equation}\end{linenomath*}
	Furthermore, the \emph{product} of two canonical factors over the same scope $\mathbf{x}$ is given by
	\begin{linenomath*}\begin{equation} \label{eq:canonical_multiply}
	\mathcal{C}(\mathbf{x};K_1,\mathbf{h}_1,g_1)\,\mathcal{C}(\mathbf{x};K_2,\mathbf{h}_2,g_2)=\mathcal{C}(\mathbf{x};K_1+K_2,\mathbf{h}_1+\mathbf{h}_2,g_1+g_2).
	\end{equation}\end{linenomath*}
	Similarly, the \emph{quotient} of two canonical factors is given by
	\begin{linenomath*}\begin{equation} \label{eq:canonical_divide}
	\frac{\mathcal{C}(\mathbf{x};K_1,\mathbf{h}_1,g_1)}{\mathcal{C}(\mathbf{x};K_2,\mathbf{h}_2,g_2)}=\mathcal{C}(\mathbf{x};K_1-K_2,\mathbf{h}_1-\mathbf{h}_2,g_1-g_2).
	\end{equation}\end{linenomath*}
	Finally, the canonical factor in \eqref{eq:multiNormal} can be reduced according to available \emph{evidence}. Setting $\mathbf{y}=\mathbf{y}_0$ results in
	\begin{linenomath*}\begin{equation} \label{eq:canonical_evid}
	\phi(\mathbf{x},\mathbf{y}_0)=\mathcal{C}(\mathbf{x};K_{\mathbf{xx}}^{},\mathbf{h}_{\mathbf{x}}^{}-K_{\mathbf{xy}}^{}\mathbf{y}_0^{},g+\mathbf{h}_{\mathbf{y}}^T\mathbf{y}_0^{}-\frac{1}{2}\mathbf{y}_0^TK_{\mathbf{yy}}\mathbf{y}_0^{}).
	\end{equation}\end{linenomath*}
	The results in \eqref{eq:canonical_marg} to \eqref{eq:canonical_evid} form the building blocks for message-passing algorithms in Gaussian networks. To perform inference using a more general factor representation that can accommodate degeneracies therefore requires similar results.
	
	\subsection{Dirac delta functions}
	
	The Dirac delta function \cite{dirac1981principles} is a distribution (or generalised function) that satisfies the two properties
	\begin{linenomath*}\begin{equation} \label{eq:dirac_prop}
	\delta(x)=0,\text{ for } x\neq 0\qquad\text{and}\qquad\int_{-\infty}^{\infty} \delta(x)\,dx=1.
	\end{equation}\end{linenomath*}
	Due to the first property in \eqref{eq:dirac_prop}, an informal (but useful) interpretation of the Dirac delta is as imposing \revision{the constraint $x=0$} on its argument. According to Shankar \cite{shankar2012principles}, a possible definition for the one-dimensional Dirac delta that makes use of a Gaussian distribution (with zero mean and variance $a$) is
	\begin{linenomath*}\begin{equation} \label{eq:Dirac_as_Gaussian}
	\delta(x) \triangleq \lim_{a \rightarrow 0}\mathcal{N}\left(x;0,a\right).
	\end{equation}\end{linenomath*}
	\noindent\revision{Furthermore,} the multidimensional Dirac delta function over a $k$-dimensional vector $\mathbf{x}$ is defined as
	\begin{linenomath*}\begin{equation} \label{eq:ND_Diracdelta}
	\delta\left(\mathbf{x}\right) \triangleq \prod_{i=1}^k\delta\left(x_i\right).
	\end{equation}\end{linenomath*}
	After combining \eqref{eq:Dirac_as_Gaussian} and \eqref{eq:ND_Diracdelta}, we can write the multidimensional Dirac delta as
	\begin{linenomath*}\begin{equation} \label{eq:ND_Dirac_as_Gaussian}
	\delta(\mathbf{x}) = \lim_{a \rightarrow 0}\mathcal{N}\left(\mathbf{x};\mathbf{0},aI\right) = \lim_{a \rightarrow 0}(2\pi a)^{-\frac{k}{2}}\exp\left(-\frac{1}{2a}\mathbf{x}^T\mathbf{x}\right).
	\end{equation}\end{linenomath*}
	This will be used in Section \ref{sec:degen_factors} to model the degenerate component of our novel factor parametrisation. Note that the definition of the Dirac delta as the limit of an exponential term conveniently resembles that of the canonical factor in \eqref{eq:canon_def}.
	
	\subsection{Fundamental subspaces and the singular value decomposition} \label{sec:linear_algebra}
	
	A useful perspective on the properties of and the operations on a given matrix $A_{m\times n}$ (with $m$ rows and $n$ columns) are what Strang \cite{strang09} calls its four fundamental subspaces. These are the column space, nullspace, row space and left nullspace. The \emph{column space} (also called the \emph{range}) of $A$ refers to the set of all possible vectors that can be constructed from a linear combination of the columns of $A$ and is denoted by $C(A)\subseteq\mathbb{R}^m$. The \emph{nullspace} (also called the \emph{kernel}) of $A$ is the set of all solutions $\mathbf{x}$ to the equation $A\mathbf{x}=\mathbf{0}$ and is denoted by $N(A)\subseteq\mathbb{R}^n$. The \emph{row space} \revision{$C\left(A^T\right)$} and the \emph{left nullspace} \revision{$N\left(A^T\right)$} of $A$ are simply the column space and nullspace of $A^T$, \revision{respectively.} For any matrix $A$, the column space and left nullspace are orthogonal complements\footnote{Two subspaces $V$ and $W$ are said to be orthogonal complements if every vector $\mathbf{v}\in V$ is orthogonal to every vector $\mathbf{w}\in W$, i.e., $\mathbf{v}^T\mathbf{w}=0$.}. \revision{We can therefore write that} $C(A)=N(A^T)^\perp$.
	
	A popular method for computing orthonormal bases for the fundamental subspaces of a matrix is using the singular value decomposition (SVD) \cite{stewart1993early}. \revision{The reason for this is that} the SVD (with complexity $\mathcal{O}(n^3)$) is numerically stable compared to other matrix decompositions and enjoys efficient implementations across many different platforms \cite{strang09}. In addition, the SVD can be used to compute the pseudo-inverse $A^+$ as well as an orthonormal basis for the orthogonal complement $C(A)^\perp$ \cite{golub2013matrix}. For a positive-definite matrix, the ratio of its largest to its smallest singular value is known as its \emph{condition number} and is denoted by $\kappa(A)$. On machines with finite precision, computations involving matrices with large condition numbers lead to a decrease in accuracy (due to the loss of significant digits) and results in numeric instability beyond a certain value. \revision{Consequently, using the SVD as part of a practical implementation often requires thresholding \cite{raphael2003bayesian}}. These linear algebra ideas are used throughout the rest of this paper -- both in our definition of degenerate factors as well as in the results and proofs for the subsequent statistical operations.
	
\section{Degenerate Gaussian factors} \label{sec:degen_factors}

	A popular approach for performing inference with continuous random variables is to employ parametrised factors. In the special case of Gaussian networks, the canonical factor in \eqref{eq:canon_def} is a good example. In this section, we combine a canonical factor with a Dirac delta function to form a novel parametrised factor that can express possible degeneracies (i.e., settings where linear dependencies exist among the random variables). We also discuss the requirements for this factor to be a normalised density function and derive its moments. Since our definition is a generalisation of traditional parametrisations, non-degenerate Gaussian densities can still be represented. Finally, we show that explicitly representing degeneracies within a joint density also enables \revision{rank-deficient} affine transformations of Gaussian random variables.

	\subsection{Definition and parametrisation}
	
\begin{figure}[htb]
		\centering
		\subfigure[]{\label{fig:degen_definition_1}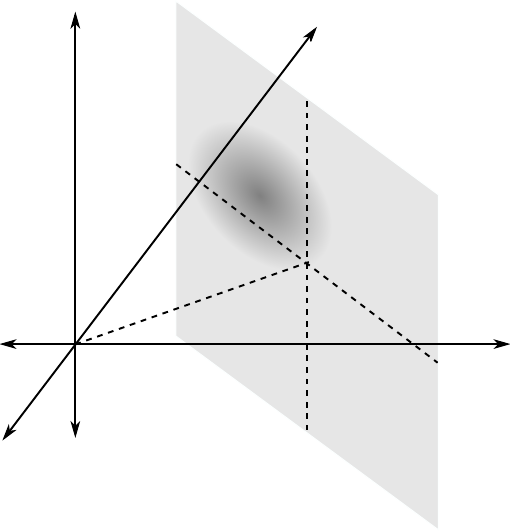}
		\hspace{0.05\linewidth}
		\subfigure[]{\label{fig:degen_definition_2}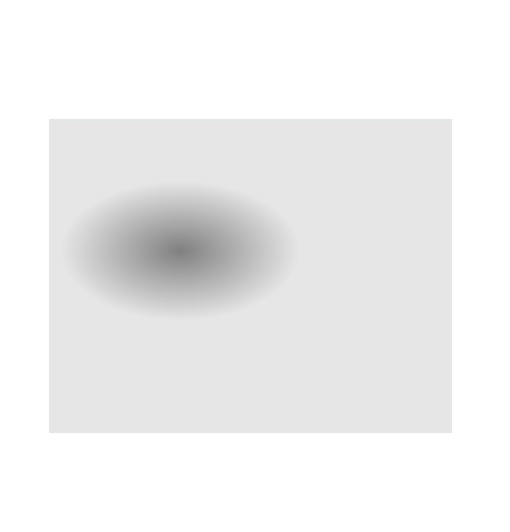}
		\caption{(a) A visualisation of a degenerate factor in $\mathbb{R}^3$ subject to the linear constraint $x_1+x_2=a$. The variation in shading represents the scalar value of the canonical \revision{component} in the affine space and the dashed lines indicate the \revision{principal} axes. Away from this plane (along the vector $\mathbf{r}$) the degenerate factor evaluates to zero.  The shortest (perpendicular) distance from the origin in $\mathbb{R}^3$ to the plane is $c$. (b) From a view perpendicular to the affine plane, we see that the vectors $\mathbf{q}_1$ and $\mathbf{q}_2$ have been chosen to align with the principal axes \revision{of the lower-dimensional distribution.} The corresponding standard deviations are related to the precisions according to $\sigma_1=1/\sqrt{\lambda_1}$ and $\sigma_2=1/\sqrt{\lambda_2}$.}
		\label{fig:degen_definition}
	\end{figure}

	Using the definitions \revision{of the canonical factor} in \eqref{eq:canon_def} and \revision{the Dirac delta function} in \eqref{eq:ND_Dirac_as_Gaussian}, we represent a degenerate Gaussian factor over an $n$-dimensional vector $\mathbf{x}$ (with $k$ degrees of degeneracy) as the product
	\begin{linenomath*}\begin{align} \label{eq:degen_def}
	\mathcal{D}(\mathbf{x};Q,R,\Lambda,\mathbf{h},\mathbf{c},g) &\triangleq \revision{\exp\left(-\frac{1}{2}\mathbf{x}^TQ\Lambda Q^T\mathbf{x}+\mathbf{h}^TQ^T\mathbf{x}+g\right)\,\delta(R^T\mathbf{x}-\mathbf{c})} \nonumber \\
	&=\mathcal{C}(Q^T\mathbf{x};\Lambda,\mathbf{h},g)\,\delta(R^T\mathbf{x}-\mathbf{c}).
	\end{align}\end{linenomath*}
	In this parametrisation, $\Lambda$ is a non-negative\footnote{We restrict our parametrisation to positive semi-definite precision matrices. Negative-definite precision matrices that may occur due to approximations (as for example in expectation propagation \cite{minka2001expectation}) should be handled in a similar way as for non-degenerate representations.}, diagonal precision matrix describing the uncertainty in the $(n-k)$-dimensional affine space where the factor is non-zero. The vector $\mathbf{h}$ and normalisation constant $g$ determine the location and height of the maximum point in this space, respectively. The vector $\mathbf{c}$ in turn describes the minimum offset from the coordinate origin to the affine space. For \revision{mathematical} convenience \revision{in later sections,} the column spaces of \revision{the semi-orthogonal matrices} $Q_{n\times (n-k)}$ and $R_{n\times k}$ are constrained to be orthogonal complements. \revision{A useful interpretation of the degenerate factor in \eqref{eq:degen_def} is as a lower-dimensional, non-degenerate factor (parametrised by $\Lambda$, $\mathbf{h}$ and $g$) expanded to a higher-dimensional space through an affine transformation (parametrised by $Q$, $R$ and $\mathbf{c}$).} Some of these properties are illustrated in Figure \ref{fig:degen_definition}.
	
	\revision{When comparing our representation in \eqref{eq:degen_def} to those by Mikheev \cite{mikheev2006multidimensional} and Raphael \cite{raphael2003bayesian}, we can highlight a couple of key differences: Firstly, neither of these existing approaches are defined for unnormalised factors, where the former is further limited to the special case where the mean is zero. In addition, Mikheev's use of a covariance matrix has the implication that variables with zero precision cannot be represented. Although Raphael addresses this limitation using three complementary subspaces, our representation provides the same capabilities with only two. Since all of our subsequent statistical operations are unchanged whether there are trivial singular values or not, it is unnecessary to explicitly separate the finite- and infinite-variance space. Furthermore, only three of Raphael's parameters are actually used in their over-parametrised factor definition and the check whether the given vector $\mathbf{x}$ lies on the lower-dimensional manifold is simplified to $R^T\mathbf{x}=\mathbf{c}$ in our parametrisation. Finally, by using a diagonal precision matrix as part of a factorisation of the quadratic term in \eqref{eq:degen_def}, the cost of some of the subsequent matrix computations is reduced.}
	
	\subsection{The degenerate density function}
	
	For the special case where the factor in \eqref{eq:degen_def} is a valid density function $p(\mathbf{x})$, we require that
	\begin{linenomath*}\begin{equation} \label{eq:integral_density}
	\int_{-\infty}^{\infty} p(\mathbf{x})\,\text{d}\mathbf{x} = \int_{-\infty}^{\infty}\mathcal{C}(Q^T\mathbf{x};\Lambda,\mathbf{h},g)\,\delta(R^T\mathbf{x}-\mathbf{c})\,\text{d}\mathbf{x} = 1.
	\end{equation}\end{linenomath*}
	In addition, we require that the density function is non-negative, but this is already true for the general factor in \eqref{eq:degen_def}. The integral in \eqref{eq:integral_density} can be rewritten using the substitutions $\boldsymbol{\epsilon}=Q^T\mathbf{x}$ and $\boldsymbol{\eta}=R^T\mathbf{x}$. Since $Q$ and $R$ are orthogonal complements, this change of variables is orthogonal and (since the integral over a Dirac delta is unity) we therefore require that
	\begin{linenomath*}\begin{equation}
	\int_{-\infty}^{\infty}\mathcal{C}(\boldsymbol{\epsilon};\Lambda,\mathbf{h},g)\,\delta(\boldsymbol{\eta}-\mathbf{c})\,\text{d}\boldsymbol{\epsilon}\,\text{d}\boldsymbol{\eta}=\int_{-\infty}^{\infty}\mathcal{C}(\boldsymbol{\epsilon};\Lambda,\mathbf{h},g)\,\text{d}\boldsymbol{\epsilon}=1.
	\end{equation}\end{linenomath*}
	Consequently, we can use the result in \eqref{eq:g_for_pdf} to see that
	\begin{linenomath*}\begin{equation} \label{eq:g_norm}
	g=-\frac{1}{2}\mathbf{h}^T\Lambda^{-1}\mathbf{h}-\frac{1}{2}\log\left|2\pi\Lambda^{-1}\right|.
	\end{equation}\end{linenomath*}
	This also requires that $\Lambda$ is non-singular. Note that, since $\Lambda$ is diagonal, the matrix inverse in \eqref{eq:g_norm} is straightforward and can be computed in linear time. This computational advantage is applicable to many operations throughout the rest of this paper.
	
	We can determine the moments (specifically the mean and the covariance) of a degenerate Gaussian density by substituting the definition of the Dirac delta function in \eqref{eq:ND_Dirac_as_Gaussian} into \eqref{eq:degen_def}, resulting in
	\begin{linenomath*}\begin{equation} \label{eq:joint_lim_degen}
	p(\mathbf{x})=\lim_{a \rightarrow 0}\mathcal{N}\left(\mathbf{x};Q\Lambda^{-1}\mathbf{h}+R\mathbf{c},Q\Lambda^{-1}Q^T+aRR^T
	\right).
	\end{equation}\end{linenomath*}
	The extended derivation of \eqref{eq:joint_lim_degen} is given in Result \ref{result:joint_lim_degen_complete} in \ref{sec:canonical_identities}. For the mean, we can use \eqref{eq:joint_lim_degen} to calculate the expected value
	\begin{linenomath*}\begin{equation} \label{eq:mean_degen}
	\mathbb{E}[\mathbf{x}]=\lim_{a \rightarrow 0}\left(Q\Lambda^{-1}\mathbf{h}+R\mathbf{c}\right)=Q\Lambda^{-1}\mathbf{h}+R\mathbf{c}.
	\end{equation}\end{linenomath*}
	For the covariance, we obtain
	\begin{linenomath*}\begin{equation} \label{eq:cov_degen}
	\text{Cov}\left[\mathbf{x}\right] = \lim_{a \rightarrow 0} \left(Q\Lambda^{-1}Q^T+aRR^T\right)=Q\Lambda^{-1}Q^T.
	\end{equation}\end{linenomath*}
	For $k$ degrees of degeneracy, the covariance will therefore have rank $n-k$.

	Conveniently, any non-degenerate Gaussian density function with mean vector $\boldsymbol{\mu}$ and covariance matrix $\Sigma$ can still be expressed as a degenerate Gaussian factor such that
	\begin{linenomath*}\begin{equation} \label{eq:density_convert}
	p(\mathbf{x})=\mathcal{N}(\mathbf{x};\boldsymbol{\mu},\Sigma)=\mathcal{D}(\mathbf{x};Q,R,\Lambda,\mathbf{h},\mathbf{c},g).
	\end{equation}\end{linenomath*}
	In this non-degenerate case, $C(Q)=\mathbb{R}^n$ and subsequently $C(R)=\{\mathbf{0}\}$. The other extreme where $C(Q)=\{\mathbf{0}\}$ and $C(R)=\mathbb{R}^n$ in turn relates to zero uncertainty and $n$ degrees of degeneracy. Although a subset of the parameters in \eqref{eq:degen_def} will be empty in both of these cases, the degenerate factor is still well defined under the convention that canonical factors and Dirac delta functions with empty arguments equate to unity.
	
	Returning to the unknown parameters in \eqref{eq:density_convert}, we can use  \eqref{eq:cov_degen} to write
	\begin{linenomath*}\begin{equation} \label{eq:svd_sigma}
	\Sigma=Q\Lambda^{-1}Q^T.
	\end{equation}\end{linenomath*}
	Since $\Sigma$ is symmetric and positive definite, $Q$ and $\Lambda$ can be calculated directly using the SVD. With the values of $Q$ and $\Lambda$ in \eqref{eq:density_convert} known, we proceed to calculate the quantities $\mathbf{h}$ and $g$. Note that in the non-degenerate case (i.e., when $k=0$), the factor in \eqref{eq:degen_def} as well as the expectation in \eqref{eq:mean_degen} no longer depend on the empty arguments $R$ and $\mathbf{c}$. Since $Q$ is orthogonal, the reduced form of the latter yields
	\begin{linenomath*}\begin{equation} \label{eq:h_mu}
	\mathbf{h}=\left(Q\Lambda^{-1}\right)^{-1}\boldsymbol{\mu}=\Lambda Q^T\boldsymbol{\mu}.
	\end{equation}\end{linenomath*}
	Finally, substituting \eqref{eq:h_mu} followed by \eqref{eq:svd_sigma} into \eqref{eq:g_norm} yields
	\begin{linenomath*}\begin{equation}
	g=-\frac{1}{2}\boldsymbol{\mu}^TQ\Lambda Q^T\boldsymbol{\mu}-\frac{1}{2}\log\left|2\pi\Lambda^{-1}\right|=-\frac{1}{2}\boldsymbol{\mu}^T\Sigma^{-1}\boldsymbol{\mu}-\log\sqrt{(2\pi)^n|\Sigma|},
	\end{equation}\end{linenomath*}
	which is equivalent to the normalisation constant in traditional parametrisations. \eqref{eq:density_convert} illustrates the idea that our representation for degenerate Gaussian factors is a generalisation of existing parametrisations that relaxes positive definiteness constraints.
	
	\subsection{Affine transformations of degenerate random variables}
	
	Among the many useful properties of Gaussian random variables is the fact that affine transformations
	\begin{linenomath*}\begin{equation} \label{eq:y_Ax_b}
	\mathbf{y}=A\mathbf{x}+\mathbf{b}
	\end{equation}\end{linenomath*}
	thereof are still Gaussian-distributed. However, one important condition for the use of traditional parametrisations to express the resulting density function is, once again, that the resulting covariance matrix must be positive definite. Since $\text{Cov}\left[\mathbf{y}\right] = A\,\text{Cov}\left[\mathbf{x}\right]A^T$, this in turn restricts the matrix $A$ to have full row rank. By instead using the parametrisation in \eqref{eq:degen_def} to express the prior density
	\begin{linenomath*}\begin{equation}
	p(\mathbf{x})=\mathcal{D}(\mathbf{x};Q,R,\Lambda,\mathbf{h},\mathbf{c},g),
	\end{equation}\end{linenomath*}
	we can relax this constraint and represent the density $p(\mathbf{y})$ for \emph{any} matrix $A$.
	
	Since an affine transformation only alters the respective subspaces associated with \revision{(a)} jointly Gaussian-distributed random variables and \revision{(b)} linear dependencies, $p(\mathbf{y})$ will also be a degenerate Gaussian density with parameters $Q'$, $R'$, $\Lambda'$, $\mathbf{h}'$, $\mathbf{c}'$ and $g'$. The procedure to calculate these parameters -- derived through moment matching using the results in \eqref{eq:mean_degen} and \eqref{eq:cov_degen} -- is summarised in Algorithm \ref{alg:affine_transform}. The detailed derivation is included in \ref{sec:proofs}. In this as well as future algorithms we use the tuple $\theta=(Q,R,\Lambda,\mathbf{h},\mathbf{c},g)$ as a shorthand for the parameters.
	
	\begin{algorithm}[htb]
		\caption{\texttt{AffineTransformation}} \label{alg:affine_transform}
		\begin{algorithmic}[1]
			\Require $\theta$, $A$, $\mathbf{b}$ such that $p(\mathbf{x})=\mathcal{D}(\mathbf{x};\theta)$ and $\mathbf{y}=A\mathbf{x}+\mathbf{b}$
			\Ensure $\theta'$ such that $p(\mathbf{y})=\mathcal{D}(\mathbf{y};\theta')$
			\State $Q^\prime,\Sigma',\_\gets\texttt{CompactSVD}\left(AQ\Lambda^{-1}Q^TA^T\right)$
			\State $\Lambda'\gets \Sigma'^{-1}$
			\State $R^\prime\gets \texttt{Complement}(Q^\prime)$
			\State $\boldsymbol{\mu}'\gets A(Q\Lambda^{-1}\mathbf{h}+R\mathbf{c})+\mathbf{b}$
			\State $\mathbf{c}^\prime \gets R^{\prime T}\boldsymbol{\mu}'$
			\State $\mathbf{h}^\prime_{}\gets \Lambda'Q^{\prime T}\boldsymbol{\mu}'$
			\State $g^\prime\gets -\frac{1}{2}\mathbf{h}'^T\Lambda'^{-1}\mathbf{h}'-\frac{1}{2}\log\left|2\pi\Lambda'^{-1}\right|$
		\end{algorithmic}
	\end{algorithm}

	Since the resulting covariance matrix might only be positive semi-definite, the quantities $Q'$ and $\Lambda'$ are calculated (in lines 1 and 2 of Algorithm \ref{alg:affine_transform}) using the compact SVD. Note that the resulting degree of degeneracy in $p(\mathbf{y})$ depends on the mutual properties of $A$ and $Q$. Next, the semi-orthogonal matrix $R'$ is computed (in line 3) such that $C(R')=C(Q')^\perp$. The mean vector $\boldsymbol{\mu}'=\mathbb{E}[\mathbf{y}]$ is calculated (in line 4) by substituting the expectation in \eqref{eq:mean_degen} into \eqref{eq:y_Ax_b}. As in \eqref{eq:h_mu}, $\mathbf{c}'$ and $\mathbf{h}'$ are then calculated (in lines 5 and 6) by exploiting the mutual orthogonality of $R'$ and $Q'$. Since the resulting density $p(\mathbf{y})$ will be normalised, $g'$ is calculated (in line 7) by direct application of \eqref{eq:g_norm}.

	In summary, our definition of degenerate Gaussian factors in \eqref{eq:degen_def} can express $k\in [0,n]$ degeneracies. It comprises a canonical factor and Dirac delta component, as necessary. For the degenerate factor to represent a normalised density function, the canonical factor must be normalised as in \eqref{eq:g_norm}. The mean of a degenerate density in \eqref{eq:mean_degen} is then the sum of two orthogonal vectors in the respective subspaces and the rank of the covariance matrix in \eqref{eq:cov_degen} is equal to $n-k$. Any non-degenerate Gaussian density can be converted to the degenerate parametrisation as in \eqref{eq:density_convert}. Finally, affine transformations of degenerate (or non-degenerate) Gaussian random variables are also distributed according to a degenerate Gaussian density.

\section{Statistical operations on degenerate factors} \label{sec:operations}

	To use degenerate Gaussian factors for inference, the appropriate statistical operations need to be derived. In the context of message-passing algorithms, these include marginalisation, multiplication, division, and reduction according to available evidence. Following the notion that the degenerate factor is a generalisation of the canonical factor, the results in this section mirror those in Section \ref{sec:can_factors}. We show a procedure to perform each operation in the form of an algorithm and provide visualisations of simple examples where appropriate. The detailed derivations of the results are included in \ref{sec:proofs}. As \revision{in the non-degenerate case,} it is desirable that the result of each operation is once again a degenerate Gaussian factor parametrised according to \eqref{eq:degen_def}.

	\subsection{Marginalisation}
	
	For the partitioned degenerate factor
	\begin{linenomath*}\begin{equation} \label{eq:jointDegen}
	\phi(\mathbf{x},\mathbf{y})=\mathcal{D}\left(\begin{bmatrix}
	\mathbf{x} \\
	\mathbf{y}
	\end{bmatrix};\begin{bmatrix}
	Q_{\mathbf{x}} \\
	Q_{\mathbf{y}}
	\end{bmatrix},\begin{bmatrix}
	R_{\mathbf{x}} \\
	R_{\mathbf{y}}
	\end{bmatrix},\Lambda,\mathbf{h},\mathbf{c},g\right),
	\end{equation}\end{linenomath*}
	marginalising over $\mathbf{y}$ results in the marginal factor
	\begin{linenomath*}\begin{equation} \label{eq:degenerate_marginal}
	\int \phi(\mathbf{x},\mathbf{y})\,\text{d}\mathbf{y}=\mathcal{D}\left(\mathbf{x};Q',R',\Lambda', \mathbf{h}',\mathbf{c}',g'\right),
	\end{equation}\end{linenomath*}
	where the parameters $Q'$, $R'$, $\Lambda'$, $\mathbf{h}'$, $\mathbf{c}'$ and $g'$ are determined according to Algorithm \ref{alg:marginal}. Since marginalisation amounts to projecting to a subspace, we use an appropriate orthogonal decomposition (according to the properties of the four block matrices $Q_{\mathbf{x}}$, $Q_{\mathbf{y}}$, $R_{\mathbf{x}}$ and $R_{\mathbf{y}}$) to handle possible degeneracies before marginalising the resulting canonical factor as in \eqref{eq:canonical_marg}.
	
	\begin{algorithm}[htb]
		\caption{\texttt{Marginalise}} \label{alg:marginal}
		\begin{algorithmic}[1]
			\revision{
			\Require $\theta$ such that $\phi(\mathbf{x},\mathbf{y})=\mathcal{D}\left(\begin{bmatrix}
			\mathbf{x} \\
			\mathbf{y}
			\end{bmatrix};\theta\right)$
			\Ensure $\theta'$ such that $\int \phi(\mathbf{x},\mathbf{y})\,\text{d}\mathbf{y}=\mathcal{D}(\mathbf{x};\theta')$
			\State $U\gets \texttt{Columnspace}\left(Q_{\mathbf{x}}\right)$, $V\gets \texttt{Nullspace}\left(Q_{\mathbf{x}}\right)$, $W\gets \texttt{Columnspace}\left(R_{\mathbf{x}}^TQ_{\mathbf{x}}^{}\right)$
			\State $R^\prime_{} \gets \texttt{Complement}(U)$
			\State $\mathbf{c}^\prime \gets R'^TR_{\mathbf{x}}\mathbf{c}$
			\State $F\gets \left(W\left(R_{\mathbf{y}}W\right)^+Q_{\mathbf{y}}\right)^T$, $G\gets \left(Q_{\mathbf{x}}^T-FR_{\mathbf{x}}^T\right)U$, $S\gets V\left(V^T\Lambda V\right)^{-1}V^T$
			\State $Z,\Lambda^\prime,\_\gets\texttt{SVD}\left(G^T(\Lambda-\Lambda S\Lambda)\,G\right)$
			\State $Q^\prime_{}\gets UZ$
			\State $\mathbf{h}^\prime\gets Z^TG^T(I-\Lambda S)(\mathbf{h}-\Lambda F\mathbf{c})$
			\State $g^\prime_{}\gets g+\left(\mathbf{h}-\frac{1}{2} \Lambda F\mathbf{c} \right)^T\,F\mathbf{c}+\frac{1}{2}(\mathbf{h}-\Lambda F\mathbf{c})^TS(\mathbf{h}-\Lambda F\mathbf{c})+\frac{1}{2}\log\frac{\left|2\pi(V^T\Lambda V)^{-1}\right|}{\left|W_{}^TR_{\mathbf{y}}^TR_{\mathbf{y}}^{}W\right|}$
			}
		\end{algorithmic}
	\end{algorithm}
	
\begin{figure}[htb]
		\centering
		\subfigure[]{\label{fig:degen_marginal_1}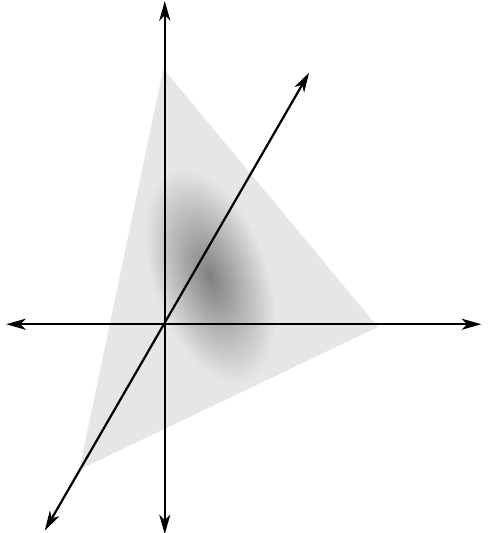}
		\hspace{0.05\linewidth}
		\subfigure[]{\label{fig:degen_marginal_2}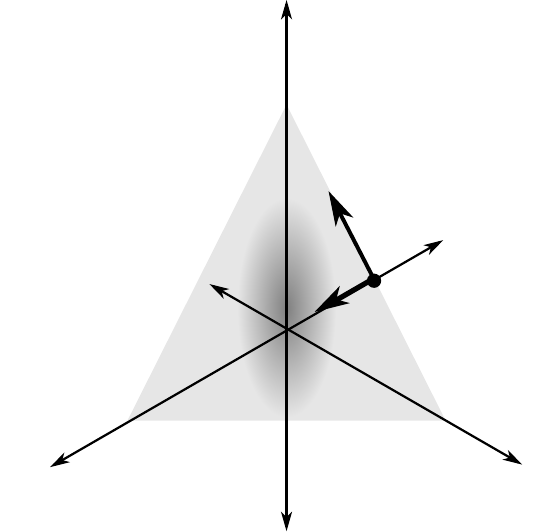}
		\caption{\revision{The marginalisation over $\{y_1,y_2\}$} of (a) a degenerate factor in $\mathbb{R}^3$ subject to the linear constraint \revision{$x+y_1+y_2=a$. The vector $\mathbf{r}$ is perpendicular to the affine plane and the two vectors $\mathbf{q}_1$ and $\mathbf{q}_2$ align with the principal axes.} (b) From a view perpendicular to \revision{this} plane, we see the 2-D point \revision{$\mathbf{f}c$. Furthermore,} the orthogonal 2-D vectors \revision{$\mathbf{g}$ and $\mathbf{v}$} align with the $x$-axis and the edge of the linear constraint in the $y_1$-$y_2$ plane, respectively. \revision{Collectively, these three components describe} an appropriate transformation within the affine plane based on the required marginalisation in the coordinate axes. The marginal factor over $x$ is then indicated by the dashed 1-D curve, where the corresponding origin is the projection of the point \revision{$\mathbf{f}c$} (where $x=0$ in $\mathbb{R}^3$).}
		\label{fig:degen_marginal}
	\end{figure}

	The first step (in line 1 of Algorithm \ref{alg:marginal}) is to \revision{compute appropriate bases $U$, $V$ and $W$ according to the block matrices $Q_{\mathbf{x}}$ and $R_{\mathbf{x}}$.} As mentioned in Section \ref{sec:linear_algebra}, this can be achieved using the SVD. Depending on the degree of degeneracy $k$, the dimension $n'$ of the vector $\mathbf{x}$ and the rank of each respective block matrix, some of these bases could be trivial. The degree of degeneracy $k'$ in the marginal factor is \revision{then} equal to \revision{$n'-\text{rank}(Q_{\mathbf{x}})$ and the basis for this marginal degeneracy in $\mathbb{R}^{n'}$ is $C(U)^\perp$ (in line 2). The corresponding offset (in line 3) is $R'^TR_{\mathbf{x}}\mathbf{c}$.} The three auxiliary matrices \revision{$F$, $G$ and $S$ (in line 4)} make the rest of the notation more concise. The resulting full-rank precision matrix is diagonalised using the SVD (in line 5). The corresponding orthogonal matrix $Z$ is then used (in line 6) to determine the $(n'-k')$-dimensional basis for the marginal support in $\mathbb{R}^{n'}$. Finally, the vector $\mathbf{h}'$ and normalisation constant $g'$ are also calculated (in lines 7 and 8, respectively). Figure \ref{fig:degen_marginal} illustrates various aspects of Algorithm \ref{alg:marginal} for a simple example.

	\subsection{Multiplication}
	
	The product of two degenerate factors over the same scope $\mathbf{x}$ is given by
	\begin{linenomath*}\begin{equation} \label{eq:degenerate_multiply}
	\mathcal{D}(\mathbf{x};Q_1,R_1,\Lambda_1,\mathbf{h}_1,\mathbf{c}_1,g_1)\,\mathcal{D}(\mathbf{x};Q_2,R_2,\Lambda_2,\mathbf{h}_2,\mathbf{c}_2,g_2)=\mathcal{D}(\mathbf{x};Q',R',\Lambda',\mathbf{h}',\mathbf{c}',g').
	\end{equation}\end{linenomath*}
	where the parameters $Q'$, $R'$, $\Lambda'$, $\mathbf{h}'$, $\mathbf{c}'$ and $g'$ are determined according to Algorithm \ref{alg:multiply}. Since each degenerate factor only has support on a lower-dimensional manifold, the affine space of the product will be the intersection of those of the two original factors. We therefore reduce each factor to this intersection and then multiply the resulting canonical factors as in \eqref{eq:canonical_multiply}.
	
	\begin{algorithm}[htb]
		\caption{\texttt{Multiply}} \label{alg:multiply}
		\begin{algorithmic}[1]
			\Require $\theta_1$, $\theta_2$ such that $\phi_1(\mathbf{x})=\mathcal{D}(\mathbf{x};\theta_1)$ and $\phi_2(\mathbf{x})=\mathcal{D}(\mathbf{x};\theta_2)$
			\Ensure $\theta'$ such that $\phi_1(\mathbf{x})\times\phi_2(\mathbf{x})=\mathcal{D}(\mathbf{x};\theta')$
			\State $V\gets\texttt{Columnspace}\left(Q_1^{}Q_1^TR_2^{}\right)$
			\State $R^\prime_{} \gets \begin{bmatrix}
			R_1 & V
			\end{bmatrix}$
			\State $\mathbf{b}\gets (R_2^TV)^{-1}(\mathbf{c}_2-R_2^TR_1^{}\mathbf{c}_1)$
			\State $\mathbf{c}^\prime \gets \begin{bmatrix}
			\mathbf{c}_1 \\
			\mathbf{b}
			\end{bmatrix}$
			\State $U\gets \texttt{Complement}\left(R'\right)$
			\State $Z,\Lambda^\prime,\_\gets\texttt{SVD}\left(U^T_{}(Q_1^{}\Lambda_1^{}Q_1^T+Q_2^{}\Lambda_2^{}Q_2^T)\,U\right)$
			\State $Q'\gets UZ$
			\State $\mathbf{h}^\prime_{}\gets Q'^T(Q_1^{}(\mathbf{h}_1^{}-\Lambda_1^{}Q_1^TV\mathbf{b})+Q_2^{}(\mathbf{h}_2^{}-\Lambda_2^{}Q_2^TR'\mathbf{c}'))$
			\State $g^\prime_{}\gets g_1^{}+g_2^{}+\left(\mathbf{h}_1-\frac{1}{2}\Lambda_1^{}Q_1^TV\mathbf{b}\right)^TQ_1^T V\mathbf{b}+\left(\mathbf{h}_2-\frac{1}{2}\Lambda_2^{}Q_2^TR'\mathbf{c}'\right)^TQ_2^T R'\mathbf{c}'-\frac{1}{2}\log\left|R_2^TV\right|^2$
		\end{algorithmic}
	\end{algorithm}

\begin{figure}[htb]
		\centering
		\subfigure[]{\label{fig:degen_multiply_1}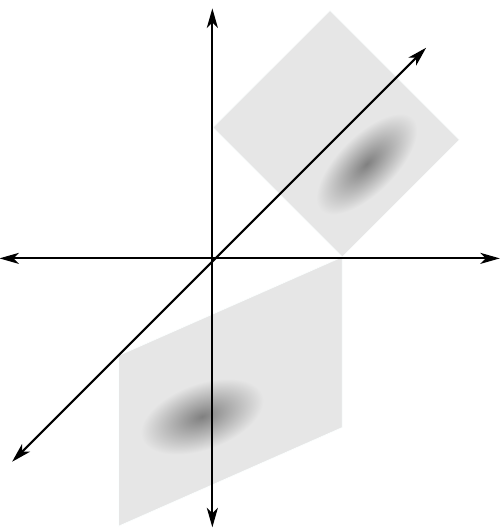}
		\hspace{0.05\linewidth}
		\subfigure[]{\label{fig:degen_multiply_2}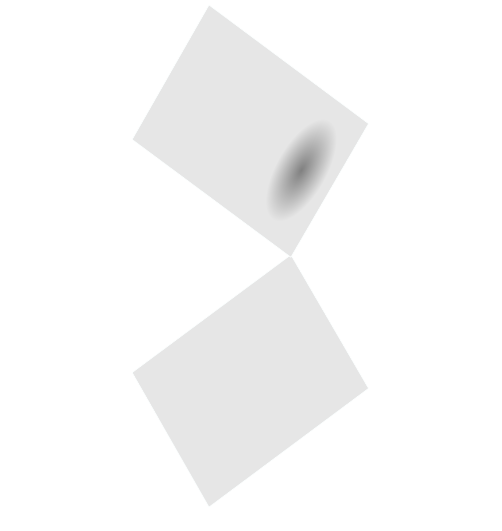}
		\caption{\revision{The multiplication of} (a) two degenerate factors in $\mathbb{R}^3$ subject to the linear constraints \revision{$x+y=a$ and $y+z=a$,} respectively. The intersection of the two affine planes is the line that goes through the points \revision{$(0,a,0)$ and $(a,0,a)$} -- along the vector $\mathbf{u}=\mathbf{q}'$. The vectors $\mathbf{r}_1$ (which is also perpendicular to the $z$-axis) and $\mathbf{r}_2$ (which is also perpendicular to the $x$-axis) are perpendicular to the respective planes. These three vectors are anchored at the point on the intersection with the minimum distance to the origin in $\mathbb{R}^3$. (b) From a view where this anchor point aligns with the origin, we see the true length of the vector $\mathbf{u}$. Each reduced factor along $\mathbf{u}$ is indicated by a dashed 1-D curve, where the product of these is equal to the product of the two degenerate factors along the line of intersection.}
		\label{fig:degen_multiply}
	\end{figure}

	The first step (in line 1 of Algorithm \ref{alg:multiply}) is to find an orthonormal basis $V$ for the column space of the projection of $R_2$ onto $N(R_1^T)=C(Q_1^{})$. This represents the additional degeneracy due to the second factor and, together with $R_1$, forms the orthonormal basis (in line 2) for the resulting degenerate space in $\mathbb{R}^n$. The offset $\mathbf{b}$ (in line 3) is determined by the intersection of the affine spaces of the two degenerate factors and this forms the second component of the offset $\mathbf{c}'$ (in line 4). In the edge case where $C(R_1)\cap C(R_2)\neq\{\mathbf{0}\}$, the matrix $R_2^TV$ will not be square and consequently its pseudo-inverse $(V_{}^TR_2^{}R_2^TV)^{-1}V_{}^TR_2^{}$ should be used instead\footnote{This will only be necessary when the same prior degeneracy exists in both factors and simplifies to $\left(R_2^TV\right)^{-1}$ otherwise.}. The orthonormal basis $U$ (in line 5) is then used to determine the resulting full-rank precision matrix, which is in turn diagonalised using the SVD (in line 6). As was the case in Algorithm \ref{alg:marginal}, the corresponding orthogonal matrix $Z$ is used (in line 7) to determine the basis for the support in $\mathbb{R}^n$. Finally, the vector $\mathbf{h}'$ and normalisation constant $g'$ are also calculated (in lines 8 and 9, respectively). In the case where $R_2^TV$ is not square, the determinant $|V_{}^TR_2^{}R_2^TV|$ should be used instead of $|R_2^TV|^2$. Figure \ref{fig:degen_multiply} illustrates various aspects of Algorithm \ref{alg:multiply} for a simple example.

	\subsection{Division} \label{sec:division}
	
	The quotient of two degenerate factors over the same scope $\mathbf{x}$ is given by
	\begin{linenomath*}\begin{equation} \label{eq:degenerate_divide}
	\frac{\mathcal{D}(\mathbf{x};Q_1,R_1,\Lambda_1,\mathbf{h}_1,\mathbf{c}_1,g_1)}{\mathcal{D}(\mathbf{x};Q_2,R_2,\Lambda_2,\mathbf{h}_2,\mathbf{c}_2,g_2)}=\mathcal{D}(\mathbf{x};Q',R',\Lambda',\mathbf{h}',\mathbf{c}',g'),
	\end{equation}\end{linenomath*}
	where the parameters $Q'$, $R'$, $\Lambda'$, $\mathbf{h}'$, $\mathbf{c}'$ and $g'$ are determined according to Algorithm \ref{alg:divide}. Due to the usual context of the division operation as part of the belief update algorithm \cite{lauritzen1988local} (where the numerator is a product of the denominator and other factors), we assume that a trivial denominator implies a trivial numerator\footnote{In addition, the results in Algorithm \ref{alg:divide} are only valid if $\Lambda'$ is non-negative as required by our definition of degenerate factors in \eqref{eq:degen_def}.}. \revision{Mathematically, this means} that $C(R_2)\subseteq C(R_1)$ and that $\mathbf{c}_2^{}=R_2^TR_1^{}\mathbf{c}_1^{}$. By choosing the affine space of the quotient \revision{ so that it is (a) a} subspace of that of the numerator \revision{and (b)} orthogonal to that of the denominator, we consequently reduce the denominator to the affine space of the numerator before dividing the resulting canonical factors as in \eqref{eq:canonical_divide}. Except for the different computation of $R'$ (in line 1 of Algorithm \ref{alg:divide}) and $Q'$ (in line 4), Algorithm \ref{alg:divide} follows a similar procedure to Algorithm \ref{alg:multiply}.
	
	\begin{algorithm}[htb]
		\caption{\texttt{Divide}} \label{alg:divide}
		\begin{algorithmic}[1]
			\Require $\theta_1$, $\theta_2$ such that $\phi_1(\mathbf{x})=\mathcal{D}(\mathbf{x};\theta_1)$ and $\phi_2(\mathbf{x})=\mathcal{D}(\mathbf{x};\theta_2)$
			\Ensure $\theta'$ such that $\phi_1(\mathbf{x})/\phi_2(\mathbf{x})=\mathcal{D}(\mathbf{x};\theta')$
			\State $R^\prime_{} \gets \texttt{Complement}\left(\begin{bmatrix}
			Q_1 & R_2
			\end{bmatrix}\right)$
			\State $\mathbf{c}^\prime \gets R^{\prime T}R_1\mathbf{c}_1$
			\State $Z,\Lambda_+,\_\gets\texttt{SVD}(\Lambda_1^{}-Q_1^TQ_2^{}\Lambda_2^{}Q_2^TQ_1^{})$
			\State $Q^\prime_{}\gets \begin{bmatrix}
			Q_1^{}Z & R_2^{}
			\end{bmatrix}$
			\State $\Lambda^\prime\gets\begin{bmatrix}
			\Lambda_+ & \textit{0} \\
			\textit{0} & \textit{0}
			\end{bmatrix}$
			\State $\mathbf{h}^\prime_{}\gets Q'^T_{}(Q_1\mathbf{h}_1^{}-Q_2^{}(\mathbf{h}_2^{}-\Lambda_2^{}Q_2^TR_1^{}\mathbf{c}_1^{}))$
			\State $g^\prime_{}\gets g_1^{}-g_2^{}-\left(\mathbf{h}_2-\frac{1}{2}\Lambda_2^{}Q_2^TR_1^{}\mathbf{c}_1^{} \right)^TQ_2^T R_1^{}\mathbf{c}_1^{}$
		\end{algorithmic}
	\end{algorithm}
	
	\subsection{Reduction}
	
	The partitioned degenerate factor in \eqref{eq:jointDegen} can also be reduced according to available evidence. Setting $\mathbf{y}=\mathbf{y}_0$ results in
	\begin{linenomath*}\begin{equation} \label{eq:cond_factor}
	\phi(\mathbf{x},\mathbf{y}_0)=\mathcal{D}\left(\mathbf{x};Q',R',\Lambda',\mathbf{h}',\mathbf{c}',g'\right),
	\end{equation}\end{linenomath*}
	where the parameters $Q'$, $R'$, $\Lambda'$, $\mathbf{h}'$, $\mathbf{c}'$ and $g'$ are determined according to Algorithm \ref{alg:reduce}. Since evidence can be regarded as imposing an additional linear constraint in the original space, we treat reduction as a special case of multiplication. Consequently, since the support of this product will be independent of the variables in $\mathbf{y}$, the lower-dimensional reduced factor is readily obtained through partitioning.
	
	\begin{algorithm}[htb]
		\caption{\texttt{Reduce}} \label{alg:reduce}
		\begin{algorithmic}[1]
			\Require $\theta$, $\mathbf{y}_0$ such that $\phi(\mathbf{x},\mathbf{y})=\mathcal{D}\left(\begin{bmatrix}
			\mathbf{x} \\
			\mathbf{y}
			\end{bmatrix};\theta\right)$
			\Ensure $\theta'$ such that $\phi(\mathbf{x},\mathbf{y}_0)=\mathcal{D}(\mathbf{x};\theta')$
			\State $R'\gets\texttt{Columnspace}\left(R_{\mathbf{x}}\right)$
			\State $\mathbf{c}'\gets \left(R_{\mathbf{x}}^TR'\right)^{-1}\left(\mathbf{c}-R_{\mathbf{y}}^T\mathbf{y}_0^{}\right)$
			\State $U_{\mathbf{x}}\gets \texttt{Complement}\left(R'\right)$
			\State $Z,\Lambda^\prime,\_\gets\texttt{SVD}(U_{\mathbf{x}}^TQ_{\mathbf{x}}^{}\Lambda Q_{\mathbf{x}}^TU_{\mathbf{x}}^{})$
			\State $Q^\prime_{}\gets U_{\mathbf{x}}^{}Z$
			\State $\mathbf{h}^\prime_{}\gets Q'^TQ_{\mathbf{x}}^{}\left(\mathbf{h}-\Lambda Q_{}^T\begin{bmatrix}
			R'\mathbf{c}' \\
			\mathbf{y}_0
			\end{bmatrix}\right)$
			\State $g^\prime_{}\gets g+\left(\mathbf{h}-\frac{1}{2}\Lambda Q_{}^T\begin{bmatrix}
			R'\mathbf{c}' \\
			\mathbf{y}_0
			\end{bmatrix}\right)^T\,Q_{}^T\begin{bmatrix}
			R'\mathbf{c}' \\
			\mathbf{y}_0
			\end{bmatrix}-\frac{1}{2}\log\left|R_{\mathbf{x}}^TR'\right|^2$
		\end{algorithmic}
	\end{algorithm}

\begin{figure}[htb]
		\centering
		\subfigure[]{\label{fig:degen_evidence_1}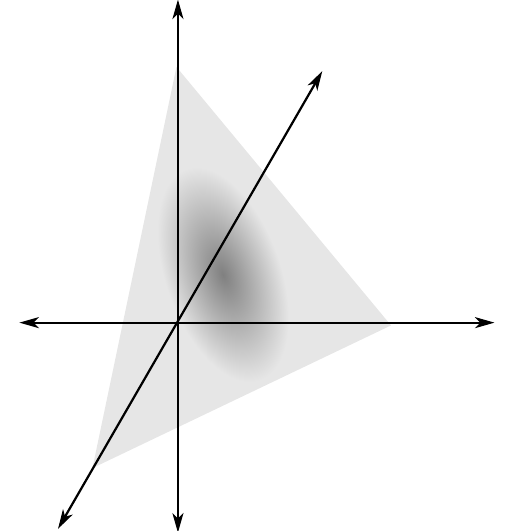}
		\hspace{0.05\linewidth}
		\subfigure[]{\label{fig:degen_evidence_2}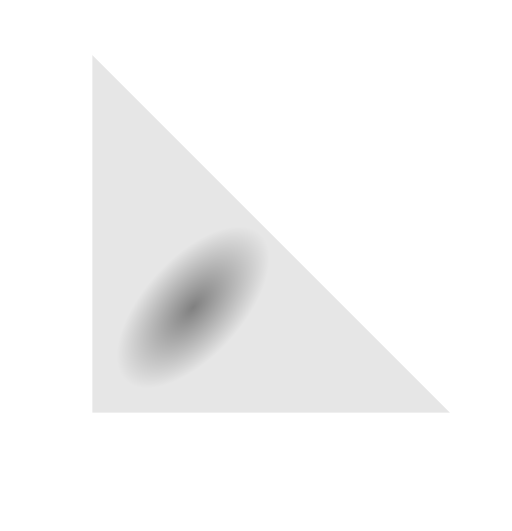}
		\caption{\revision{The reduction of} (a) a degenerate factor in $\mathbb{R}^3$ subject to the linear constraint \revision{$x_1+x_2+y=a$.} The evidence $y=y_0$ can be regarded as a horizontal plane and its intersection with the affine plane is indicated by the dashed line (along the vector $\mathbf{u}$). (b) From a view perpendicular to the $x_1$-$x_2$ plane, we see the orthogonal 2-D components $\mathbf{r}_{\mathbf{x}}$ and $\mathbf{u}_{\mathbf{x}}=\mathbf{q}'$ (where the latter is a unit vector). The reduced factor over $\{x_1,x_2\}$ then has uncertainty along $\mathbf{q}'$ as indicated by the dashed 1-D curve and is equal to zero away from the dashed line (in the direction of $\mathbf{r}'$ with corresponding offset $c'$).}
		\label{fig:degen_evidence}
	\end{figure}

	The first step (in line 1 of Algorithm \ref{alg:reduce}) is to find an orthonormal basis for the resulting degenerate space in $\mathbb{R}^{n'}$, where $n'$ is the dimension of the vector $\mathbf{x}$. The corresponding offset $\mathbf{c}'$ (in line 2) is determined by the intersection of the original affine space and the additional constraint $\mathbf{y}=\mathbf{y}_0$. In the edge case where the block matrix $R_{\mathbf{x}}$ does not have full column rank, the matrix $R_{\mathbf{x}}^TR'$ will not be square and consequently its pseudo-inverse $(R_{}'^TR_{\mathbf{x}}^{}R_{\mathbf{x}}^TR')^{-1}R_{}'^TR_{\mathbf{x}}^{}$ should be used instead\footnote{This will only be necessary when a prior degeneracy is again included in the evidence and simplifies to $\left(R_{\mathbf{x}}^TR'\right)^{-1}$ otherwise.}. The orthonormal basis $U_{\mathbf{x}}$ (in line 3) is then used to determine the resulting full-rank precision matrix, which is in turn diagonalised using the SVD (in line 4). As was the case in Algorithms \ref{alg:marginal}, \ref{alg:multiply} and \ref{alg:divide}, the corresponding orthogonal matrix $Z$ is once again used (in line 5) to determine the basis for the support in $\mathbb{R}^{n'}$. Finally, the vector $\mathbf{h}'$ and normalisation constant $g'$ are also calculated (in lines 6 and 7, respectively). In the case where $R_{\mathbf{x}}^TR'$ is not square, the determinant $|R_{}'^TR_{\mathbf{x}}^{}R_{\mathbf{x}}^TR'|$ should be used instead of $|R_{\mathbf{x}}^TR'|^2$. Figure \ref{fig:degen_evidence} illustrates various aspects of Algorithm \ref{alg:reduce} for a simple example.

	\subsection{Computational complexity}
	
	Due to our use of the SVD, the asymptotic time complexity of the operations in Algorithms \ref{alg:marginal} to \ref{alg:reduce} is $\mathcal{O}(n^3)$. This is similar to that for Raphael's \cite{raphael2003bayesian} representation. In comparison, the multiplication, division and reduction operations for Koller and Friedman's \cite{koller2009} (non-degenerate) canonical factors require only $\mathcal{O}(n^2)$. However, due to the necessary matrix inversion\footnote{Matrix inversion can be done in $\mathcal{O}(n^{2.807})$ \cite{strassen1969gaussian} or even $\mathcal{O}(n^{2.376})$ \cite{coppersmith1987}, although the large constant coefficient in the latter renders it impractical for typical matrices.}, their marginalisation operation also requires $\mathcal{O}(n^3)$ and therefore often dominates the computational time in practice. The expressiveness of degenerate factors therefore comes at little additional computational cost.
	
\revision{\section{Additional operations necessary for inference}} \label{sec:add_operations}

	In addition to the four operations on degenerate factors discussed in Section \ref{sec:operations}, we present further \revision{operations} that are necessary for performing inference on Bayesian networks. \revision{A Bayesian network is a type of probabilistic graphical model where each of the variable nodes are associated with a conditional distribution and the (absence of) edges indicate conditional (in)dependencies among the random variables \cite{pearl1986fusion}.} These \revision{operations} include correctly manipulating degenerate factors with different scopes, modelling stochastic systems by representing conditional densities (associated with both linear and nonlinear dependencies) and checking for convergence of message-passing algorithms. The detailed derivations are once again included in \ref{sec:proofs} where applicable.
	
	\subsection{Extending and rearranging factor scopes}
	
	Performing inference on graphical models often amounts to multiplying (or dividing) two factors with different scopes. This is true even for the simple case of computing the joint density $p(\mathbf{x},\mathbf{y})=p(\mathbf{y}|\mathbf{x})\,p(\mathbf{x})$, where the scope of the second factor is a subset of that of the first. In such cases, we need to extend the scope of each factor in the product to be the union of all the scopes.	For this purpose, we can extend the scope of a given degenerate factor with scope $\mathbf{x}$	to include the vector $\mathbf{y}$, such that
	\begin{linenomath*}\begin{equation} \label{eq:extension}
	\mathcal{D}\left(\begin{bmatrix}
	\mathbf{x} \\
	\mathbf{y}
	\end{bmatrix};Q',R',\Lambda',\mathbf{h}',\mathbf{c},g\right)=\mathcal{D}\left(\mathbf{x};Q,R,\Lambda,\mathbf{h},\mathbf{c},g\right).
	\end{equation}\end{linenomath*}
	A choice for the parameters that conforms to the definition of the degenerate factor in \eqref{eq:degen_def} is
	\begin{linenomath*}\begin{equation} \label{eq:extension_result}
	Q'=\begin{bmatrix}
	Q & \textit{0} \\
	\textit{0} & I
	\end{bmatrix},\enspace R'=\begin{bmatrix}
	R \\
	\textit{0}
	\end{bmatrix},\enspace \Lambda'=\begin{bmatrix}
	\Lambda & \textit{0} \\
	\textit{0} & \textit{0}
	\end{bmatrix}\enspace\text{and}\enspace\mathbf{h}'=\begin{bmatrix}
	\mathbf{h} \\
	\mathbf{0}
	\end{bmatrix}.
	\end{equation}\end{linenomath*}
	Note that $\Lambda'$ is diagonal and that $C\left(Q'\right)=C\left(R'\right)^\perp$ as required.
	
	Another operation that is necessary for a message-passing implementation is rearranging the scope of a given factor. For instance, consider a degenerate Gaussian factor with scope $\{\mathbf{x},\mathbf{y}\}$ where the matrices $Q$ and $R$ are partitioned accordingly. We can determine another factor with scope $\{\mathbf{y},\mathbf{x}\}$, such that
	\begin{linenomath*}\begin{equation}
	\mathcal{D}\left(\begin{bmatrix}
	\mathbf{y} \\
	\mathbf{x}
	\end{bmatrix};Q',R',\Lambda,\mathbf{h},\mathbf{c},g\right)=\mathcal{D}\left(\begin{bmatrix}
	\mathbf{x} \\
	\mathbf{y}
	\end{bmatrix};\begin{bmatrix}
	Q_{\mathbf{x}} \\
	Q_{\mathbf{y}}
	\end{bmatrix},\begin{bmatrix}
	R_{\mathbf{x}} \\
	R_{\mathbf{y}}
	\end{bmatrix},\Lambda,\mathbf{h},\mathbf{c},g\right),
	\end{equation}\end{linenomath*}
	by simply rearranging the rows of the two matrices
	\begin{linenomath*}\begin{equation}
	Q'=\begin{bmatrix}
	Q_{\mathbf{y}} \\
	Q_{\mathbf{x}}
	\end{bmatrix}\quad\text{and}\quad R'=\begin{bmatrix}
	R_{\mathbf{y}} \\
	R_{\mathbf{x}}
	\end{bmatrix}.
	\end{equation}\end{linenomath*}
	This operation is necessary to insure that, prior to applying any statistical operations, the scopes of multiple factors not only have the same dimension but are also aligned.
	
	\revision{\subsection{Representing conditional density functions}}
	
	To model a stochastic system using a Bayesian network, it is necessary to represent conditional density functions -- in our case as parametrised degenerate factors. \revision{We start with the special case of linear transformations, where the expression for the conditional density is exact. However, since it is usually also necessary to represent more general nonlinear relationships, we propose a method for approximating these.}

	\revision{\subsubsection{Linear dependencies}} \label{sec:represent}
	
	Given the affine transformation
	\begin{linenomath*}\begin{equation} \label{eq:affine_noise}
	\mathbf{y}=A\mathbf{x}+\mathbf{b}+\mathbf{w}
	\end{equation}\end{linenomath*}
	subject to (possibly degenerate) independent noise
	\begin{linenomath*}\begin{equation} \label{eq:p_w}
	\mathbf{w}\sim\mathcal{D}(\mathbf{w};Q,R,\Lambda,\mathbf{h},\mathbf{c},g),
	\end{equation}\end{linenomath*}
	we can represent the conditional density
	\begin{linenomath*}\begin{equation}
	p(\mathbf{y}|\mathbf{x})=\mathcal{D}\left(\begin{bmatrix}
	\mathbf{x} \\
	\mathbf{y}
	\end{bmatrix};Q',R',\Lambda',\mathbf{h}',\mathbf{c}',g'\right),
	\end{equation}\end{linenomath*}
	where the parameters $Q'$, $R'$, $\Lambda'$, $\mathbf{h}'$, $\mathbf{c}'$ and $g'$ are determined according to Algorithm \ref{alg:conditional_factor}. By treating the vector $\mathbf{x}$ as constant, we use an affine transformation of the random vector $\mathbf{w}$ (according to Algorithm \ref{alg:affine_transform}) to express $p(\mathbf{y}|\mathbf{x})$ as a factor over $\mathbf{y}$. We then rewrite this factor as a parametrised degenerate factor with scope $\{\mathbf{x},\mathbf{y}\}$.
	
	\begin{algorithm}[htb]
		\caption{\texttt{RepresentConditional}} \label{alg:conditional_factor}
		\begin{algorithmic}[1]
			\Require $\theta$, $A$, $\mathbf{b}$ such that $p(\mathbf{w})=\mathcal{D}(\mathbf{w};\theta)$ and $\mathbf{y}=A\mathbf{x}+\mathbf{b}+\mathbf{w}$
			\Ensure $\theta'$ such that $p(\mathbf{y}|\mathbf{x})=\mathcal{D}\left(\begin{bmatrix}
			\mathbf{x} \\
			\mathbf{y}
			\end{bmatrix};\theta'\right)$
			\State $F \gets \begin{bmatrix}
			-A & I
			\end{bmatrix}$
			\State $R^\prime\gets \texttt{Columnspace}\left(F^TR\right)$
			\State $Z\gets R^{\prime T}F^TR\left(R^T(I+AA^T)R\right)^{-1}$
			\State $\mathbf{c}^\prime \gets Z(\mathbf{c}+R^T\mathbf{b})$
			\State $P\gets\left(I-R'R'^T\right)$
			\State $Q_+^{},\Lambda_+^{},\_\gets\texttt{CompactSVD}\left(PF^TQ\Lambda Q^TFP\right)$
			\State $Q_\infty\gets\texttt{Complement}\left(\begin{bmatrix}
			Q_+ & R'
			\end{bmatrix}\right)$
			\State $Q^\prime\gets \begin{bmatrix}
			Q_+ & Q_\infty
			\end{bmatrix}$
			\State $\Lambda^\prime\gets\begin{bmatrix}
			\Lambda_+ & \textit{0} \\
			\textit{0} & \textit{0}
			\end{bmatrix}$
			\State $\mathbf{h}^\prime_{}\gets Q^{\prime T}F^TQ(\mathbf{h}+\Lambda Q^T(\mathbf{b}-FR'\mathbf{c}'))$
			\State $g^\prime\gets g-(\mathbf{h}+\frac{1}{2}\Lambda Q^T\mathbf{b})^T \,Q^T\mathbf{b}+(\mathbf{h}+\Lambda Q^T(\mathbf{b}-\frac{1}{2}FR'\mathbf{c}'))^TQ^TFR'\mathbf{c}'+\frac{1}{2}\log|Z|^2$
		\end{algorithmic}
	\end{algorithm}
	
\begin{figure}[htb]
		\centering
		\subfigure[]{\label{fig:degen_conditional_1}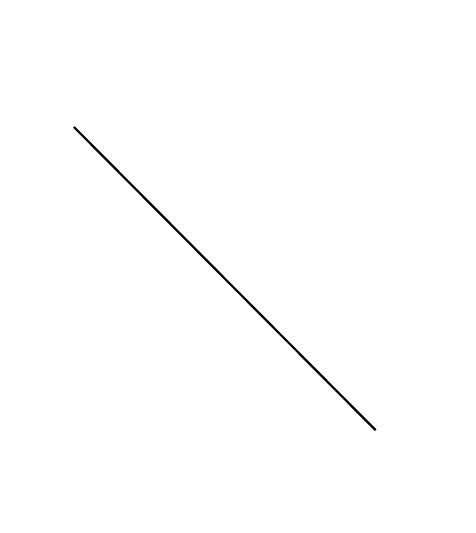}
		\hspace{0.05\linewidth}
		\subfigure[]{\label{fig:degen_conditional_3}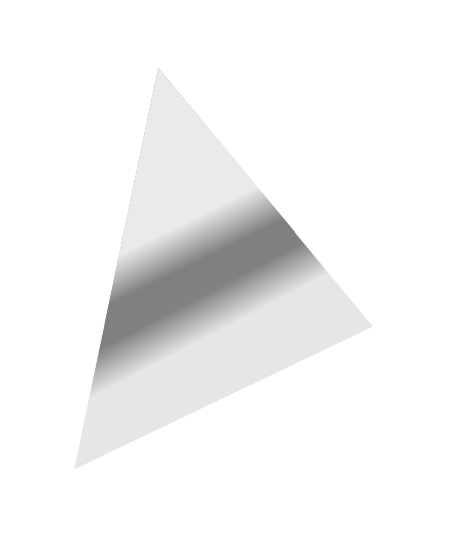}
		\caption{(a) A visualisation of a degenerate factor in $\mathbb{R}^2$ subject to the linear \revision{noise} constraint \revision{$w_1+w_2=a$.} The dashed 1-D density function indicates the uncertainty along the constraint (and in the direction of the vector $\mathbf{q}$). Away from this line (in the direction of $\mathbf{r}$) the factor is equal to zero. (b) The constructed degenerate factor $p(y_1,y_2|x)$ in $\mathbb{R}^3$, where $y_1=w_1+x$ and $y_2=w_2$. \revision{This} factor is consequently subject to the linear constraint \revision{$y_1+y_2-x=a$} and the vector $\mathbf{r}'$ is perpendicular to the resulting affine plane. In contrast, the vectors $\mathbf{q}_{+}$ and $\mathbf{q}_\infty$ lie in the plane, where the factor is invariant to any change along the latter.}
		\label{fig:degen_conditional}
	\end{figure}

	The first step (in line 2 of Algorithm \ref{alg:conditional_factor}) is to use the auxiliary matrix $F$ (in line 1) to find an orthonormal basis $R'$ for the resulting degenerate space in $\mathbb{R}^{n_{\mathbf{x}}+n_{\mathbf{y}}}$, where the matrix $A$ has shape $n_{\mathbf{y}}\times n_{\mathbf{x}}$. The corresponding $k$-dimensional offset $\mathbf{c}'$ (in line 4) is then determined using the auxiliary matrix $Z$ (in line 3). To ensure that $C(Q')\perp C(R')$, the projection matrix (in line 5) is applied before the compact SVD (in line 6). This yields the basis $Q_+$ corresponding to the $n_{\mathbf{y}}-k$ non-trivial singular values $\Lambda_+$, where we use the same subscripts as Raphael \cite{raphael2003bayesian}. The remainder of the orthogonal decomposition of $\mathbb{R}^{n_{\mathbf{x}}+n_{\mathbf{y}}}$ corresponding to the $n_{\mathbf{x}}$ trivial singular values is then determined (in line 7) and used to augment $Q'$ and $\Lambda'$ (in lines 8 and 9, respectively). Finally, the vector $\mathbf{h}'$ and normalisation constant $g'$ are also calculated (in lines 10 and 11, respectively). Figure \ref{fig:degen_conditional} illustrates various aspects of Algorithm \ref{alg:conditional_factor} for a simple example.
	
	\revision{\subsubsection{Nonlinear dependencies}} \label{sec:equivalent}

	Now consider the more general nonlinear transformation
	\begin{linenomath*}\begin{equation} \label{eq:non_linear}
	\mathbf{y}=\mathbf{f}(\mathbf{x},\mathbf{w})
	\end{equation}\end{linenomath*}
	and suppose that we once again need to represent the conditional density
	\begin{linenomath*}\begin{equation} \label{eq:non_linear_cpd}
	p(\mathbf{y}|\mathbf{x})\approx\mathcal{D}\left(\begin{bmatrix}
	\mathbf{x} \\
	\mathbf{y}
	\end{bmatrix};Q',R',\Lambda',\mathbf{h}',\mathbf{c}',g'\right).
	\end{equation}\end{linenomath*}
	In the nonlinear case, \revision{however,} we need to resort to approximation through linearisation, \revision{where two of the most popular techniques are Taylor series expansion and the unscented transform \cite{julier1997}. While the former uses partial derivatives of the given non-linear transform to linearise around a single point, multiple deterministic samples (known as sigma points) are individually propagated through the non-linear transform in the latter. In the non-degenerate case, Thrun et al. \cite{thrun2005probabilistic} propose $2n+1$ sigma points determined using the Cholesky decomposition. Although each has its advantages, the performance of the unscented transform has been proven to be at least as good as that of the Taylor approximation on average \cite{wan2000}.
		
	When using the Taylor series expansion, the approach is very similar to the non-degenerate case, where Jacobian matrices can be used to obtain an affine approximation of \eqref{eq:non_linear} before applying the results in Algorithm \ref{alg:conditional_factor}. However, to} use the unscented transform and sigma points drawn from the prior distribution\footnote{For statistically independent random vectors $\mathbf{x}$ and $\mathbf{w}$, this joint factor can easily be computed using the extension operation in \eqref{eq:extension_result} on each marginal factor followed by (a special case of) the multiplication operation in Algorithm \ref{alg:multiply}.}
	\begin{linenomath*}\begin{equation} \label{eq:p_xw}
	p(\mathbf{x},\mathbf{w})=p(\mathbf{x})\,p(\mathbf{w})=\mathcal{D}\left(\begin{bmatrix}
	\mathbf{x} \\
	\mathbf{w}
	\end{bmatrix};Q,R,\Lambda,\mathbf{h},\mathbf{c},g\right)
	\end{equation}\end{linenomath*}
	\revision{requires a more subtle approach. The idea is to use samples from the canonical component of the prior in \eqref{eq:p_xw}} to obtain an equivalent affine transformation $\mathbf{x}\mapsto\mathbf{y}$, given by
	\begin{linenomath*}\begin{equation} \label{eq:equiv_linear}
	\mathbf{y}\approx\widetilde{A}\mathbf{x}+\widetilde{\mathbf{b}}+\widetilde{\mathbf{w}}
	\end{equation}\end{linenomath*}
	and where the affine transform noise is
	\begin{linenomath*}\begin{equation} \label{eq:noise_linear}
	\widetilde{\mathbf{w}}\sim\mathcal{D}(\widetilde{\mathbf{w}};\widetilde{Q},\widetilde{R},\widetilde{\Lambda},\widetilde{\mathbf{h}},\widetilde{\mathbf{c}},\widetilde{g}).
	\end{equation}\end{linenomath*}
	This is achieved in Algorithm \ref{alg:equiv_transform} through moment-matching of the joint density $p(\mathbf{x},\mathbf{y})$ resulting from each transformation in \eqref{eq:non_linear} and \eqref{eq:equiv_linear}. Once the parameters in \eqref{eq:equiv_linear} and \eqref{eq:noise_linear} have been calculated, we can approximate the conditional density in \eqref{eq:non_linear_cpd} using Algorithm \ref{alg:conditional_factor}.
	
	\begin{algorithm}[htb]
		\caption{\texttt{EquivalentTransformation}} \label{alg:equiv_transform}
		\begin{algorithmic}[1]
			\Require $\theta$, $\mathbf{f}(\cdot)$ such that $p(\mathbf{x},\mathbf{w})=\mathcal{D}\left(\begin{bmatrix}
			\mathbf{x} \\
			\mathbf{w}
			\end{bmatrix};\theta\right)$ and $\mathbf{y}=\mathbf{f}(\mathbf{x},\mathbf{w})$
			\Ensure $\widetilde{\theta}$, $\widetilde{A}$, $\widetilde{\mathbf{b}}$ such that $p(\mathbf{\widetilde{w}})=\mathcal{D}(\mathbf{\widetilde{w}};\widetilde{\theta})$ and $\mathbf{y}\approx \widetilde{A}\mathbf{x}+\widetilde{\mathbf{b}}+\widetilde{\mathbf{w}}$
			\State $\begin{bmatrix}
			\mathcal{X} \\
			\mathcal{W}
			\end{bmatrix}\gets\begin{bmatrix}
			\mathbf{0}, & \gamma Q\sqrt{\Lambda^{-1}}, & {}-\gamma Q\sqrt{\Lambda^{-1}}
			\end{bmatrix}+(Q\Lambda^{-1}\mathbf{h}+R\mathbf{c})\,\mathbf{1}^T$
			\For{$i=0,2(n-k)$}
			\State $\mathbf{y}^{[i]}\gets\mathbf{f}\left(\mathbf{x}^{[i]},\mathbf{w}^{[i]}\right)$
			\EndFor
			\State $\boldsymbol{\mu}\gets\sum_{i=0}^{2(n-k)}w_m^{[i]}\begin{bmatrix}
			\mathbf{x}^{[i]} \\
			\mathbf{y}^{[i]}
			\end{bmatrix}$
			\State $\Sigma\gets\sum_{i=0}^{2(n-k)}w_c^{[i]}\,\left(\begin{bmatrix}
			\mathbf{x}^{[i]} \\
			\mathbf{y}^{[i]}
			\end{bmatrix}-\boldsymbol{\mu}'\right)\left(\begin{bmatrix}
			\mathbf{x}^{[i]} \\
			\mathbf{y}^{[i]}
			\end{bmatrix}-\boldsymbol{\mu}'\right)^T$
			\State $\widetilde{A}\gets\Sigma_{\mathbf{xy}}^T\Sigma_{\mathbf{xx}}^+$
			\State $\widetilde{\mathbf{b}}\gets\boldsymbol{\mu}_{\mathbf{y}}-\widetilde{A}\boldsymbol{\mu}_{\mathbf{x}}$
			\State $\widetilde{Q},\widetilde{\Sigma},\_\gets\texttt{CompactSVD}\left(\Sigma_{\mathbf{yy}}-\widetilde{A}\Sigma_{\mathbf{xy}}\right)$
			\State $\widetilde{\Lambda}\gets \widetilde{\Sigma}^{-1}$
			\State $\widetilde{R}\gets \texttt{Complement}(\widetilde{Q})$
			\State $\widetilde{\mathbf{h}}\gets\mathbf{0}$, $\widetilde{\mathbf{c}}\gets\mathbf{0}$
			\State $\widetilde{g}\gets -\frac{1}{2}\log\left|2\pi\widetilde{\Lambda}^{-1}\right|$
		\end{algorithmic}
	\end{algorithm}

	The first step (in line 1 of Algorithm \ref{alg:equiv_transform}) is to draw $2(n-k)$ sigma points from the prior distribution $p(\mathbf{x},\mathbf{w})$, where $k$ is the degree of degeneracy and $\gamma$ is a scaling parameter. Each sigma point is then propagated through the nonlinear transform (in line 3). As outlined by Thrun et al. \cite{thrun2005probabilistic}, the mean (in line 5) and covariance (in line 6) of the joint density $p(\mathbf{x},\mathbf{y})$ are approximated using the sigma points and corresponding weights. By partitioning the moments according to $\mathbf{x}$ and $\mathbf{y}$ and making use of the pseudo-inverse, the matrix $\widetilde{A}$ is determined (in line 7). This value of $\widetilde{A}$ is then used to determine that of $\widetilde{\mathbf{b}}$ (in line 8) as well as in the compact SVD of the noise covariance (in line 9). The quantities $\widetilde{\Lambda}$ (in line 10) and $\widetilde{R}$ (in line 11) are determined as in Algorithm \ref{alg:affine_transform}. Without loss of generality, the noise $\widetilde{\mathbf{w}}$ can be chosen as zero-mean, which implies that $\widetilde{\mathbf{h}}=\mathbf{0}$ and $\widetilde{\mathbf{c}}=\mathbf{0}$ (in line 12). Finally, the normalisation constant (in line 13) is calculated according to \eqref{eq:g_norm}.
	
	\subsection{Kullback-Leibler divergence} \label{sec:KL}
	
	\revision{According to Koller and Friedman \cite{koller2009},} inference on \revision{probabilistic graphical models} containing loops is approximate. In addition, even for tree-structured graphs, approximations could be introduced due to linearisation. For message-passing algorithms, this implies that messages typically need to be computed iteratively until convergence. A popular method to check for convergence of such an algorithm is to use the Kullback-Leibler (KL) divergence
	\begin{linenomath*}\begin{equation} \label{eq:KL}
	D_\text{KL}(P||Q)\triangleq\int_{-\infty}^{\infty}p(\mathbf{x})\log\left(\frac{p(\mathbf{x})}{q(\mathbf{x})}\right)\,\text{d}\mathbf{x}=\mathbb{E}_{p(\mathbf{x})}\left[\log p(\mathbf{x})-\log q(\mathbf{x})\right].
	\end{equation}\end{linenomath*}
	This provides a measure of the relative entropy from density $Q$ to density $P$, and $D_\text{KL}(P||Q)=0$ only when $p(\mathbf{x})=q(\mathbf{x})$.
	
	In the case of two degenerate Gaussian densities
	\begin{linenomath*}\begin{equation} \label{eq:KL_densities}
	p(\mathbf{x}) = \mathcal{D}(\mathbf{x};Q_1,R_1,\Lambda_1,\mathbf{h}_1,\mathbf{c}_1,g_1)\quad\text{and}\quad q(\mathbf{x}) = \mathcal{D}(\mathbf{x};Q_2,R_2,\Lambda_2,\mathbf{h}_2,\mathbf{c}_2,g_2),
	\end{equation}\end{linenomath*}
	the KL divergence can be calculated according to
	\begin{linenomath*}\begin{align} \label{eq:KL_result}
	D_\text{KL}(P||Q)&=\frac{1}{2}\Big(\text{tr}\left(Q_2^{}\Lambda_2^{}Q_2^TQ_1^{}\Lambda_1^{-1}Q_1^T\right)+\mathbf{h}_1^T\Lambda_1^{-1}Q_1^TQ_2^{}\Lambda_2^{}Q_2^TQ_1^{}\Lambda_1^{-1}\mathbf{h}_1^{}+{} \nonumber \\
	&\quad\qquad\mathbf{h}_1^T\Lambda_1^{-1}\mathbf{h}_1^{}-n+k\Big)-\mathbf{h}_1^T\Lambda_1^{-1}Q_1^TQ_2^{}\mathbf{h}_2^{}+g_1^{}-g_2^{}.
	\end{align}\end{linenomath*}
	Note, however, that this result is only valid when the two degenerate densities have support on the same lower-dimensional manifold, i.e., if $C\left(R_1\right)=C\left(R_2\right)$ and $R_1\mathbf{c}_1=R_2\mathbf{c}_2$. Conversely, if these conditions are not satisfied, the KL divergence will instead be infinite. In the context of message-passing algorithms, however, this should not be the case near convergence.
	
	\revision{Together with the results from Section \ref{sec:operations}, this section provided the means to perform inference on Bayesian networks using degenerate Gaussian factors. This includes representing both linear and nonlinear models (as outlined in Algorithms \ref{alg:conditional_factor} and \ref{alg:equiv_transform}) as well as practical aspects such as aligning factor scopes and checking for convergence of message-passing algorithms. Note, however, that this does not limit the application of degenerate factors to Bayesian networks alone, where many of the results derived thus far can readily form part of more general inference processes. In the next section, we support our theoretical development by applying these results to a recursive state estimation problem.}
	
\section{An example: Recursive state estimation for mobile robots} \label{sec:experiments}

	To illustrate the advantages of performing inference with degenerate factors, we consider a representative example. More specifically, we demonstrate a scenario where it is essential to account for degeneracies \revision{(a) during the} modelling \revision{process as well as (b) when performing subsequent} computations. For this purpose, we first provide an overview of recursive state estimation -- specifically in the context of mobile robotics. In order to keep the discussion pertinent yet not overly complex, we choose to model the estimation problem with a \revision{Bayesian network \cite{pearl1986fusion}.} Since we perform inference on a nonlinear model using the belief propagation algorithm \cite{shenoy1986propagating}, all the operations outlined in Sections \ref{sec:operations} and \ref{sec:add_operations} (except for division) are relevant. We also comment on the shortcomings of existing approaches using appropriate qualitative and quantitative results.
	
	\subsection{Recursive state estimation} \label{sec:recursive}
	
\begin{figure}[htb]
		\centering
		\subfigure[]{\label{fig:bayesian_network}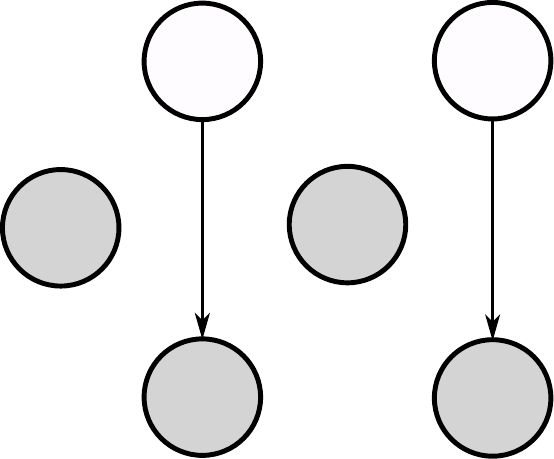}
		\hspace{0.03\linewidth}
		\subfigure[]{\label{fig:factor_graph}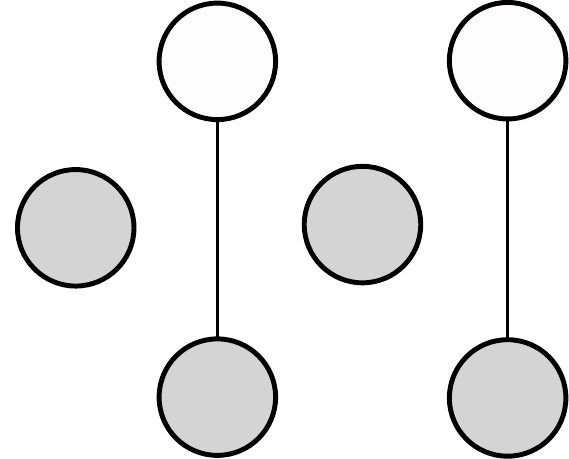}
		\caption{(a) A Bayesian network that models the recursive state estimation problem (as adapted from Thrun et al. \cite{thrun2005probabilistic}). Round nodes represent random variables (where a shaded node indicates that the variable is observed) and directed edges indicate causal dependencies. For each discrete time step (as indicated by the subscript $k$), the state $\mathbf{x}_k$ is latent while the control inputs $\mathbf{u}_k$ and measurement $\mathbf{z}_k$ are observed. (b) The corresponding factor graph. Dark squares indicate factors, where each factor is connected to all the random variables in its scope via undirected edges. We distinguish between motion factors $\psi_k$ and measurement factors $\rho_k$.}
		\label{fig:model}
	\end{figure}

	Recursive state estimation (or Bayesian filtering) refers to a setting where the posterior distribution over a time-dependent latent state must be inferred using noisy measurements thereof \revision{\cite{schweppe1968recursive}.} Common applications are found in mobile robotics \cite{thrun2005probabilistic}, where accurate control of a vehicle heavily relies on a good estimate of its state. \revision{For practical robotic systems,} the estimation usually occurs online and in real time. \revision{Since regularisation techniques are often used to improve numerical robustness in robotic estimation problems \cite{koshizen1999sensor,inoue2016extended}, this is an appropriate context for showcasing the advantages of degenerate Gaussian factors.
	
	} The standard Markovian formulation of the estimation problem consists of a \revision{motion and a measurement model. The} \emph{motion} model \revision{is} a discrete-time, nonlinear function
	\begin{linenomath*}\begin{equation} \label{eq:motion}
	\mathbf{x}_k=\mathbf{g}(\mathbf{x}_{k-1},\mathbf{u}_k,\mathbf{w}_k)
	\end{equation}\end{linenomath*}
	relating the current state $\mathbf{x}_k$ to the previous state $\mathbf{x}_{k-1}$, the control inputs $\mathbf{u}_k$ and the process noise $\mathbf{w}_k$. \revision{In turn, the} \emph{measurement} model \revision{is} a function
	\begin{linenomath*}\begin{equation} \label{eq:measurement}
	\mathbf{z}_k=\mathbf{h}(\mathbf{x}_k,\mathbf{v}_k)
	\end{equation}\end{linenomath*}
	relating the measurement $\mathbf{z}_k$ to the state $\mathbf{x}_{k}$ and the measurement noise $\mathbf{v}_k$. \revision{Additionally,} the prior distributions over the process noise $\mathbf{w}_k$ and measurement noise $\mathbf{v}_k$ are \revision{generally assumed to be} known. The aim of the estimator is then to calculate what Thrun et al. \cite{thrun2005probabilistic} call the \emph{belief} \revision{for every time step $k\in\{1,2,\dots K\}$. The belief is} the posterior distribution
	\begin{linenomath*}\begin{equation} \label{eq:belief}
	\text{bel}(\mathbf{x}_k)\triangleq p(\mathbf{x}_k|\mathbf{u}_{1:k},\mathbf{z}_{1:k})
	\end{equation}\end{linenomath*}
	over the state $\mathbf{x}_k$ given all the control inputs $\mathbf{u}_{1:k}$ and measurements $\mathbf{z}_{1:k}$ up to time $k$. Under the Markov assumption, the estimate in \eqref{eq:belief} can be calculated recursively. The Bayesian network and subsequent factor graph \cite{koller2009} representing this inference problem are shown in Figure \ref{fig:model}, where we distinguish between motion factors
	\begin{linenomath*}\begin{equation} \label{eq:motion_factor}
	\psi_k(\mathbf{x}_{k-1},\mathbf{x}_k,\mathbf{u}_k)\triangleq p(\mathbf{x}_k|\mathbf{x}_{k-1},\mathbf{u}_k)
	\end{equation}\end{linenomath*}
	and measurement factors
	\begin{linenomath*}\begin{equation} \label{eq:meas_factor}
	\rho_k(\mathbf{x}_k,\mathbf{z}_k)\triangleq p(\mathbf{z}_k|\mathbf{x}_k)
	\end{equation}\end{linenomath*}
	in the latter. Due to the nonlinear models in \eqref{eq:motion} and \eqref{eq:measurement}, it is generally necessary to approximate the factors in \eqref{eq:motion_factor} and \eqref{eq:meas_factor} using a linearisation technique such as the unscented transform.
	
\begin{figure}[htb]
	\centering
	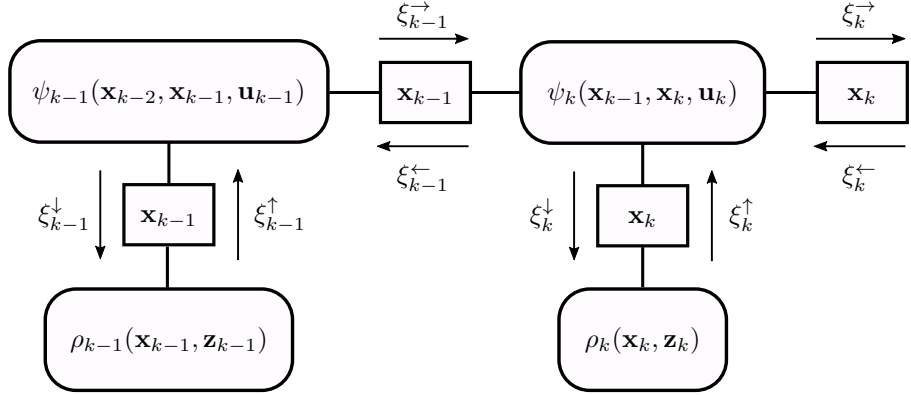
	\caption{A possible cluster graph corresponding to the model of the recursive state estimation problem in Figure \ref{fig:model}, where each factor has been placed in its own cluster. Each cluster potential is therefore equal to the appropriate motion or measurement factor. The sepsets are indicated by rectangular nodes and are connected to the clusters via undirected edges. Messages are indicated by arrows and are differentiated (by their labels) according to both their scopes and their directions.}
	\label{fig:cluster}
	\end{figure}

	To solve the estimation problem (i.e., to calculate the belief over every state $\mathbf{x}_k$), we subsequently construct the cluster graph \cite{koller2009} shown in Figure \ref{fig:cluster}. Note, however, that this choice of cluster graph is not unique. According to the belief propagation algorithm, the rightward, leftward, downward and upward messages are defined as
	\begin{linenomath*}\begin{align} \label{eq:messages}
	\xi_k^{\rightarrow}(\mathbf{x}_k)&=\int\psi_k(\mathbf{x}_{k-1},\mathbf{x}_k,\mathbf{u}_k)\,\xi_{k-1}^{\rightarrow}(\mathbf{x}_{k-1})\,\xi_k^{\uparrow}(\mathbf{x}_k)\,\text{d}\mathbf{x}_{k-1} \nonumber \\
	\xi_k^{\leftarrow}(\mathbf{x}_k)&=\int\psi_{k+1}(\mathbf{x}_{k},\mathbf{x}_{k+1},\mathbf{u}_{k+1})\,\xi_{k+1}^{\leftarrow}(\mathbf{x}_{k+1})\,\xi_{k+1}^{\uparrow}(\mathbf{x}_{k+1})\,\text{d}\mathbf{x}_{k+1} \nonumber \\
	\xi_k^{\downarrow}(\mathbf{x}_k)&=\int\psi_k(\mathbf{x}_{k-1},\mathbf{x}_k,\mathbf{u}_k)\,\xi_{k-1}^{\rightarrow}(\mathbf{x}_{k-1})\,\xi_k^{\leftarrow}(\mathbf{x}_k)\,\text{d}\mathbf{x}_{k-1}  \nonumber \\
	\xi_k^{\uparrow}(\mathbf{x}_k)&=\rho_k(\mathbf{x}_k,\mathbf{z}_k),
	\end{align}\end{linenomath*}
	respectively. The computations in the first three cases each makes use of two multiplication operations followed by a single marginalisation operation. In addition, all four computations make use of a reduction operation based on either the available control inputs $\mathbf{u}_k$ or the observed measurement $\mathbf{z}_k$. Once all of the messages have been computed (usually through iteration until convergence), the posterior distribution
	\begin{linenomath*}\begin{equation} \label{eq:belief_smoothing}
	p(\mathbf{x}_k|\mathbf{u}_{1:K},\mathbf{z}_{1:K})=\xi_k^{\rightarrow}(\mathbf{x}_k)\,\xi_k^{\leftarrow}(\mathbf{x}_k)
	\end{equation}\end{linenomath*}
	can be determined for each time step. Note the slight difference in conditioning between \eqref{eq:belief} (known as filtering) and \eqref{eq:belief_smoothing} (known as smoothing). Although the former inference problem can also be solved using \eqref{eq:messages} and \eqref{eq:belief_smoothing} by simply omitting all leftward messages $\xi_k^{\leftarrow}$, we will only consider the latter going forward as it is \revision{(a) usually more accurate and (b)} a more natural context for graphical models.

	Even though the computations in \eqref{eq:messages} and \eqref{eq:belief_smoothing} are valid for any factor representation, the advantages of Gaussian factors (for example the fact that they are closed under the required operations) make them a popular approximation in robotics applications. The most natural context that further warrants the use of \emph{degenerate} Gaussian factors is one where the models in \eqref{eq:motion} and \eqref{eq:measurement} include deterministic relationships between a subset of the variables. Note, however, that if a given degeneracy is present for all time steps, it could be sufficient to construct an equivalent, lower-dimensional model of the system without requiring degenerate factors. An appropriate example to illustrate the advantages of degenerate factors is therefore one with time-dependent models, where degeneracies arise inconsistently and unpredictably.

	\subsection{Cooperative \revision{transportation} robots} \label{sec:cooperative}
	
	\revision{For our illustrative example, we draw inspiration from the practical system by Loianno and Kumar \cite{loianno2017cooperative}, where they develop a fleet of micro aerial vehicles (MAVs) that can transport a rigid body using permanent electromagnets. If the geometry of the transported object is assumed to be known a priori, they show that this additional information can aid the localisation of the vehicles by formulating the estimation problem as an (expensive) optimisation problem. In this section, we show how such additional information can be incorporated automatically into the well-established state estimation formulation (outlined in the previous section) using degenerate Gaussian factors. To keep the dimensionality manageable, however, we limit our discussion to ground vehicles with three degrees of freedom.
	
	In particular,} consider a scenario where a fleet of mobile robots need to work together to transport objects on a warehouse floor. Suppose that these objects can vary in shape and size, and consequently multiple and varying subsets of robots cooperate at any given time. Since each robot operates independently for the majority of the time, we use the \revision{nonlinear} odometry motion model
	\begin{linenomath*}\begin{equation} \label{eq:motion_odometry}
	\mathbf{x}_k^i=\begin{bmatrix}
	x_k^i \\
	y_k^i \\
	\theta_k^i
	\end{bmatrix}=\begin{bmatrix}
	x_{k-1}^i+r_k^i\cos\left(\theta_{k-1}^i+\alpha_k^i\right) \\
	y_{k-1}^i+r_k^i\sin\left(\theta_{k-1}^i+\alpha_k^i\right) \\
	\theta_{k-1}^i+\alpha_k^i+\beta_k^i
	\end{bmatrix}+\mathbf{w}_k^i
	\end{equation}\end{linenomath*}
	as proposed by Thrun et al. \cite{thrun2005probabilistic}. \revision{In \eqref{eq:motion_odometry},} the $i$'th robot has position $(x_k^i,y_k^i)$ and orientation $\theta_k^i$ and the corresponding control inputs comprise a rotation $\alpha_k^i$ followed by a translation $r_k^i$ and another rotation $\beta_k^i$. Suppose that each robot also receives a noisy measurement of its position
	\begin{linenomath*}\begin{equation} \label{eq:meas_position}
	\mathbf{z}_k^i=\begin{bmatrix}
	x_k^i \\
	y_k^i
	\end{bmatrix}+\mathbf{v}_k^i
	\end{equation}\end{linenomath*}
	at every time step and that the noise distributions
	\begin{linenomath*}\begin{equation}
	p(\mathbf{w}_k^i)=\mathcal{N}(\mathbf{w}_k^i;\mathbf{0},\Sigma_w)\qquad\text{and}\qquad p(\mathbf{v}_k^i)=\mathcal{N}(\mathbf{v}_k^i;\mathbf{0},\Sigma_v)
	\end{equation}\end{linenomath*}
	are Gaussian. To keep this model consistent with the one in Figure \ref{fig:model}, 
	we then combine the states (and similarly the measurements and controls) of all the individual robots into a single state vector $\mathbf{x}_k$ \revision{(as well as} measurement vector $\mathbf{z}_k$ and control vector $\mathbf{u}_k$).
	
	Now consider a moment in time $k'$ when a group of $N$ robots $\{j_1,j_2,\dots j_N\}$ are transporting an object. If the shape of the object is known, this provides additional information that would be useful for estimating the states of the robots. In this example, we assume that the distances between these robots as well as their relative orientations are specified exactly\footnote{For the sake of simplicity, we only consider translations of the object (as opposed to rotations as well).}. This additional information can therefore be \revision{captured} in a noiseless \emph{auxiliary} measurement $\hat{\mathbf{z}}_{k'}$, where its $n$'th component is given by
	\begin{linenomath*}\begin{equation} \label{eq:meas_distance}
	\hat{\mathbf{z}}_{k'}^n=\begin{bmatrix}
	\sqrt{\left(x_{k'}^{j_n}-x_{k'}^{j_{n-1}}\right)^2+\left(y_{k'}^{j_n}-y_{k'}^{j_{n-1}}\right)^2} \\
	\theta_{k'}^{j_n}-\theta_{k'}^{j_{n-1}}
	\end{bmatrix}.
	\end{equation}\end{linenomath*}
	The \emph{augmented} measurement model for time step $k'$ then appends all of the \revision{nonlinear} auxiliary measurements in \eqref{eq:meas_distance} to the default measurements in \eqref{eq:meas_position} to produce a single measurement vector $\mathbf{z}_{k'}$.
	
	\revision{During physical operation, each robot will receive its odometry information and sensor measurements as normal. If this were naively used to perform state estimation separately for each robot, the correlation between cooperating robots could not be utilised to improve the accuracy of the estimated beliefs. Instead, all the robot poses should be combined into a single state vector and auxiliary measurements of the form in \eqref{eq:meas_distance} included where applicable. Note that, although such an auxiliary measurement is expressed in terms of the robot states, this is not a causal relationship. This equation is merely used to construct the necessary conditional densities as required by the model in Figure \ref{fig:model}. The actual observations are instead determined by the geometry of the particular object and will be the same at every time step for the duration of the transportation task. Omitting the auxiliary measurements in \eqref{eq:meas_distance} altogether would be equivalent to ignoring the additional information provided by the known shape of the object.
	
	} A closer look at the necessary inference operations (as outlined in Section \ref{sec:recursive}) reveals the need for computing with degenerate factors in this example. Since the augmented measurement model contains noiseless components, the covariance of the measurement noise will be rank deficient and consequently the measurement factor $\rho_{k'}$ in \eqref{eq:meas_factor} cannot be represented using non-degenerate parametrisations. We therefore need to represent the noise distribution using a degenerate factor as in \eqref{eq:p_w}. Furthermore, the upward message $\xi^{\uparrow}_{k'}$ as computed in \eqref{eq:messages} will also be degenerate and will in turn be used to compute other messages (for example $\xi^{\rightarrow}_{k'}$ and $\xi^{\leftarrow}_{k'-1}$). This requires the reduction, multiplication and marginalisation operations as outlined in Section \ref{sec:operations}. Even the context for linearising the motion factor $\psi_{k'+1}$ -- typically the message $\xi^{\rightarrow}_{k'}$ -- will be degenerate. \revision{This reveals that,} once degeneracies arise during inference, it is necessary that \emph{all} of the downstream operations handle such cases appropriately.
	
	\subsection{Experiments and results}
	
\begin{figure}[htb]
		\centering
		\subfigure[]{\label{fig:estimation_a}\includegraphics[width=0.49\textwidth]{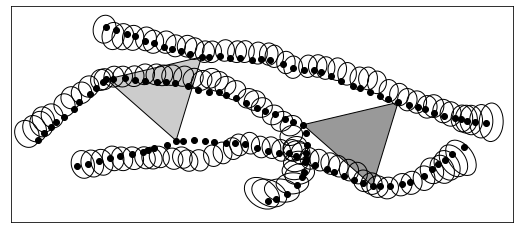}}
		\subfigure[]{\label{fig:estimation_b}\includegraphics[width=0.49\textwidth]{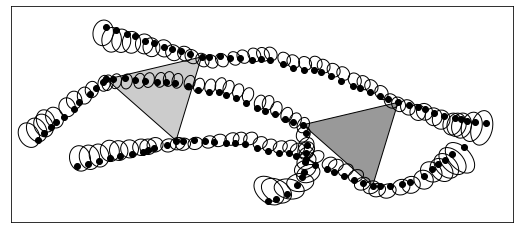}}
		\subfigure[]{\label{fig:estimation_c}\includegraphics[width=0.49\textwidth]{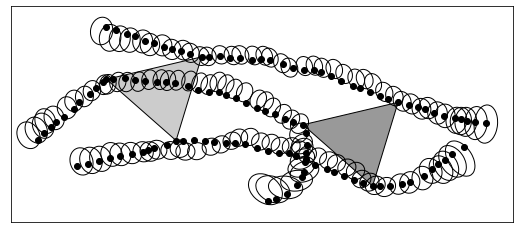}}
		\caption{State estimation for three robots transporting a triangular object (from left to right) on a 2-D warehouse floor when \revision{(a) ignoring the known shape of the object,} (b) using degenerate factors and (c) using canonical factors with ridge regularisation. The object's initial and final positions are indicated by the lighter and darker triangles, respectively. At each time step, the robots' actual positions are represented with solid dots and the inferred beliefs with 67\% confidence ellipses. \revision{Note that these ellipses are smaller in (b) compared to (a) and (c), as seen by the separation between the trajectories (specifically near the top right vertex of the lighter triangle and the left vertex of the darker triangle).}}
		\label{fig:result}
	\end{figure}

	\revision{To support this argument, the transportation task in Section \ref{sec:cooperative} was simulated, where generated control inputs and sampled noise values were used to calculate the trajectories and measurements of the robots. The control inputs and measurements were then used to calculate the messages in Figure \ref{fig:cluster} until convergence. As an example,} the beliefs for a group of three robots and a single object is shown in Figure \ref{fig:result}. By computing with degenerate factors, we \revision{were} able to handle cases where the motion of the three robots are independent (first and last 10 time steps) as well as highly-correlated (middle 20 time steps) -- both automatically and without encountering numerical errors. In comparison, \revision{ignoring the additional information provided by the known shape resulted in more uncertain beliefs. This was also the case when} using ridge regularisation (i.e., adding a small scalar value $\lambda$ to the diagonal terms of the \revision{singular} covariance matrix) as an approximate solution.
	
\begin{figure}[htb]
		\centering
		\input{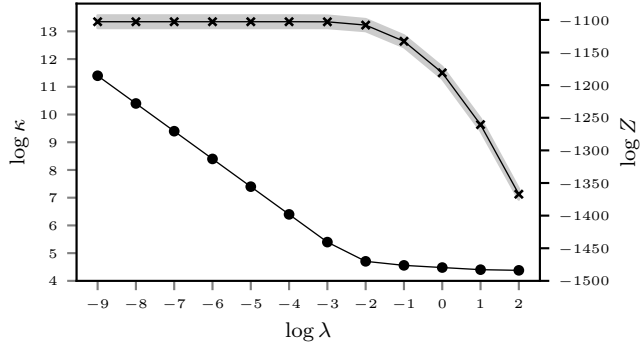}
		\caption{The effect of ridge regularisation (for varying values of $\lambda$) on the maximum condition number $\kappa$ (indicated by the line with circular markers) as well as on the model likelihood $Z$ (indicated by the line with x-shaped markers). The standard deviation for the latter (based on multiple experiments with \revision{randomly-generated} control inputs and noise values) are indicated by the shaded region.}
		\label{fig:conditon_number}
	\end{figure}

	\revision{For repeated experiments,} Figure \ref{fig:conditon_number} shows the trade-off between high accuracy for small regularisation values and well-conditioned matrices for larger values. \revision{In general,} the critical point when too large a condition number results in numerical errors depends on the machine precision and other application-specific details. \revision{However,} it is worth noting that even for this example, there is \revision{only} a limited range of regularisation values that achieve acceptable accuracy without affecting the conditioning of the problem adversely. This is in contrast to our principled solution using degenerate factors, where the condition numbers are determined by the problem definition alone \revision{and not increased} by unnecessary approximations.

	Since the degenerate parametrisation proposed by Raphael \cite{raphael2003bayesian} does not include a procedure for approximating nonlinear models, we need to convert to (and from) our parametrisation to enable a comparison. The more significant drawback of the former, however, is the absence of normalisation constants due to their use of indicator functions (as opposed to our use of Dirac delta functions) for representing degeneracies. The implication is that, although the first- and second-order moments of the computed beliefs are equivalent in both cases, \revision{Bayesian} model comparison is only possible with our parametrisation (where the messages remain unnormalised).
	
\begin{figure}[htb]
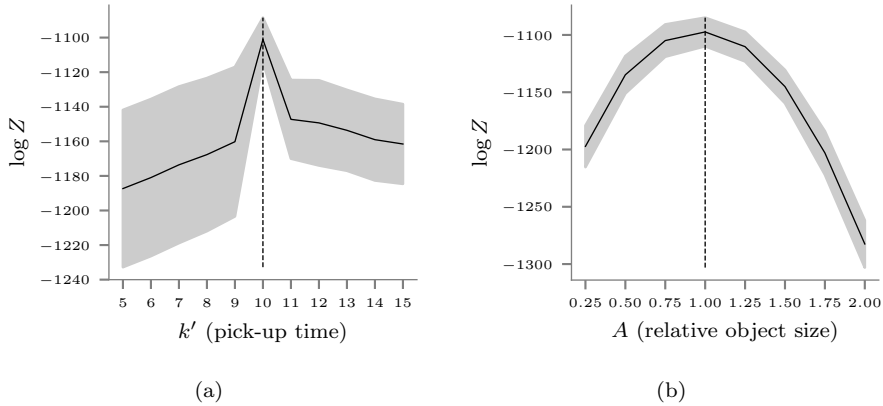

		\centering
		\subfigure[]{\label{fig:model_1}\input{figures/timestep.pgf}}
		\subfigure[]{\label{fig:model_2}\input{figures/size.pgf}}
		\caption{Model comparison for the state estimation problem in Figure \ref{fig:result} to determine (a) the time step $k'$ when the object was picked up and (b) the size $A$ of the object (relative to the true object's size). In both cases, the mean and standard deviation of the model likelihood $Z$ are indicated by the solid line and shaded region, respectively. The peak where the model likelihood is a maximum (and therefore corresponding to the most-likely model) is indicated by the dashed line.}
		\label{fig:model_evidence}
	\end{figure}

	To illustrate this advantage, suppose that the time when the object was picked up is unknown but of interest. By considering multiple hypotheses (i.e., one for each time step), performing inference for each model and then extracting the model likelihood using the normalisation information, we are able to identify the correct model. This is shown in Figure \ref{fig:model_1}, where the peak at $k'=10$ corresponds to the true time step when the object was picked up in Figure \ref{fig:result}. Similarly, Figure \ref{fig:model_2} illustrates an alternative context where the transported object could be any one out of a finite set of candidates and the relative size of the object is correctly identified (as $A=1$) using the model likelihood.
	
	\revision{As a final comparison of our methodology to both ridge regularisation and Raphael's factor representation, we investigate the overall execution times for inferring the posterior belief over the robots' trajectories. In the case of ridge regularisation, the use of Koller and Friedman's \cite{koller2009} canonical factors resulted in the shortest execution times, as expected. This approach required an average of 0.67 seconds to solve the entire inference problem\footnote{These runtime results were obtained using unoptimised implementations and standard Python libraries such as \texttt{numpy} and \texttt{scipy}. The computations were performed on a single thread of an Intel\text{\textregistered} Core\text{\texttrademark} i7 CPU.}. Next, using the parametrisation for degenerate factors by Raphael \cite{raphael2003bayesian} took 1.58 seconds on average. In some cases, however, this resulted in numerical errors when using the Cholesky decomposition to draw the necessary sigma points. This reveals that, although these factors can be used to approximate nonlinear models, the use of a \emph{diagonal} precision matrix as in \eqref{eq:degen_def} improves the numerical stability significantly, since its inverse and square root can be computed in an element-wise manner.
	
	Finally, using our degenerate Gaussian factors to solve this recursive state estimation problem required 1.51 seconds on average. Although this is only slightly faster than Raphael's representation, recall that the normalisation constants are explicitly kept track of throughout the former, but not computed at all in the latter. For a fair comparison, we can therefore adapt our solution by omitting the final line in each of Algorithms \ref{alg:marginal} to \ref{alg:equiv_transform}. Consequently, if we are only interested in the first- and second-order moments of the computed beliefs, the execution time drops to 1.34 seconds on average. In summary, ridge regularisation therefore had the shortest execution time, but requires approximations in degenerate cases. Raphael's representation can accommodate degeneracies exactly, but cannot be used for model comparison and required the longest execution time in this example. As illustrated, using our parametrised factors instead enjoys the advantages of both while requiring more time than the former, but less than the latter.}

\section{Conclusions} \label{sec:conclusion}

	Probabilistic inference involving Gaussian factors is only valid for positive-definite covariance matrices. In positive \emph{semi}-definite settings, linear dependencies among the random variables instead warrant the explicit representation of generalised \emph{degenerate} Gaussian factors. To this end, this paper introduced a \revision{novel} parametrisation \revision{comprising} a lower-dimensional, non-degenerate component as well as a Dirac delta function describing possible degeneracies. \revision{By using a Dirac delta function we can keep track of normalisation constants explicitly, which in turn enables model comparison in degenerate cases. Our factor definition can still express non-degenerate Gaussian distributions as a special case and provides a principled way of handling rank-deficient transformations of Gaussian random variables.}
	
	We further derived \revision{algorithms for} typical statistical operations \revision{on degenerate Gaussian factors from first principles. This means} that all the capabilities of non-degenerate factors are preserved. \revision{In addition, we developed a method for linearising nonlinear models via the unscented transform when the prior distribution is degenerate. Specifically, by only drawing sigma points from the lower-dimensional distribution, we avoid the positive-definite constraints associated with the Cholesky decomposition. Collectively, these contributions} enable accurate and automatic inference in (possibly) degenerate settings at little additional computational cost. In contrast, approximate solutions sacrifice either accuracy or numerical stability as illustrated by the recursive state estimation example.
	
	\revision{Future work should investigate further applications of Gaussian random variables where it would be advantageous to compute with explicit degenerate factors. In addition, since inference problems can become near-degenerate due to machine precision limitations, the numerical stability of inference with ill-conditioned matrices could be improved significantly.} Finally, by utilising the lower-dimensional, diagonal precision matrix of our representation as part of an optimised implementation, future work could reduce the computational cost of performing inference on models with a significant degree of degeneracy.

\section*{CRediT authorship contribution statement}
\markright{}
\addcontentsline{toc}{section}{CRediT authorship contribution statement}

\textbf{J.C. Schoeman:} Conceptualization, Methodology, Software, Formal analysis, Writing - Original Draft, Writing - Review \& Editing. \textbf{C.E. van Daalen:} Conceptualization, Writing - Review \& Editing, Supervision. \textbf{J.A. du Preez:} Writing - Review \& Editing, Supervision.

\section*{Declaration of competing interest}
\markright{}
\addcontentsline{toc}{section}{Declaration of competing interest}

The authors declare that they have no known competing financial interests or personal relationships that could have appeared to influence the work reported in this paper.

\section*{Acknowledgements}
\markright{}
\addcontentsline{toc}{section}{Acknowledgements}

This work was supported by grants from the Wilhelm Frank Bursary Fund as administered by Stellenbosch University from 2017 to 2019. \revision{The authors would also like to thank anonymous reviewers for their valuable suggestions and comments, which improved the final version of the paper significantly.}

\appendix

\section{Identities related to canonical factors and Dirac delta functions} \label{sec:canonical_identities}

\begin{result} \label{result:canonical_scope}
	A canonical factor with scope $A\mathbf{x}+\mathbf{b}$ can be expressed over scope $\mathbf{x}$ according to
	\begin{linenomath*}\begin{equation} \label{eq:canonical_scope}
		\mathcal{C}(A\mathbf{x}+\mathbf{b};K,\mathbf{h},g)=\mathcal{C}\left(\mathbf{x};A^TKA,A^T(\mathbf{h}-K\mathbf{b}),g+\left(\mathbf{h}-\frac{1}{2}K\mathbf{b}\right)^T\mathbf{b}\right).
	\end{equation}\end{linenomath*}
\end{result}
\begin{proof}
	Using the definition of a canonical factor in \eqref{eq:canon_def} we can write
	\begin{linenomath*}\begin{equation}
	\phi=\mathcal{C}(A\mathbf{x}+\mathbf{b};K,\mathbf{h},g)=\exp\left(-\frac{1}{2}(A\mathbf{x}+\mathbf{b})^TK(A\mathbf{x}+\mathbf{b})+\mathbf{h}^T(A\mathbf{x}+\mathbf{b})+g\right).
	\end{equation}\end{linenomath*}
	Rearranging terms yields
	\begin{linenomath*}\begin{equation}
	\phi=\exp\left(-\frac{1}{2}\mathbf{x}^TA^TKA\mathbf{x}+(\mathbf{h}-K\mathbf{b})^TA\mathbf{x}+g+\left(\mathbf{h}-\frac{1}{2}K\mathbf{b}\right)^T\mathbf{b}\right).
	\end{equation}\end{linenomath*}
	This concludes the proof for Result \ref{result:canonical_scope}.
\end{proof}

\begin{result} \label{result:norm_can_Gaussian}
	A normalised canonical factor can be expressed as a Gaussian density with covariance parametrisation according to
	\begin{linenomath*}\begin{equation} \label{eq:norm_can_Gaussian}
		\mathcal{C}(\mathbf{x};K,\mathbf{h},g)=\mathcal{N}(\mathbf{x};K^{-1}\mathbf{h},K^{-1}).
	\end{equation}\end{linenomath*}
\end{result}
\begin{proof}
	By substituting the condition for a normalised factor in \eqref{eq:g_for_pdf} into the definition of a canonical factor in \eqref{eq:canon_def} we can write
	\begin{linenomath*}\begin{equation}
	\mathcal{C}(\mathbf{x};K,\mathbf{h},g)=\exp\left(-\frac{1}{2}\mathbf{x}^TK\mathbf{x}+\mathbf{h}^T\mathbf{x}-\frac{1}{2}\mathbf{h}^TK^{-1}\mathbf{h}-\frac{1}{2}\log\left|2\pi K^{-1}\right|\right).
	\end{equation}\end{linenomath*}
	Rearranging terms and moving the normalisation constant out of the exponent yields a factor in the covariance form, i.e.,
	\begin{linenomath*}\begin{equation}
	\mathcal{C}(\mathbf{x};K,\mathbf{h},g)=\frac{1}{\sqrt{\left|2\pi K^{-1}\right|}}\exp\left(-\frac{1}{2}(\mathbf{x}-K^{-1}\mathbf{h})^TK(\mathbf{x}-K^{-1}\mathbf{h})\right).
	\end{equation}\end{linenomath*}
	This concludes the proof for Result \ref{result:norm_can_Gaussian}.
\end{proof}

\begin{result} \label{result:joint_lim_degen_complete}
	A normalised degenerate factor can be expressed as the limit of a non-degenerate Gaussian density according to
	\begin{linenomath*}\begin{equation}
	\mathcal{D}(\mathbf{x};Q,R,\Lambda,\mathbf{h},\mathbf{c},g)=\lim_{a \rightarrow 0}\mathcal{N}\left(\mathbf{x};Q\Lambda^{-1}\mathbf{h}+R\mathbf{c},Q\Lambda^{-1}Q^T+aRR^T
	\right).
	\end{equation}\end{linenomath*}
\end{result}
\begin{proof}
	By substituting the result in \eqref{eq:norm_can_Gaussian} as well as the definition of a multidimensional Dirac delta in \eqref{eq:ND_Dirac_as_Gaussian} into the definition of a degenerate factor in \eqref{eq:degen_def} we can write
	\begin{linenomath*}\begin{equation} \label{eq:joint_lim_degen_complete_2}
		\phi=\mathcal{D}(\mathbf{x};Q,R,\Lambda,\mathbf{h},\mathbf{c},g)=\mathcal{N}(Q^T\mathbf{x};\Lambda^{-1}\mathbf{h},\Lambda^{-1})\lim_{a \rightarrow 0}\mathcal{N}\left(R^T\mathbf{x}-\mathbf{c};\mathbf{0},aI\right).
	\end{equation}\end{linenomath*}
	By moving the first term into the limit and expressing the factored product as a joint density, \eqref{eq:joint_lim_degen_complete_2} becomes
	\begin{linenomath*}\begin{equation} \label{eq:joint_lim_degen_complete_3}
	\phi=\lim_{a \rightarrow 0}\mathcal{N}\left(\begin{bmatrix}
	Q^T \\
	R^T\end{bmatrix}\mathbf{x};\begin{bmatrix}
	\Lambda^{-1}\mathbf{h} \\
	\mathbf{c}
	\end{bmatrix},\begin{bmatrix}
	\Lambda^{-1} & \textit{0} \\
	\textit{0} & aI
	\end{bmatrix}
	\right).
	\end{equation}\end{linenomath*}
	Since $\mathcal{N}(U^T\mathbf{x};\boldsymbol{\mu},\Sigma)=\mathcal{N}(\mathbf{x};U\boldsymbol{\mu},U\Sigma U^T)$ for an orthogonal transformation $U$, \eqref{eq:joint_lim_degen_complete_3} can further be simplified according to
	\begin{linenomath*}\begin{align} 
	\phi&=\lim_{a \rightarrow 0}\mathcal{N}\left(\mathbf{x};\begin{bmatrix}
	Q & R
	\end{bmatrix}\begin{bmatrix}
	\Lambda^{-1}\mathbf{h} \\
	\mathbf{c}
	\end{bmatrix},\begin{bmatrix}
	Q & R
	\end{bmatrix}\begin{bmatrix}
	\Lambda^{-1} & \textit{0} \\
	\textit{0} & aI
	\end{bmatrix}\begin{bmatrix}
	Q & R
	\end{bmatrix}^T
	\right) \nonumber \\
	&=\lim_{a \rightarrow 0}\mathcal{N}\left(\mathbf{x};Q\Lambda^{-1}\mathbf{h}+R\mathbf{c},Q\Lambda^{-1}Q^T+aRR^T
	\right).
	\end{align}\end{linenomath*}
	This concludes the proof for Result \ref{result:joint_lim_degen_complete}.
\end{proof}

\begin{result} \label{result:dirac_properties}
	Three further properties of the Dirac delta function follow from those in \eqref{eq:dirac_prop}. Firstly, multiplying a function $f(x)$ that is continuous at \revision{$x=a$} with the Dirac delta $\delta(x-a)$ yields
	\begin{linenomath*}\begin{equation} \label{eq:Dirac_multiply}
	f(x)\,\delta(x-a)=f(a)\,\delta(x-a).
	\end{equation}\end{linenomath*}
	Furthermore, the sampling (or sifting) property states that
	\begin{linenomath*}\begin{equation} \label{eq:Dirac_sift}
	\int_{-\infty}^{\infty}f(x)\,\delta(x-a)\,dx=f(a).
	\end{equation}\end{linenomath*}
	Finally, the scaling property states that
	\begin{linenomath*}\begin{equation} \label{eq:Dirac_scale}
	\delta(kx)=\frac{1}{|k|}\delta(x).
	\end{equation}\end{linenomath*}
\end{result}
\revision{\begin{proof}
	Since $\delta(x-a)=0$ where $x\neq a$, the product in \eqref{eq:Dirac_multiply} is independent of $f(x)$ where $x\neq a$. This subsequently means that
	\begin{linenomath*}\begin{equation}
	\int_{-\infty}^{\infty}f(x)\,\delta(x-a)\,dx=\int_{-\infty}^{\infty}f(a)\,\delta(x-a)\,dx=f(a).
	\end{equation}\end{linenomath*}
	To see that $|k|\delta(kx)=\delta(x)$, we check that the two properties in \eqref{eq:dirac_prop} are satisfied. The first follows directly and for the second we have that
	\begin{linenomath*}\begin{align}
	\int_{-\infty}^{\infty}|k|\delta(kx)\,dx&=\begin{cases}
	|k|\int_{-\infty}^{\infty}\delta(y)\,\frac{dy}{k} & \quad k>0 \\
	|k|\int_{\infty}^{-\infty}\delta(y)\,\frac{dy}{k} & \quad k<0
	\end{cases} \nonumber \\
	&=\frac{k}{k}\int_{-\infty}^{\infty}\delta(y)\,dy=1.
	\end{align}\end{linenomath*}
	This concludes the proof for Result \ref{result:dirac_properties}.
\end{proof}}

\begin{result}
	Using the definition in \eqref{eq:ND_Diracdelta} we can use element-wise arguments to show that similar properties (to those in \eqref{eq:Dirac_multiply} and \eqref{eq:Dirac_sift}) hold for the multidimensional Dirac delta function, i.e.,
	\begin{linenomath*}\begin{equation} \label{eq:dirac_prod_multi}
	f(\mathbf{x})\,\delta(\mathbf{x}-\mathbf{a})=f(\mathbf{a})\,\delta(\mathbf{x}-\mathbf{a})
	\end{equation}\end{linenomath*}
	and
	\begin{linenomath*}\begin{equation} \label{eq:dirac_sift_multi}
	\int_{-\infty}^{\infty}f(\mathbf{x})\,\delta(\mathbf{x}-\mathbf{a})\,\text{d}\mathbf{x}=f(\mathbf{a}).
	\end{equation}\end{linenomath*}
\end{result}

\begin{result} \label{result:rotated_Dirac_delta}
	For an $n\times n$ non-singular matrix $A$, we can write
	\begin{linenomath*}\begin{equation} \label{eq:rotated_Dirac_delta_2}
	\delta\left(A\mathbf{x}+\mathbf{b}\right) = \frac{1}{\sqrt{|A|^2}}\delta\left(\mathbf{x}+A^{-1}\mathbf{b}\right).
	\end{equation}\end{linenomath*}
	If $A$ is orthogonal and $\mathbf{b}=\mathbf{0}$, the result reduces to
	\begin{linenomath*}\begin{equation} \label{eq:rotated_Dirac_delta_3}
	\delta\left(A\mathbf{x}\right) = \delta\left(\mathbf{x}\right).
	\end{equation}\end{linenomath*}
\end{result}
\begin{proof}
	By using the definition of the multidimensional Dirac delta in \eqref{eq:ND_Dirac_as_Gaussian}, the SVD $A = U\Sigma V^T$, and the fact that $U^TU = I$, we can write
	\begin{linenomath*}\begin{equation} \label{eq:A_sigma_V}
	\delta\left(A\mathbf{x}\right) = \lim_{a \rightarrow 0}(2\pi a)^{-\frac{n}{2}}\exp\left(-\frac{1}{2a}\mathbf{x}^TA^TA\mathbf{x}\right) = \lim_{a \rightarrow 0}(2\pi a)^{-\frac{n}{2}}\exp\left(-\frac{1}{2a}\mathbf{x}^TV\Sigma^T \Sigma V^T\mathbf{x}\right) = \delta\left(\Sigma V^T\mathbf{x}\right).
	\end{equation}\end{linenomath*}
	Since $\Sigma$ is diagonal, we can use \eqref{eq:ND_Diracdelta} to write
	\begin{linenomath*}\begin{equation} \label{eq:sigma_V_prod}
	\delta\left(\Sigma V^T\mathbf{x}\right) = \prod_{i=1}^n\delta\left(\sigma_i^{}\mathbf{v}_i^T\mathbf{x}\right),
	\end{equation}\end{linenomath*}
	where $\sigma_i$ is the $i$th singular value of $A$ and $\mathbf{v}_i$ is the $i$th column of $V$. Using the scaling property of the Dirac delta function in \eqref{eq:Dirac_scale}, we can combine \eqref{eq:A_sigma_V} and \eqref{eq:sigma_V_prod} to write
	\begin{linenomath*}\begin{equation} \label{eq:dirac_A_V}
	\delta\left(A\mathbf{x}\right)  = \prod_{i=1}^n\delta\left(\sigma_i\mathbf{v}_i^T\mathbf{x}\right)
	 = \prod_{i=1}^n\frac{1}{|\sigma_i|}\delta\left(\mathbf{v}_i^T\mathbf{x}\right)
	 = \left(\prod_{i=1}^n\frac{1}{|\sigma_i|}\right)\delta\left(V^T\mathbf{x}\right).
	\end{equation}\end{linenomath*}
	Using the definition in \eqref{eq:ND_Dirac_as_Gaussian} again, and since $VV^T=I$ and $|A|=|\Sigma|=\prod\sigma_i$, we can further expand the Dirac delta in \eqref{eq:dirac_A_V} to write
	\begin{linenomath*}\begin{equation} \label{eq:rotated_dirac_result}
	\delta\left(A\mathbf{x}\right) = \left(\prod_{i=1}^n\frac{1}{|\sigma_i|}\right) \lim_{a\rightarrow 0}(2\pi a)^{-\frac{n}{2}}\exp\left(-\frac{1}{2a}\mathbf{x}^T\mathbf{x}\right)
	 = \frac{1}{\sqrt{|A|^2}}\delta\left(\mathbf{x}\right).
	\end{equation}\end{linenomath*}
	The more general result for a non-zero offset $\mathbf{b}$ is obtained by substituting $\mathbf{x}=\mathbf{y}+A^{-1}\mathbf{b}$ into \eqref{eq:rotated_dirac_result} and if $A$ is orthogonal, $|A|=1$. This concludes the proof for Result \ref{result:rotated_Dirac_delta}.
\end{proof}

\section{Derivations of main results} \label{sec:proofs}

\subsection{Affine transformations of degenerate random variables}

To determine the parameters of the degenerate factor representing the density $p(\mathbf{y})$ as calculated according to Algorithm \ref{alg:affine_transform}, we can use \eqref{eq:mean_degen}, \eqref{eq:cov_degen} and \eqref{eq:y_Ax_b} to calculate the mean
\begin{linenomath*}\begin{equation} \label{eq:mean_y}
\mathbb{E}[\mathbf{y}]= A\,\mathbb{E}[\mathbf{x}]+\mathbf{b}\quad\Longrightarrow\quad Q'\Lambda'^{-1}\mathbf{h}'+R'\mathbf{c}'=A(Q\Lambda^{-1}\mathbf{h}+R\mathbf{c})+\mathbf{b}
\end{equation}\end{linenomath*}
and covariance
\begin{linenomath*}\begin{equation} \label{eq:cov_y}
\text{Cov}\left[\mathbf{y}\right] = A\,\text{Cov}\left[\mathbf{x}\right]A^T\quad\Longrightarrow\quad Q'\Lambda'^{-1}Q'^T=AQ\Lambda^{-1}Q^TA^T.
\end{equation}\end{linenomath*}
The quantities $Q'$ and $\Lambda'$ can then be calculated directly from \eqref{eq:cov_y} using the compact SVD. By definition, $C(R')=C(Q')^\perp$, and $R'$ can be determined accordingly. With these quantities known, $\mathbf{h}'$ and $\mathbf{c}'$ can be determined from \eqref{eq:mean_y} by exploiting the mutual orthogonality of $Q'$ and $R'$, where we can write
\begin{linenomath*}\begin{equation}
R'^T\left(Q'\Lambda'^{-1}\mathbf{h}'+R'\mathbf{c}'\right)=\mathbf{c}'=R'^T\left(A(Q\Lambda^{-1}\mathbf{h}+R\mathbf{c})+\mathbf{b}\right)
\end{equation}\end{linenomath*}
and
\begin{linenomath*}\begin{equation}
\Lambda'Q'^T\left(Q'\Lambda'^{-1}\mathbf{h}'+R'\mathbf{c}'\right)=\mathbf{h}'=\Lambda'Q'^T\left(A(Q\Lambda^{-1}\mathbf{h}+R\mathbf{c})+\mathbf{b}\right).
\end{equation}\end{linenomath*}
Finally, since the resulting density $p(\mathbf{y})$ will be normalised, $g'$ is calculated by direct application of \eqref{eq:g_norm}. This concludes the derivation of Algorithm \ref{alg:affine_transform}.

\subsection{Marginalisation}
	
To determine the parameters of the marginal factor $\phi(\mathbf{x})$ as calculated according to Algorithm \ref{alg:marginal}, we need to evaluate the integral
\begin{linenomath*}\begin{equation} \label{eq:int_marg_degen}
\mathcal{J}=\int \mathcal{C}\left(\begin{bmatrix}
Q_{\mathbf{x}}^T & Q_{\mathbf{y}}^T
\end{bmatrix}\begin{bmatrix}
\mathbf{x} \\
\mathbf{y}
\end{bmatrix};\Lambda,\mathbf{h},g\right)\,\delta\left(\begin{bmatrix}
R_{\mathbf{x}}^T & R_{\mathbf{y}}^T
\end{bmatrix}\begin{bmatrix}
\mathbf{x} \\
\mathbf{y}
\end{bmatrix}-\mathbf{c}\right)\,\text{d}\mathbf{y}.
\end{equation}\end{linenomath*}
\revision{Since the matrix $\begin{bmatrix}
Q & R
\end{bmatrix}$ is orthogonal, some important relationships that will be useful throughout this derivation are
\begin{linenomath*}\begin{align} \label{eq:QxRx}
Q_{\mathbf{x}}^TR_{\mathbf{x}}^{}+Q_{\mathbf{y}}^TR_{\mathbf{y}}^{}=\textit{0}\quad&\Longrightarrow\quad Q_{\mathbf{x}}^TR_{\mathbf{x}}^{}=-Q_{\mathbf{y}}^TR_{\mathbf{y}}^{} \nonumber \\
Q_{\mathbf{x}}^TQ_{\mathbf{x}}^{}+Q_{\mathbf{y}}^TQ_{\mathbf{y}}^{}=I\quad&\Longrightarrow\quad Q_{\mathbf{x}}^TQ_{\mathbf{x}}^{}=I-Q_{\mathbf{y}}^TQ_{\mathbf{y}}^{} \nonumber \\
R_{\mathbf{x}}^TR_{\mathbf{x}}^{}+R_{\mathbf{y}}^TR_{\mathbf{y}}^{}=I\quad&\Longrightarrow\quad R_{\mathbf{x}}^TR_{\mathbf{x}}^{}=I-R_{\mathbf{y}}^TR_{\mathbf{y}}^{}.
\end{align}\end{linenomath*}}
The first step \revision{for calculating the integral in \eqref{eq:int_marg_degen}} is to define an appropriate substitution. This is achieved by determining orthogonal bases for the nullspaces of the four block matrices $Q_{\mathbf{x}}$, $Q_{\mathbf{y}}$, $R_{\mathbf{x}}$ and $R_{\mathbf{y}}$ \revision{such that
\begin{linenomath*}\begin{equation} \label{eq:C_ABUV}
C(V_1)=N(Q_{\mathbf{x}}), \quad C(V_2)=N(Q_{\mathbf{y}}), \quad C(V_3)=N(R_{\mathbf{x}}) \quad \text{and} \quad C(V_4)=N(R_{\mathbf{y}}).
\end{equation}\end{linenomath*}}
Since the matrix $\begin{bmatrix}
Q & R
\end{bmatrix}$ has full rank, $N(Q_{\mathbf{x}})\perp N(Q_{\mathbf{y}})$ and $N(R_{\mathbf{x}})\perp N(R_{\mathbf{y}})$, although the pairs are not necessarily orthogonal complements. This is because there could for example be a vector $\mathbf{w}\in\mathbb{R}^{n-k}$ such that $\mathbf{w}\notin N(Q_{\mathbf{x}})$ and  $\mathbf{w}\notin N(Q_{\mathbf{y}})$. The implication is that the semi-orthogonal matrices $\begin{bmatrix}
V_1 & V_2
\end{bmatrix}$ and $\begin{bmatrix}
V_3 & V_4
\end{bmatrix}$ are not necessarily square. We therefore need additional bases to guarantee an orthogonal decomposition of $\mathbb{R}^n$. For this purpose, we define two ad-hoc matrices \revision{$A$ and $B$} such that $A_{\mathbf{y}}=\textit{0}$ and $B_{\mathbf{x}}=\textit{0}$. Using the definitions in \eqref{eq:C_ABUV}, the orthogonal matrix
\begin{linenomath*}\begin{equation}
\begin{bmatrix}
QV_1 & QV_2 & RV_3 & RV_4 & A & B
\end{bmatrix}=\begin{bmatrix}
\textit{0} & Q_{\mathbf{x}}V_2 & \textit{0} & R_{\mathbf{x}}V_4 & A_{\mathbf{x}}^{} & \textit{0} \\
Q_{\mathbf{y}}V_1 & \textit{0} & R_{\mathbf{y}}V_3 & \textit{0} & \textit{0} & B_{\mathbf{y}}^{}
\end{bmatrix}
\end{equation}\end{linenomath*}
reveals that
\begin{linenomath*}\begin{equation} \label{eq:F_G}
C(A_{\mathbf{x}})=\left(C\left(Q_{\mathbf{x}}V_2\right)+C\left(R_{\mathbf{x}}V_4\right)\right)^\perp \quad\text{and}\quad C(B_{\mathbf{y}})=\left(C\left(Q_{\mathbf{y}}V_1\right)+C\left(R_{\mathbf{y}}V_3\right)\right)^\perp.
\end{equation}\end{linenomath*}
\revision{From \eqref{eq:QxRx} and \eqref{eq:C_ABUV}, we further have that $Q_{\mathbf{x}}^TR_{\mathbf{x}}^{}V_4^{}=\textit{0}$ and consequently
\begin{linenomath*}\begin{equation} \label{eq:C_Qx}
C(Q_{\mathbf{x}})=C\left(R_{\mathbf{x}}V_4\right)^\perp=C\left(\begin{bmatrix}
Q_{\mathbf{x}}V_2 & A_{\mathbf{x}}
\end{bmatrix}\right).
\end{equation}\end{linenomath*}}
Returning to the integral in \eqref{eq:int_marg_degen} and using the substitution
\begin{linenomath*}\begin{equation} \label{eq:xy_alpha}
\begin{bmatrix}
\mathbf{x} \\
\mathbf{y}
\end{bmatrix}=\begin{bmatrix}
QV_1 & QV_2 & RV_3 & RV_4 & A & B
\end{bmatrix}\begin{bmatrix}
\boldsymbol{\alpha} \\
\boldsymbol{\beta} \\
\boldsymbol{\gamma} \\
\boldsymbol{\theta} \\
\boldsymbol{\omega} \\
\boldsymbol{\rho}
\end{bmatrix},
\end{equation}\end{linenomath*}
as well as the fact that $Q^TR=\textit{0}$, we can write
\begin{linenomath*}\begin{equation} \label{eq:xy_alpha_2}
\begin{bmatrix}
Q_{\mathbf{x}}^T & Q_{\mathbf{y}}^T
\end{bmatrix}\begin{bmatrix}
\mathbf{x} \\
\mathbf{y}
\end{bmatrix}=\begin{bmatrix}
V_1^{} & V_2^{} & Q_{\mathbf{x}}^TA_{\mathbf{x}}^{} & Q_{\mathbf{y}}^TB_{\mathbf{y}}^{}
\end{bmatrix}\begin{bmatrix}
\boldsymbol{\alpha} \\
\boldsymbol{\beta} \\
\boldsymbol{\omega} \\
\boldsymbol{\rho}
\end{bmatrix}
\end{equation}\end{linenomath*}
and
\begin{linenomath*}\begin{equation} \label{eq:xy_alpha_3}
\begin{bmatrix}
R_{\mathbf{x}}^T & R_{\mathbf{y}}^T
\end{bmatrix}\begin{bmatrix}
\mathbf{x} \\
\mathbf{y}
\end{bmatrix}=\begin{bmatrix}
V_3^{} & V_4^{} & R_{\mathbf{x}}^TA_{\mathbf{x}}^{} & R_{\mathbf{y}}^TB_{\mathbf{y}}^{}
\end{bmatrix}\begin{bmatrix}
\boldsymbol{\gamma} \\
\boldsymbol{\theta} \\
\boldsymbol{\omega} \\
\boldsymbol{\rho}
\end{bmatrix}.
\end{equation}\end{linenomath*}
The next step is to decompose the Dirac delta in \eqref{eq:int_marg_degen} into three convenient components. This is achieved by multiplying its argument with the orthogonal matrix $\begin{bmatrix}
V_3^{} & V_4^{} & W
\end{bmatrix}^T$, where
\revision{\begin{linenomath*}\begin{equation}
C(W)=\left(N\left(R_{\mathbf{x}}\right)+N\left(R_{\mathbf{y}}\right)\right)^\perp=C\left(R_{\mathbf{x}}^T\right)\cap C\left(R_{\mathbf{y}}^T\right)=C\left(R_{\mathbf{x}}^TQ_{\mathbf{x}}^{}\right)=C\left(R_{\mathbf{y}}^TQ_{\mathbf{y}}^{}\right)
\end{equation}\end{linenomath*}
follows from \eqref{eq:QxRx} and \eqref{eq:C_ABUV}.} Using the substitution in \eqref{eq:xy_alpha_3} and the result in \eqref{eq:rotated_Dirac_delta_3}, \revision{after this multiplication} the Dirac delta becomes
\begin{linenomath*}\begin{equation}
\delta\left(\begin{bmatrix}
R_{\mathbf{x}}^T & R_{\mathbf{y}}^T
\end{bmatrix}\begin{bmatrix}
\mathbf{x} \\
\mathbf{y}
\end{bmatrix}-\mathbf{c}\right)=\delta\left(\begin{bmatrix}
\boldsymbol{\gamma}-V_3^{T}\mathbf{c} \\
\boldsymbol{\theta}-V_4^T\mathbf{c} \\
W^TR_{\mathbf{x}}^TA_{\mathbf{x}}^{}\boldsymbol{\omega} + W^TR_{\mathbf{y}}^TB_{\mathbf{y}}^{}\boldsymbol{\rho}-W^T\mathbf{c}
\end{bmatrix}\right).
\end{equation}\end{linenomath*}
Note that $V_3^TR_{\mathbf{x}}^TA_{\mathbf{x}}^{}=\textit{0}$ and $V_4^TR_{\mathbf{y}}^TB_{\mathbf{y}}^{}=\textit{0}$ from \eqref{eq:C_ABUV} and $V_3^TR_{\mathbf{y}}^TB_{\mathbf{y}}^{}=\textit{0}$ and $V_4^TR_{\mathbf{x}}^TA_{\mathbf{x}}^{}=\textit{0}$ from \eqref{eq:F_G}. By also using the substitution in \eqref{eq:xy_alpha_2} and the definition in \eqref{eq:ND_Diracdelta}, and since $\boldsymbol{\theta}$ is independent of the variables of integration and the integral over a Dirac delta is unity, the integral in \eqref{eq:int_marg_degen} becomes
\begin{linenomath*}\begin{equation} \label{eq:J_3}
\mathcal{J}=\delta(\boldsymbol{\theta} - V_4^T\mathbf{c})\int \mathcal{C}\left(\begin{bmatrix}
V_1^{} & V_2^{} & Q_{\mathbf{x}}^TA_{\mathbf{x}}^{} & Q_{\mathbf{y}}^TB_{\mathbf{y}}^{}
\end{bmatrix}\begin{bmatrix}
\boldsymbol{\alpha} \\
\boldsymbol{\beta} \\
\boldsymbol{\omega} \\
\boldsymbol{\rho}
\end{bmatrix};\Lambda,\mathbf{h},g\right)\delta\left(W^TR_{\mathbf{x}}^TA_{\mathbf{x}}^{}\boldsymbol{\omega} + W^TR_{\mathbf{y}}^TB_{\mathbf{y}}^{}\boldsymbol{\rho}-W^T\mathbf{c}\right)\,\text{d}\boldsymbol{\alpha}\,\text{d}\boldsymbol{\rho}.
\end{equation}\end{linenomath*}
The next step is to use the sifting property in \eqref{eq:dirac_sift_multi}. For this purpose, we first use \eqref{eq:rotated_Dirac_delta_2} to express the argument of the Dirac delta inside the integral in terms of $\boldsymbol{\rho}$, i.e.,
\begin{linenomath*}\begin{equation} \label{eq:dirac_rho}
\frac{1}{\sqrt{\left|W_{}^TR_{\mathbf{y}}^TB_{\mathbf{y}}^{}\right|^2}}\,\delta\left(\boldsymbol{\rho} + \left(W_{}^TR_{\mathbf{y}}^TB_{\mathbf{y}}^{}\right)^{-1}W_{}^T\left(R_{\mathbf{x}}^TA_{\mathbf{x}}^{}\boldsymbol{\omega} - \mathbf{c}\right)\right).
\end{equation}\end{linenomath*}
\revision{By using \eqref{eq:QxRx}, \ref{eq:C_ABUV} and \ref{eq:F_G} once again, we further obtain
\begin{linenomath*}\begin{align}
B_{\mathbf{y}}^{}B_{\mathbf{y}}^TR_{\mathbf{y}}^{}W&=(I-Q_{\mathbf{y}}^{}V_1^{}V_1^TQ_{\mathbf{y}}^T-R_{\mathbf{y}}^{}V_3^{}V_3^TR_{\mathbf{y}}^T)\,R_{\mathbf{y}}^{}W \nonumber \\
&=R_{\mathbf{y}}^{}W+Q_{\mathbf{y}}^{}V_1^{}V_1^TQ_{\mathbf{x}}^TR_{\mathbf{x}}^{}W-R_{\mathbf{y}}^{}V_3^{}V_3^T(I-R_{\mathbf{x}}^TR_{\mathbf{x}}^{})W=R_{\mathbf{y}}^{}W.
\end{align}\end{linenomath*}
Therefore,} substituting the expression in \eqref{eq:dirac_rho} into the integral in \eqref{eq:J_3} and using the sifting property in \eqref{eq:dirac_sift_multi} yields
\begin{linenomath*}\begin{equation} \label{eq:J_4}
\mathcal{J}=\frac{1}{\sqrt{\revision{\left|W_{}^TR_{\mathbf{y}}^TR_{\mathbf{y}}^{}W\right|}}}\,\delta(\boldsymbol{\theta} - V_4^T\mathbf{c})\int \mathcal{C}\left(\begin{bmatrix}
V_1^{} & V_2^{} & M
\end{bmatrix}\begin{bmatrix}
\boldsymbol{\alpha} \\
\boldsymbol{\beta} \\
\boldsymbol{\omega}
\end{bmatrix}+F\mathbf{c};\Lambda,\mathbf{h},g\right)\,\text{d}\boldsymbol{\alpha},
\end{equation}\end{linenomath*}
where we made use of the matrix definitions
\begin{linenomath*}\begin{equation}
F=Q_{\mathbf{y}}^TB_{\mathbf{y}}^{}(W^T_{}R_{\mathbf{y}}^TB_{\mathbf{y}}^{})^{-1}W^T_{}\revision{=\left(W(R_{\mathbf{y}}^{}W)^+Q_{\mathbf{y}}^{}\right)^T}
\end{equation}\end{linenomath*}
and
\begin{linenomath*}\begin{equation}
M=Q_{\mathbf{x}}^TA_{\mathbf{x}}^{}-FR_{\mathbf{x}}^TA_{\mathbf{x}}^{}=(Q_{\mathbf{x}}^T-FR_{\mathbf{x}}^T)\,A_{\mathbf{x}}^{}.
\end{equation}\end{linenomath*}
Next, the canonical factor inside the integral in \eqref{eq:J_4} can be rewritten to have scope $\boldsymbol{\alpha}$, $\boldsymbol{\beta}$ and $\boldsymbol{\omega}$ according to \eqref{eq:canonical_scope}, namely
\begin{linenomath*}\begin{equation}
\mathcal{C}\left(\begin{bmatrix}
\boldsymbol{\alpha} \\
\boldsymbol{\beta} \\
\boldsymbol{\omega}
\end{bmatrix};\begin{bmatrix}
V_1^T \\ V_2^T \\ M_{}^T
\end{bmatrix}\Lambda\begin{bmatrix}
V_1^T \\ V_2^T \\ M_{}^T
\end{bmatrix}^T,\begin{bmatrix}
V_1^T \\ V_2^T \\ M_{}^T
\end{bmatrix}(\mathbf{h}-\Lambda F\mathbf{c}),g+\left(\mathbf{h}-\frac{1}{2} \Lambda F\mathbf{c} \right)^T\,F\mathbf{c}\right).
\end{equation}\end{linenomath*}
Using the result for marginalisation of a canonical factor in \eqref{eq:canonical_marg}, the integral in \eqref{eq:J_4} becomes
\begin{linenomath*}\begin{equation} \label{eq:J_5}
\mathcal{J}=\frac{1}{\sqrt{\left|W_{}^TR_{\mathbf{y}}^TR_{\mathbf{y}}^{}W\right|}}\,\delta(\boldsymbol{\theta} - V_4^T\mathbf{c})\,\mathcal{C}\left(\begin{bmatrix}
\boldsymbol{\beta} \\
\boldsymbol{\omega}
\end{bmatrix};\hat{K},\hat{\mathbf{h}},\hat{g}\right),
\end{equation}\end{linenomath*}
where
\begin{linenomath*}\begin{align} \label{eq:Khg_hat}
\hat{K} &= \begin{bmatrix}
V_2 & M
\end{bmatrix}^T(\Lambda-\Lambda S\Lambda)\begin{bmatrix}
V_2 & M
\end{bmatrix} \nonumber \\
\hat{\mathbf{h}} &= \begin{bmatrix}
V_2 & M
\end{bmatrix}^T(I-\Lambda S)(\mathbf{h}-\Lambda F\mathbf{c}) \nonumber \\
\hat{g} &= g+\left(\mathbf{h}-\frac{1}{2} \Lambda F\mathbf{c} \right)^T\,F\mathbf{c}+\frac{1}{2}(\mathbf{h}-\Lambda F\mathbf{c})^TS(\mathbf{h}-\Lambda F\mathbf{c})+\frac{1}{2}\log\left|2\pi(V_1^T\Lambda V_1^{})^{-1}\right|
\end{align}\end{linenomath*}
and where we have made use of the matrix definition
\begin{linenomath*}\begin{equation}
S =  V_1^{}\left(V_1^T\Lambda V_1^{}\right)^{-1}V_1^T.
\end{equation}\end{linenomath*}
Reversing the change of variables according to \eqref{eq:xy_alpha} then yields
\begin{linenomath*}\begin{equation} \label{eq:degen_marg_result}
\mathcal{J}=\frac{1}{\sqrt{\left|W_{}^TR_{\mathbf{y}}^TR_{\mathbf{y}}^{}W\right|}}\,\mathcal{C}\left(\begin{bmatrix}
Q_{\mathbf{x}}V_2 & A_{\mathbf{x}}
\end{bmatrix}^T\mathbf{x};\hat{K},\hat{\mathbf{h}},\hat{g}\right)\,\delta(V_4^TR_{\mathbf{x}}^T\mathbf{x}-V_4^T\mathbf{c}).
\end{equation}\end{linenomath*}
\revision{Although the specific orthonormal bases $\begin{bmatrix}
Q_{\mathbf{x}}V_2 & A_{\mathbf{x}}
\end{bmatrix}$ and $R_{\mathbf{x}}V_4$ were convenient for deriving this result, recall from \eqref{eq:C_Qx} that the former is also a basis for $C(Q_{\mathbf{x}})$. Consequently, using an arbitrary basis $U$ such that $C(U)=C(Q_{\mathbf{x}})$ results in a simpler computation of the canonical factor in \eqref{eq:degen_marg_result} (without requiring the explicit calculation of $V_2$, $V_3$, $V_4$, $A_{\mathbf{x}}$ or $B_{\mathbf{y}}$). However, it is then also necessary to adapt $\hat{K}$ and $\hat{\mathbf{h}}$ to correspond to this new basis. Specifically, we require that
\begin{linenomath*}\begin{equation}
U\hat{K}U^T=\begin{bmatrix}
Q_{\mathbf{x}}V_2 & A_{\mathbf{x}}
\end{bmatrix}\begin{bmatrix}
V_2 & M
\end{bmatrix}^T(\Lambda-\Lambda S\Lambda)\begin{bmatrix}
V_2 & M
\end{bmatrix}\begin{bmatrix}
Q_{\mathbf{x}}V_2 & A_{\mathbf{x}}
\end{bmatrix}^T.
\end{equation}\end{linenomath*}
Expanding the first two factors and using \eqref{eq:QxRx}, \eqref{eq:C_ABUV} and \eqref{eq:F_G} yields
\begin{linenomath*}\begin{align}
\begin{bmatrix}
Q_{\mathbf{x}}V_2 & A_{\mathbf{x}}
\end{bmatrix}\begin{bmatrix}
V_2 & M
\end{bmatrix}^T&=Q_{\mathbf{x}}^{}V_2^{}V_2^T+A_{\mathbf{x}}^{}A_{\mathbf{x}}^T(Q_{\mathbf{x}}^{}-R_{\mathbf{x}}^{}F) \nonumber \\
&=Q_{\mathbf{x}}^{}V_2^{}V_2^T+(I-Q_{\mathbf{x}}^{}V_2^{}V_2^TQ_{\mathbf{x}}^T-R_{\mathbf{x}}^{}V_4^{}V_4^TR_{\mathbf{x}}^T)(Q_{\mathbf{x}}^{}-R_{\mathbf{x}}^{}F) \nonumber \\
&=Q_{\mathbf{x}}^{}(V_2^{}V_2^T(I-Q_{\mathbf{x}}^TQ_{\mathbf{x}}^{})+I)-(I-R_{\mathbf{x}}^{}V_4^{}V_4^TR_{\mathbf{x}}^T)R_{\mathbf{x}}^{}F \nonumber \\
&=Q_{\mathbf{x}}^{}(V_2^{}V_2^TQ_{\mathbf{y}}^TQ_{\mathbf{y}}^{}+I)-UU^T_{}R_{\mathbf{x}}^{}F \nonumber \\
&=UU^T_{}(Q_{\mathbf{x}}^{}-R_{\mathbf{x}}^{}F)
\end{align}\end{linenomath*}
and consequently the precision matrix
\begin{linenomath*}\begin{equation} \label{eq:lambda_prime}
\hat{K}=U^T_{}(Q_{\mathbf{x}}^{}-R_{\mathbf{x}}^{}F)(\Lambda-\Lambda S\Lambda)(Q_{\mathbf{x}}^{}-R_{\mathbf{x}}^{}F)^TU=Z\hat{\Lambda} Z^T
\end{equation}\end{linenomath*}
can be diagonalised using the SVD. Similarly,
\begin{linenomath*}\begin{equation}
U\hat{\mathbf{h}}=\begin{bmatrix}
Q_{\mathbf{x}}V_2 & A_{\mathbf{x}}
\end{bmatrix}\begin{bmatrix}
V_2 & M
\end{bmatrix}^T(I-\Lambda S)(\mathbf{h}-\Lambda F\mathbf{c})
\end{equation}\end{linenomath*}
and therefore
\begin{linenomath*}\begin{equation}
\hat{\mathbf{h}}=U^T_{}(Q_{\mathbf{x}}^{}-R_{\mathbf{x}}^{}F)(I-\Lambda S)(\mathbf{h}-\Lambda F\mathbf{c}).
\end{equation}\end{linenomath*}
Finally, for any $R'$ such that $C(R')=C(U)^\perp=C(R_{\mathbf{x}}V_4)$, by multiplying its argument with $R'^TR_{\mathbf{x}}V_4$ and using the result in \eqref{eq:rotated_Dirac_delta_3}, we can express the Dirac delta in \eqref{eq:degen_marg_result} as
\begin{linenomath*}\begin{equation}
\delta\left(V_4^TR_{\mathbf{x}}^T\mathbf{x}-V_4^T\mathbf{c}\right)=\delta\left(V_4^TR_{\mathbf{x}}^T\mathbf{x}-V_4^T(R_{\mathbf{x}}^TR_{\mathbf{x}}^{}+R_{\mathbf{y}}^TR_{\mathbf{y}}^{})\,\mathbf{c}\right)=\delta\left(R'^T\mathbf{x}-R'^TR_{\mathbf{x}}\mathbf{c}\right).
\end{equation}\end{linenomath*}
Substituting these new definitions into \eqref{eq:degen_marg_result} then yields
\begin{linenomath*}\begin{align}
\mathcal{J}&=\frac{1}{\sqrt{\left|W_{}^TR_{\mathbf{y}}^TR_{\mathbf{y}}^{}W\right|}}\,\mathcal{C}\left(U^T\mathbf{x};Z\hat{\Lambda} Z^T,\hat{\mathbf{h}},\hat{g}\right)\,\delta(R'^T\mathbf{x}-R'^TR_{\mathbf{x}}\mathbf{c}) \nonumber \\
&=\mathcal{D}\left(\mathbf{x};UZ,R',\hat{\Lambda},Z^T\hat{\mathbf{h}},R'^T_{}R_{\mathbf{x}}^{}\mathbf{c},\hat{g}-\frac{1}{2}\log\left|W_{}^TR_{\mathbf{y}}^TR_{\mathbf{y}}^{}W\right|\right).
\end{align}\end{linenomath*}}
Note that since both \revision{$U$} and $Z$ \revision{are semi-orthogonal, so is} their product (as required). This concludes the derivation of Algorithm \ref{alg:marginal}.

\subsection{Multiplication}

To determine the parameters of the resulting degenerate factor $\phi(\mathbf{x})$ as calculated according to Algorithm \ref{alg:multiply}, we need to compute the product
\begin{linenomath*}\begin{equation} \label{eq:product_degen}
\mathcal{P}=\mathcal{C}(Q_1^T\mathbf{x};\Lambda_1,\mathbf{h}_1,g_1)\,\delta(R_1^T\mathbf{x}-\mathbf{c}_1)\,\mathcal{C}(Q_2^T\mathbf{x};\Lambda_2,\mathbf{h}_2,g_2)\,\delta(R_2^T\mathbf{x}-\mathbf{c}_2).
\end{equation}\end{linenomath*}
We start by finding an orthonormal basis for the column space of $R^\prime$, which is
\begin{linenomath*}\begin{equation} \label{eq:C_Rp}
C(R^\prime)=C(R_1)+C(R_2).
\end{equation}\end{linenomath*}
In general, this basis is not unique. However, a convenient choice is to have the $n-k_1$ columns of $R_1$ as part of the basis. This will always be possible since the columns of $R_1$ are orthonormal per definition and $\text{rank}(R^\prime)\geq\text{rank}(R_1)$. The rest of the basis vectors of $C(R^\prime)$ can be obtained by projecting the columns of $R_2$ onto $N(R_1^T)=C(Q_1)$ and computing its column space. Since $Q_1$ has orthonormal columns, the necessary projection matrix is $Q_1^{}Q_1^T$. By defining the matrix $V$ (with orthonormal columns) such that
\begin{linenomath*}\begin{equation} \label{eq:def_V}
C(V)=C(Q_1^{}Q_1^TR_2^{}),
\end{equation}\end{linenomath*}
the columns of the matrix $\begin{bmatrix}
R_1 & V
\end{bmatrix}$ form an orthonormal basis for $C(R^\prime)$. As with the marginalisation operation, we now define a convenient orthogonal change of variables
\begin{linenomath*}\begin{equation} \label{eq:UVR}
\mathbf{x}=\begin{bmatrix}
U & V & R_1
\end{bmatrix}\begin{bmatrix}
\boldsymbol{\alpha} \\
\boldsymbol{\beta} \\
\boldsymbol{\gamma}
\end{bmatrix},
\end{equation}\end{linenomath*}
where
\begin{linenomath*}\begin{equation} \label{eq:def_U}
C(U)=(C(V)+C(R_1))^\perp.
\end{equation}\end{linenomath*}
Also note that $C(Q_1)=(C(U)+C(V))$. Returning to the product in \eqref{eq:product_degen} and using the substitution in \eqref{eq:UVR} yields
\begin{linenomath*}\begin{equation} \label{eq:product_alpha_beta}
\mathcal{P}=\mathcal{C}(Q_1^T(U\boldsymbol{\alpha}+V\boldsymbol{\beta});\Lambda_1,\mathbf{h}_1,g_1)\,\delta\left(\boldsymbol{\gamma}-\mathbf{c}_1\right) \\\mathcal{C}(Q_2^T(U\boldsymbol{\alpha}+V\boldsymbol{\beta}+R_1\boldsymbol{\gamma});\Lambda_2,\mathbf{h}_2,g_2)\,\delta\left(R_2^T(V^{}\boldsymbol{\beta} + R_1^{}\boldsymbol{\gamma})-\mathbf{c}_2\right).
\end{equation}\end{linenomath*}
Note that $R_2^TU=\textit{0}$ because, from \eqref{eq:C_Rp}, $C(R_2)\subseteq C(R')=C(U)^\perp$. Using the property in \eqref{eq:dirac_prod_multi} followed by the result in \eqref{eq:rotated_Dirac_delta_2}, we can express the product of the two Dirac delta functions in \eqref{eq:product_alpha_beta} as
\begin{linenomath*}\begin{align} \label{eq:delta_b}
\delta\left(\boldsymbol{\gamma}-\mathbf{c}_1\right)\,\delta\left(R_2^TV\boldsymbol{\beta} + R_2^TR_1^{}\boldsymbol{\gamma}-\mathbf{c}_2\right)&=\delta\left(\boldsymbol{\gamma}-\mathbf{c}_1\right)\,\delta\left(R_2^TV\boldsymbol{\beta} + R_2^TR_1^{}\mathbf{c}_1-\mathbf{c}_2\right) \nonumber \\
&=\frac{1}{\sqrt{\left|R_2^TV\right|^2}}\delta\left(\boldsymbol{\gamma}-\mathbf{c}_1\right)\,\delta\left(\boldsymbol{\beta}-\mathbf{b}\right),
\end{align}\end{linenomath*}
where
\begin{linenomath*}\begin{equation} \label{eq:def_b}
\mathbf{b}=\left(R_2^TV\right)^{-1}(\mathbf{c}_2-R_2^TR_1^{}\mathbf{c}_1).
\end{equation}\end{linenomath*}
Substituting \eqref{eq:delta_b} into the product in \eqref{eq:product_alpha_beta} and using the property in \eqref{eq:dirac_prod_multi} once again, we can write
\begin{linenomath*}\begin{equation}
\mathcal{P}=\frac{1}{\sqrt{\left|R_2^TV\right|^2}}\mathcal{C}(Q_1^T(U\boldsymbol{\alpha}+V\mathbf{b});\Lambda_1,\mathbf{h}_1,g_1)\,\mathcal{C}(Q_2^T(U\boldsymbol{\alpha}+V\mathbf{b}+R_1\mathbf{c}_1);\Lambda_2,\mathbf{h}_2,g_2)\,\delta\left(\begin{bmatrix}
\boldsymbol{\gamma} \\
\boldsymbol{\beta}
\end{bmatrix}-\begin{bmatrix}
\mathbf{c}_1 \\
\mathbf{b}
\end{bmatrix}\right).
\end{equation}\end{linenomath*}
Each of the two canonical factors can then be rewritten to have scope $\boldsymbol{\alpha}$ according to \eqref{eq:canonical_scope} and can be combined using the result in \eqref{eq:canonical_multiply}, yielding
\begin{linenomath*}\begin{equation} \label{eq:P_5}
\mathcal{P}=\frac{1}{\sqrt{\left|R_2^TV\right|^2}}\,\mathcal{C}(\boldsymbol{\alpha};\hat{K},\hat{\mathbf{h}},\hat{g})\,\delta\left(\begin{bmatrix}
\boldsymbol{\gamma} \\
\boldsymbol{\beta}
\end{bmatrix}-\begin{bmatrix}
\mathbf{c}_1 \\
\mathbf{b}
\end{bmatrix}\right),
\end{equation}\end{linenomath*}
where we have made use of the definitions
\begin{linenomath*}\begin{align} \label{eq:Khg_hat_2}
\hat{K}&=U^T_{} (Q_1^{} \Lambda_1^{} Q_1^T+Q_2^{} \Lambda_2^{} Q_2^T)\, U \nonumber \\
\hat{\mathbf{h}}&=U^T_{}(Q_1^{}(\mathbf{h}_1^{}-\Lambda_1^{} Q_1^TV\mathbf{b})+Q_2^{}(\mathbf{h}_2^{}-\Lambda_2^{} Q_2^T(V\mathbf{b}+R_1\mathbf{c}_1))) \nonumber \\
\hat{g}&=g_1^{}+g_2^{}+\left(\mathbf{h}_1-\frac{1}{2}\Lambda_1^{}Q_1^TV\mathbf{b}\right)^TQ_1^TV\mathbf{b}+\left(\mathbf{h}_2-\frac{1}{2}\Lambda_2^{}Q_2^T(V\mathbf{b}+R_1\mathbf{c}_1)\right)^TQ_2^T (V\mathbf{b}+R_1\mathbf{c}_1).
\end{align}\end{linenomath*}
The precision matrix in \eqref{eq:Khg_hat_2} can be diagonalised using the SVD
\begin{linenomath*}\begin{equation} \label{eq:lambda_prime_2}
\hat{K}=Z\hat{\Lambda} Z^T.
\end{equation}\end{linenomath*}
Reversing the change of variables in \eqref{eq:UVR} and substituting \eqref{eq:lambda_prime_2} into \eqref{eq:P_5} yields
\begin{linenomath*}\begin{align} \label{eq:degen_prod_result}
\mathcal{P}&=\frac{1}{\sqrt{\left|R_2^TV\right|^2}}\mathcal{C}(U^T\mathbf{x};Z\hat{\Lambda} Z^T,\hat{\mathbf{h}},\hat{g})\,\delta\left(\begin{bmatrix}
R_1 & V
\end{bmatrix}^T\mathbf{x}-\begin{bmatrix}
\mathbf{c}_1 \\
\mathbf{b}
\end{bmatrix}\right) \nonumber \\
&=\mathcal{C}\left(Z^TU^T\mathbf{x};\hat{\Lambda},Z^T\hat{\mathbf{h}},\hat{g}-\frac{1}{2}\log\left|R_2^TV\right|^2\right)\,\delta\left(\begin{bmatrix}
R_1 & V
\end{bmatrix}^T\mathbf{x}-\begin{bmatrix}
\mathbf{c}_1 \\
\mathbf{b}
\end{bmatrix}\right) \nonumber \\
&=\mathcal{D}\left(\mathbf{x};UZ,\begin{bmatrix}
R_1 & V
\end{bmatrix},\hat{\Lambda},Z^T\hat{\mathbf{h}},\begin{bmatrix}
\mathbf{c}_1 \\
\mathbf{b}
\end{bmatrix},\hat{g}-\frac{1}{2}\log\left|R_2^TV\right|^2\right).
\end{align}\end{linenomath*}
Note that since both $U$ and $Z$ have orthonormal columns, so does their product (as required). This concludes the derivation of Algorithm \ref{alg:multiply}.

\subsection{Division}

To determine the parameters of the resulting degenerate factor $\phi(\mathbf{x})$ as calculate according to Algorithm \ref{alg:divide}, we need to compute the quotient
\begin{linenomath*}\begin{equation} \label{eq:quotient_degen}
\mathcal{Q}=\frac{\mathcal{C}(Q_1^T\mathbf{x};\Lambda_1,\mathbf{h}_1,g_1)\,\delta(R_1^T\mathbf{x}-\mathbf{c}_1)}{\mathcal{C}(Q_2^T\mathbf{x};\Lambda_2,\mathbf{h}_2,g_2)\,\delta(R_2^T\mathbf{x}-\mathbf{c}_2)}=\mathcal{D}(\mathbf{x};Q',R',\Lambda',\mathbf{h}',\mathbf{c}',g').
\end{equation}\end{linenomath*}
Since multiplication of Dirac delta functions is better defined than division, a more precise formulation is to calculate the quantities $Q'$, $R'$, $\Lambda'$, $\mathbf{h}'$, $\mathbf{c}'$ and $g'$ such that
\begin{linenomath*}\begin{align} \label{eq:quotient_degen_prod}
\mathcal{P}&=\mathcal{C}(Q'^T\mathbf{x};\Lambda',\mathbf{h}',g')\,\delta(R'^T\mathbf{x}-\mathbf{c}')\,\mathcal{C}(Q_2^T\mathbf{x};\Lambda_2,\mathbf{h}_2,g_2)\,\delta(R_2^T\mathbf{x}-\mathbf{c}_2) \nonumber \\
&=\mathcal{C}(Q_1^T\mathbf{x};\Lambda_1,\mathbf{h}_1,g_1)\,\delta(R_1^T\mathbf{x}-\mathbf{c}_1).
\end{align}\end{linenomath*}
This is similar to the product in \eqref{eq:product_degen}, where the $x'$-superscripts and $x_1$-subscripts have been reversed. Therefore, just as in \eqref{eq:C_Rp}, \eqref{eq:quotient_degen_prod} requires that
\begin{linenomath*}\begin{equation} \label{eq:C_Rp_divide}
C(R_1)=C(R')+C(R_2).
\end{equation}\end{linenomath*}
This satisfies the assumption in Section \ref{sec:division} that $C(R_2)\subseteq C(R_1)$ and consequently $Q_1^TR_2^{}=\textit{0}$. To prove that the proposed parameters
\begin{linenomath*}\begin{align}
R^\prime_{} &= \texttt{Complement}\left(\begin{bmatrix}
Q_1 & R_2
\end{bmatrix}\right) \nonumber \\
\mathbf{c}^\prime &= R^{\prime T}R_1\mathbf{c}_1 \nonumber \\
Q'&=\begin{bmatrix}
Q_1Z & R_2
\end{bmatrix} \nonumber \\
\Lambda^\prime&=\begin{bmatrix}
\Lambda_+ & \textit{0} \\
\textit{0} & \textit{0}
\end{bmatrix} \nonumber \\
\mathbf{h}^\prime_{}&= Q'^T_{}(Q_1^{}\mathbf{h}_1^{}-Q_2^{}(\mathbf{h}_2^{}-\Lambda_2^{}Q_2^TR_1^{}\mathbf{c}_1^{})) \nonumber \\
g^\prime_{}&= g_1^{}-g_2^{}-\left(\mathbf{h}_2-\frac{1}{2}\Lambda_2^{}Q_2^TR_1^{}\mathbf{c}_1^{}\right)^TQ_2^T R_1^{}\mathbf{c}_1^{}
\end{align}\end{linenomath*}
(where $Z\Lambda_+Z^T=\Lambda_1^{}-Q_1^TQ_2^{}\Lambda_2^{}Q_2^TQ_1^{}$) satisfy \eqref{eq:quotient_degen_prod}, we use the results in Algorithm \ref{alg:multiply}, with the $x_1$-subscripts replaced by $x'$-superscripts and the latter in turn replaced by $x''$-superscripts. Since $V=R_2$ and $U=Q_1$, this yields
\begin{linenomath*}\begin{equation} \label{eq:result_division_proof}
\mathcal{P}=\mathcal{C}(Q''^T\mathbf{x};\Lambda'',\mathbf{h}'',g'')\,\delta(R''^T\mathbf{x}-\mathbf{c}''),
\end{equation}\end{linenomath*}
where
\begin{linenomath*}\begin{align}
R''&=\begin{bmatrix}
R' & R_2
\end{bmatrix} \nonumber \\
\mathbf{c}'' &= \begin{bmatrix}
\mathbf{c}' \\
\mathbf{c}_2-R_2^TR'\mathbf{c}'
\end{bmatrix} = \begin{bmatrix}
R' & R_2
\end{bmatrix}^TR_1\mathbf{c}_1 \nonumber \\
\Lambda''&=Q_1^T\left(\begin{bmatrix}
Q_1Z & R_2
\end{bmatrix}\begin{bmatrix}
\Lambda_+ & \textit{0} \\
\textit{0} & \textit{0}
\end{bmatrix}\begin{bmatrix}
Q_1Z & R_2
\end{bmatrix}^T+Q_2^{}\Lambda_2^{}Q_2^T\right)\,Q_1 \nonumber \\
&=Q_1^T\left(Q_1^{}\left(\Lambda_1^{}-Q_1^TQ_2^{}\Lambda_2^{}Q_2^TQ_1^{}\right)Q_1^T+Q_2^{}\Lambda_2^{}Q_2^T\right)\,Q_1 \nonumber \\
&=\Lambda_1^{} \nonumber \\
Q''&= Q_1 \nonumber \\
\mathbf{h}''&= Q_1^T\left(\begin{bmatrix}
Q_1Z & R_2
\end{bmatrix}\left(\begin{bmatrix}
Q_1Z & R_2
\end{bmatrix}^T_{}(Q_1^{}\mathbf{h}_1^{}-Q_2^{}(\mathbf{h}_2^{}-\Lambda_2^{}Q_2^TR_1^{}\mathbf{c}_1^{}))-{}\right.\right. \nonumber \\
&\quad \left.\begin{bmatrix}
\Lambda_+ & \textit{0} \\
\textit{0} & \textit{0}
\end{bmatrix}\begin{bmatrix}
Q_1Z & R_2
\end{bmatrix}^TR_2^{}\mathbf{c}_2^{}\right)+Q_2^{}(\mathbf{h}_2^{}-\Lambda_2^{}Q_2^TR''\mathbf{c}'')\Bigg) \nonumber \\
&=\mathbf{h}_1 \nonumber \\
g''&= g_1^{}-g_2^{}-\left(\mathbf{h}_2-\frac{1}{2}\Lambda_2^{}Q_2^TR_1^{}\mathbf{c}_1^{}\right)^TQ_2^T R_1^{}\mathbf{c}_1^{}+g_2^{}+{} \nonumber \\
&\quad\left(\mathbf{h}'-\frac{1}{2}\Lambda'Q'^TR_2^{}\mathbf{c}_2^{}\right)^TQ'^T R_2^{}\mathbf{c}_2^{}+\left(\mathbf{h}_2-\frac{1}{2}\Lambda_2^{}Q_2^TR''\mathbf{c}''\right)^TQ_2^T R''\mathbf{c}'' \nonumber \\
&=g_1^{}+\begin{bmatrix}
Z^T_{}Q_1^T(Q_1^{}\mathbf{h}_1^{}-Q_2^{}(\mathbf{h}_2^{}-\Lambda_2^{}Q_2^TR_1^{}\mathbf{c}_1^{})) \\
\mathbf{0}
\end{bmatrix}^T\begin{bmatrix}
\mathbf{0} \\
\mathbf{c}_2
\end{bmatrix} \nonumber \\
&=g_1^{}.
\end{align}\end{linenomath*}
This reveals that the product of the canonical factors in \eqref{eq:quotient_degen_prod} is correct. Since $C\left(R_1\right)=C\left(\begin{bmatrix}
R' & R_2
\end{bmatrix}\right)=C\left(Q_1\right)^\perp$, by multiplying its argument with $R_1^T\begin{bmatrix}
R' & R_2
\end{bmatrix}$ and using the result in \eqref{eq:rotated_Dirac_delta_3}, we can further express the Dirac delta in \eqref{eq:result_division_proof} as
\begin{linenomath*}\begin{equation}
\delta\left(\begin{bmatrix}
R' & R_2
\end{bmatrix}^T\mathbf{x}-\begin{bmatrix}
R' & R_2
\end{bmatrix}^TR_1\mathbf{c}_1\right)=\delta\left(R_1^T\mathbf{x}-\mathbf{c}_1^{}\right).
\end{equation}\end{linenomath*}
This concludes the derivation of Algorithm \ref{alg:divide}.

\subsection{Reduction}

To determine the parameters of the resulting degenerate factor $\phi(\mathbf{x})$ as calculated according to Algorithm \ref{alg:reduce}, we need to substitute the evidence $\mathbf{y}_0$ into \eqref{eq:jointDegen} to obtain $\phi(\mathbf{x},\mathbf{y}_0)$ and then express this factor over scope $\mathbf{x}$. To this end, we start by defining an auxiliary factor with scope $\{\mathbf{x},\mathbf{y}\}$, namely
\begin{linenomath*}\begin{equation} \label{eq:phi_0}
\phi_0(\mathbf{x},\mathbf{y})=\phi(\mathbf{x},\mathbf{y})\,\delta\left(\mathbf{y}-\mathbf{y}_0\right).
\end{equation}\end{linenomath*}
This multiplication of the joint factor with a Dirac delta can be interpreted as imposing an additional linear constraint due to the evidence $\mathbf{y}=\mathbf{y}_0$. If we further multiply by a carefully-chosen vacuous canonical factor (with $K=\textit{0}$, $\mathbf{h}=\mathbf{0}$ and $g=0$) and rearrange the factors, \eqref{eq:phi_0} becomes
\begin{linenomath*}\begin{align} \label{eq:evid_as_prod}
\phi_0(\mathbf{x},\mathbf{y})&=\mathcal{C}\left(\mathbf{x};\textit{0},\mathbf{0},0\right)\,\delta\left(\mathbf{y}-\mathbf{y}_0\right)\,\phi(\mathbf{x},\mathbf{y}) \nonumber \\
&=\mathcal{C}\left(\begin{bmatrix}
I & \textit{0}
\end{bmatrix}\begin{bmatrix}
\mathbf{x} \\
\mathbf{y}
\end{bmatrix};\textit{0},\mathbf{0},0\right)\,\delta\left(\begin{bmatrix}
\textit{0} & I
\end{bmatrix}\begin{bmatrix}
\mathbf{x} \\
\mathbf{y}
\end{bmatrix}-\mathbf{y}_0\right)\mathcal{C}\left(\begin{bmatrix}
Q_{\mathbf{x}}^T & Q_{\mathbf{y}}^T
\end{bmatrix}\begin{bmatrix}
\mathbf{x} \\
\mathbf{y}
\end{bmatrix};\Lambda,\mathbf{h},g\right)\,\delta\left(\begin{bmatrix}
R_{\mathbf{x}}^T & R_{\mathbf{y}}^T
\end{bmatrix}\begin{bmatrix}
\mathbf{x} \\
\mathbf{y}
\end{bmatrix}-\mathbf{c}\right).
\end{align}\end{linenomath*}
The motivation for this expansion becomes apparent when comparing \eqref{eq:product_degen} and \eqref{eq:evid_as_prod}. In particular, note that the evidence operation for degenerate Gaussian factors becomes a special case of the multiplication operation, where $Q_1=\begin{bmatrix}
I & \textit{0}
\end{bmatrix}^T$, $R_1=\begin{bmatrix}
\textit{0} & I
\end{bmatrix}^T$ and $\mathbf{c}_1=\mathbf{y}_0$. We can therefore use the results from \eqref{eq:degen_prod_result} to determine the product in \eqref{eq:evid_as_prod}. Returning to the definition of the matrix $V$ in \eqref{eq:def_V}, however, we see that in this case
\begin{linenomath*}\begin{equation} \label{eq:new_V}
C(V)=C\left(\begin{bmatrix}
I \\
\textit{0}
\end{bmatrix}\begin{bmatrix}
I & \textit{0}
\end{bmatrix}\begin{bmatrix}
R_{\mathbf{x}} \\
R_{\mathbf{y}}
\end{bmatrix}\right)=C\left(\begin{bmatrix}
R_{\mathbf{x}} \\
\textit{0}
\end{bmatrix}\right).
\end{equation}\end{linenomath*}
This means that $V_{\mathbf{y}}=\textit{0}$. Therefore, the expression for the vector $\mathbf{b}$ in \eqref{eq:def_b} can be simplified to
\begin{linenomath*}\begin{equation} \label{eq:new_b}
\mathbf{b}=(R_{\mathbf{x}}^TV_{\mathbf{x}}^{})^{-1}(\mathbf{c}-R_{\mathbf{y}}^T\mathbf{y}_0^{}).
\end{equation}\end{linenomath*}
Furthermore, since $C(U)\perp C\left(\begin{bmatrix}
\textit{0} & I
\end{bmatrix}^T\right)$, $U_{\mathbf{y}}=\textit{0}$. This means that
\begin{linenomath*}\begin{equation} \label{eq:new_U}
UZ=\begin{bmatrix}
U_{\mathbf{x}} \\
\textit{0}
\end{bmatrix}Z=\begin{bmatrix}
U_{\mathbf{x}}Z \\
\textit{0}
\end{bmatrix}\qquad\text{and}\qquad U^TQ=\begin{bmatrix}
U_{\mathbf{x}} \\
\textit{0}
\end{bmatrix}^T\begin{bmatrix}
Q_{\mathbf{x}} \\
Q_{\mathbf{y}}
\end{bmatrix}=U_{\mathbf{x}}^TQ_{\mathbf{x}}^{}.
\end{equation}\end{linenomath*}
Using \eqref{eq:degen_prod_result} (where $\Lambda_1=\textit{0}$, $\mathbf{h}_1=\mathbf{0}$ and $g_1=0$) and substituting \eqref{eq:new_V} to \eqref{eq:new_U}, \eqref{eq:evid_as_prod} becomes
\begin{linenomath*}\begin{equation} \label{eq:phi_0_xy}
\phi_0(\mathbf{x},\mathbf{y})=\mathcal{D}\left(\begin{bmatrix}
\mathbf{x} \\
\mathbf{y}
\end{bmatrix};\begin{bmatrix}
U_{\mathbf{x}}Z \\
\textit{0}
\end{bmatrix},\begin{bmatrix}
\textit{0} & V_{\mathbf{x}} \\
I & \textit{0}
\end{bmatrix},\hat{\Lambda},\hat{\mathbf{h}},\begin{bmatrix}
\mathbf{y}_0 \\
\mathbf{b}
\end{bmatrix},\hat{g}\right),
\end{equation}\end{linenomath*}
where
\begin{linenomath*}\begin{align}
Z\hat{\Lambda} Z^T&=\texttt{SVD}(U_{\mathbf{x}}^TQ_{\mathbf{x}}^{}\Lambda Q_{\mathbf{x}}^TU_{\mathbf{x}}^{}) \nonumber \\
\hat{\mathbf{h}}&=Z_{}^TU_{\mathbf{x}}^TQ_{\mathbf{x}}^{}\left(\mathbf{h}-\Lambda Q_{}^T\begin{bmatrix}
V_{\mathbf{x}}\mathbf{b} \\
\mathbf{y}_0
\end{bmatrix}\right) \nonumber \\
\hat{g}&=g+\left(\mathbf{h}-\frac{1}{2}\Lambda Q_{}^T\begin{bmatrix}
V_{\mathbf{x}}\mathbf{b} \\
\mathbf{y}_0
\end{bmatrix}\right)^T\,Q_{}^T\begin{bmatrix}
V_{\mathbf{x}}\mathbf{b} \\
\mathbf{y}_0
\end{bmatrix}-\frac{1}{2}\log\left|R_{\mathbf{x}}^TV_{\mathbf{x}}^{}\right|^2.
\end{align}\end{linenomath*}
Using the definition of the degenerate factor in \eqref{eq:degen_def} and that of the multidimensional Dirac delta function in \eqref{eq:ND_Diracdelta}, \eqref{eq:phi_0_xy} becomes
\begin{linenomath*}\begin{align} \label{eq:phi_0_result}
\phi_0(\mathbf{x},\mathbf{y})&=\mathcal{C}\left(\begin{bmatrix}
Z_{}^TU_{\mathbf{x}}^T & \textit{0}
\end{bmatrix}\begin{bmatrix}
\mathbf{x} \\
\mathbf{y}
\end{bmatrix};\hat{\Lambda},\hat{\mathbf{h}},\hat{g}\right)\,\delta\left(\begin{bmatrix}
\textit{0} & I \\
V_{\mathbf{x}}^T & \textit{0}
\end{bmatrix}\begin{bmatrix}
\mathbf{x} \\
\mathbf{y}
\end{bmatrix}-\begin{bmatrix}
\mathbf{y}_0 \\
\mathbf{b}
\end{bmatrix}\right) \nonumber \\
&=\mathcal{C}\left(Z_{}^TU_{\mathbf{x}}^T\mathbf{x};\hat{\Lambda},\hat{\mathbf{h}},\hat{g}\right)\,\delta\left(V_{\mathbf{x}}^T\mathbf{x}-\mathbf{b}\right)\,\delta\left(\mathbf{y}-\mathbf{y}_0\right).
\end{align}\end{linenomath*}
Substituting $\mathbf{y}=\mathbf{y}_0$ into \eqref{eq:phi_0_result} yields
\begin{linenomath*}\begin{equation} \label{eq:phi_0_y_0_1}
\phi_0(\mathbf{x},\mathbf{y}_0)=\mathcal{C}\left(Z_{}^TU_{\mathbf{x}}^T\mathbf{x};\hat{\Lambda},\hat{\mathbf{h}},\hat{g}\right)\,\delta\left(V_{\mathbf{x}}^T\mathbf{x}-\mathbf{b}\right)\,\delta\left(\mathbf{y}_0-\mathbf{y}_0\right).
\end{equation}\end{linenomath*}
However, using \eqref{eq:phi_0}, we also obtain
\begin{linenomath*}\begin{align} \label{eq:phi_0_y_0_2}
\phi_0(\mathbf{x},\mathbf{y}_0)=\phi(\mathbf{x},\mathbf{y}_0)\,\delta(\mathbf{y}_0-\mathbf{y}_0).
\end{align}\end{linenomath*}
By comparing \eqref{eq:phi_0_y_0_1} and \eqref{eq:phi_0_y_0_2}, we can conclude that
\begin{linenomath*}\begin{equation} \label{eq:reduce_result}
\phi(\mathbf{x},\mathbf{y}_0)=\mathcal{C}\left(Z_{}^TU_{\mathbf{x}}^T\mathbf{x};\hat{\Lambda},\hat{\mathbf{h}},\hat{g}\right)\,\delta\left(V_{\mathbf{x}}^T\mathbf{x}-\mathbf{b}\right)=\mathcal{D}(\mathbf{x};U_{\mathbf{x}}^{}Z,V_{\mathbf{x}}^{},\hat{\Lambda},\hat{\mathbf{h}},\mathbf{b},\hat{g}).
\end{equation}\end{linenomath*}
This concludes the derivation of Algorithm \ref{alg:reduce}.

\subsection{Conditional density functions}

To determine the parameters of the degenerate factor representing the conditional density $p(\mathbf{y}|\mathbf{x})$ as calculated according to Algorithm \ref{alg:conditional_factor}, we rewrite \eqref{eq:affine_noise} as
\begin{linenomath*}\begin{equation}
\mathbf{y}=I\mathbf{w}+(A\mathbf{x}+\mathbf{b}).
\end{equation}\end{linenomath*}
Using the results in Algorithm \ref{alg:affine_transform} for the special case of the identity matrix, we can then express the conditional density as
\begin{linenomath*}\begin{equation} \label{eq:p_ygx_hat}
p(\mathbf{y}|\mathbf{x})=\mathcal{D}(\mathbf{y};Q,R,\Lambda,\hat{\mathbf{h}},\hat{\mathbf{c}},\hat{g}),
\end{equation}\end{linenomath*}
where
\begin{linenomath*}\begin{align}
\hat{\mathbf{c}}&=R^T\left((Q\Lambda^{-1}\mathbf{h}+R\mathbf{c})+(A\mathbf{x}+\mathbf{b})\right)=\mathbf{c}+R^T(A\mathbf{x}+\mathbf{b}) \nonumber \\
\hat{\mathbf{h}}&=\Lambda Q^T\left((Q\Lambda^{-1}\mathbf{h}+R\mathbf{c})+(A\mathbf{x}+\mathbf{b})\right)=\mathbf{h}+\Lambda Q^T(A\mathbf{x}+\mathbf{b}) \nonumber \\
\hat{g}&=-\frac{1}{2}(\mathbf{h}+\Lambda Q^T(A\mathbf{x}+\mathbf{b}))^T\Lambda^{-1}(\mathbf{h}+\Lambda Q^T(A\mathbf{x}+\mathbf{b}))-\frac{1}{2}\log\left|2\pi\Lambda^{-1}\right| \nonumber \\
&=g-\frac{1}{2} (A\mathbf{x}+\mathbf{b})^TQ\Lambda Q^T(A\mathbf{x}+\mathbf{b})-\mathbf{h}^TQ^T(A\mathbf{x}+\mathbf{b})
\end{align}\end{linenomath*}
and where we have made use of the result in \eqref{eq:g_norm} in the last line. We can then use the definition of the degenerate factor in \eqref{eq:degen_def} to expand \eqref{eq:p_ygx_hat} according to
\begin{linenomath*}\begin{equation} \label{eq:p_ygx_expand}
p(\mathbf{y}|\mathbf{x})=\mathcal{C}\left(Q^T\mathbf{y};\Lambda,\mathbf{h}+\Lambda Q^T(A\mathbf{x}+\mathbf{b}),\hat{g}\right)\,\delta\left(R^T\mathbf{y}-\left(\mathbf{c}+R^T\left(A\mathbf{x}+\mathbf{b}\right)\right)\right).
\end{equation}\end{linenomath*}
By using the definition in \eqref{eq:canon_def} and rearranging terms, the canonical factor in \eqref{eq:p_ygx_expand} can further be rewritten as
\begin{linenomath*}\begin{align} \label{eq:p_ygx_rewrite_1}
&\quad\exp\left(-\frac{1}{2}\mathbf{y}^TQ\Lambda Q^T\mathbf{y}+(\mathbf{h}+\Lambda Q^T(A\mathbf{x}+\mathbf{b}))^TQ^T\mathbf{y}+\hat{g}\right) \nonumber \\
&=\exp\left(-\frac{1}{2}\begin{bmatrix}
\mathbf{x} \\
\mathbf{y}
\end{bmatrix}
^T\begin{bmatrix}
-A & I
\end{bmatrix}^TQ\Lambda Q^T\begin{bmatrix}
-A & I
\end{bmatrix}\begin{bmatrix}
\mathbf{x} \\
\mathbf{y}
\end{bmatrix}+{}\right. \nonumber \\
&\quad\left.(\mathbf{h}+\Lambda Q^T\mathbf{b})^TQ^T\begin{bmatrix}
-A & I
\end{bmatrix}\begin{bmatrix}
\mathbf{x} \\
\mathbf{y}
\end{bmatrix}+ g-(\mathbf{h}+\frac{1}{2}\Lambda Q^T\mathbf{b})^T \,Q^T\mathbf{b}\right).
\end{align}\end{linenomath*}
Similarly, the Dirac delta in \eqref{eq:p_ygx_expand} can be rewritten as
\begin{linenomath*}\begin{equation} \label{eq:p_ygx_rewrite_2}
\delta\left(R^T\left(\mathbf{y}-A\mathbf{x}\right)-\left(\mathbf{c}+R^T\mathbf{b}\right)\right)=\delta\left(R^T\begin{bmatrix}
-A & I
\end{bmatrix}\begin{bmatrix}
\mathbf{x} \\
\mathbf{y}
\end{bmatrix}-(\mathbf{c}+R^T\mathbf{b})\right).
\end{equation}\end{linenomath*}
To guarantee that the basis $R'$ has orthonormal columns, we define the basis $U$ such that
\begin{linenomath*}\begin{equation}
C(U)=C\left(\begin{bmatrix}
-A & I
\end{bmatrix}^TR\right).
\end{equation}\end{linenomath*}
Since the two corresponding projection matrices are also equal, we multiply the argument of the Dirac delta in \eqref{eq:p_ygx_rewrite_2} by the carefully-chosen matrix
\begin{linenomath*}\begin{equation}
Z=U^T\begin{bmatrix}
-A & I
\end{bmatrix}^TR\left(R^T\left(I+AA^T\right)R\right)^{-1}
\end{equation}\end{linenomath*}
and use \eqref{eq:rotated_Dirac_delta_2} to write
\begin{linenomath*}\begin{equation} \label{eq:p_ygx_dirac_Z}
\delta\left(R^T\begin{bmatrix}
-A & I
\end{bmatrix}\begin{bmatrix}
\mathbf{x} \\
\mathbf{y}
\end{bmatrix}-(\mathbf{c}+R^T\mathbf{b})\right)=\sqrt{|Z|^2}\,\delta\left(R'^T\begin{bmatrix}
\mathbf{x} \\
\mathbf{y}
\end{bmatrix}-\mathbf{c}'\right),
\end{equation}\end{linenomath*}
where
\begin{linenomath*}\begin{equation}
R'=U\qquad\text{and}\qquad\mathbf{c}'=Z(\mathbf{c}+R^T\mathbf{b}).
\end{equation}\end{linenomath*}
To find the remainder of the parameters $Q'$, $\Lambda'$, $\mathbf{h}'$ and $g'$, the quadratic term in the exponent in \eqref{eq:p_ygx_rewrite_1} should be expanded according to
\begin{linenomath*}\begin{equation} \label{eq:p_ygx_quadratic}
-\frac{1}{2}\begin{bmatrix}
\mathbf{x} \\
\mathbf{y}
\end{bmatrix}
^TS\begin{bmatrix}
\mathbf{x} \\
\mathbf{y}
\end{bmatrix}=-\frac{1}{2}\begin{bmatrix}
\mathbf{x} \\
\mathbf{y}
\end{bmatrix}
^T(I-R'R'^T)S(I-R'R'^T)\begin{bmatrix}
\mathbf{x} \\
\mathbf{y}
\end{bmatrix}+\frac{1}{2}\begin{bmatrix}
\mathbf{x} \\
\mathbf{y}
\end{bmatrix}
^TR'R'^TSR'R'^T\begin{bmatrix}
\mathbf{x} \\
\mathbf{y}
\end{bmatrix}-\begin{bmatrix}
\mathbf{x} \\
\mathbf{y}
\end{bmatrix}
^TR'R'^TS\begin{bmatrix}
\mathbf{x} \\
\mathbf{y}
\end{bmatrix},
\end{equation}\end{linenomath*}
where we made use of the matrix definitions
\begin{linenomath*}\begin{equation}
S=F^TQ\Lambda Q^TF\qquad\text{and}\qquad F=\begin{bmatrix}
-A & I
\end{bmatrix}.
\end{equation}\end{linenomath*}
Substituting \eqref{eq:p_ygx_rewrite_1}, \eqref{eq:p_ygx_rewrite_2}, \eqref{eq:p_ygx_dirac_Z} and \eqref{eq:p_ygx_quadratic} into \eqref{eq:p_ygx_expand} and using the multiplicative property in \eqref{eq:dirac_prod_multi} yields
\begin{linenomath*}\begin{align} \label{eq:p_ygx_2}
p(\mathbf{y}|\mathbf{x})&=\exp\left(-\frac{1}{2}\begin{bmatrix}
\mathbf{x} \\
\mathbf{y}
\end{bmatrix}
^T(I-R'R'^T)S(I-R'R'^T)\begin{bmatrix}
\mathbf{x} \\
\mathbf{y}
\end{bmatrix}+{}\right. \nonumber \\
&\quad\qquad\left. \frac{1}{2}\mathbf{c}'^TR'^TSR'\mathbf{c}'-\mathbf{c}'^TR'^TS\begin{bmatrix}
\mathbf{x} \\
\mathbf{y}
\end{bmatrix}+(\mathbf{h}+\Lambda Q^T\mathbf{b})^TQ^TF\begin{bmatrix}
\mathbf{x} \\
\mathbf{y}
\end{bmatrix}+{}\right. \nonumber \\
&\quad\qquad\left. g-(\mathbf{h}+\frac{1}{2}\Lambda Q^T\mathbf{b})^T \,Q^T\mathbf{b}+\frac{1}{2}\log|Z|^2\right)\,\delta\left(R'^T\begin{bmatrix}
\mathbf{x} \\
\mathbf{y}
\end{bmatrix}-\mathbf{c}'\right).
\end{align}\end{linenomath*}
This reveals the motivation behind the expansion in \eqref{eq:p_ygx_quadratic}, namely that, for the compact SVD of the quadratic coefficient in \eqref{eq:p_ygx_2}
\begin{linenomath*}\begin{equation} \label{eq:Q+}
Q_+^{}\Lambda_+^{}Q_+^T=(I-R'R'^T)S(I-R'R'^T),
\end{equation}\end{linenomath*}
we ensure that $C(Q_+)\perp C(R')$. However, if the density function over the vector $\mathbf{w}\in\mathbb{R}^{n_{\mathbf{y}}}$ in \eqref{eq:p_w} has $k$ degrees of degeneracy, then
\begin{linenomath*}\begin{equation}
\text{rank}(R')=\text{rank}(R)=k\qquad\text{and}\qquad\text{rank}(Q_+)=\text{rank}(Q)=n_{\mathbf{y}}-k
\end{equation}\end{linenomath*}
and consequently $C(Q_+)+C(R')\subset\mathbb{R}^{n_{\mathbf{x}}+n_{\mathbf{y}}}$, where $\mathbf{x}\in\mathbb{R}^{n_{\mathbf{x}}}$. We therefore define a third orthonormal basis $Q_\infty$ (with dimension $n_{\mathbf{x}}$) to complete the decomposition such that
\begin{linenomath*}\begin{equation}
C(Q_\infty)=(C(Q_+)+C(R'))^\perp.
\end{equation}\end{linenomath*}
For such an orthogonal decomposition, we can relate the respective projection matrices according to
\begin{linenomath*}\begin{equation} \label{eq:I-RR}
I-R'R'^T=Q_+^{}Q_+^T+Q_\infty^{}Q_\infty^T=\begin{bmatrix}
Q_+ & Q_\infty
\end{bmatrix}\begin{bmatrix}
Q_+ & Q_\infty
\end{bmatrix}^T.
\end{equation}\end{linenomath*}
Similar to \eqref{eq:p_ygx_quadratic}, we now expand the linear term in the exponent in \eqref{eq:p_ygx_2} according to
\begin{linenomath*}\begin{equation} \label{eq:p_ygx_linear}
\mathbf{w}^T\begin{bmatrix}
\mathbf{x} \\
\mathbf{y}
\end{bmatrix}=\mathbf{w}^T(I-R'R'^T)\begin{bmatrix}
\mathbf{x} \\
\mathbf{y}
\end{bmatrix}+\mathbf{w}^TR'R'^T\begin{bmatrix}
\mathbf{x} \\
\mathbf{y}
\end{bmatrix},
\end{equation}\end{linenomath*}
where we have made use of the definition
\begin{linenomath*}\begin{equation}
\mathbf{w}=F^TQ(\mathbf{h}+\Lambda Q^T\mathbf{b})-SR'\mathbf{c}'=F^TQ(\mathbf{h}+\Lambda Q^T(\mathbf{b}-FR'\mathbf{c}')).
\end{equation}\end{linenomath*}
Finally, substituting \eqref{eq:Q+} and \eqref{eq:p_ygx_linear} followed by \eqref{eq:I-RR} into \eqref{eq:p_ygx_2} and once again using the multiplicative property in \eqref{eq:dirac_prod_multi} yields
\begin{linenomath*}\begin{equation} \label{eq:conditional_result}
p(\mathbf{y}|\mathbf{x})=\mathcal{D}\left(\begin{bmatrix}
\mathbf{x} \\
\mathbf{y}
\end{bmatrix};\begin{bmatrix}
Q_+ & Q_\infty
\end{bmatrix},R',\begin{bmatrix}
\Lambda_+ & \textit{0} \\
\textit{0} & \textit{0}
\end{bmatrix},\begin{bmatrix}
Q_+ & Q_\infty
\end{bmatrix}^T\mathbf{w},\mathbf{c}',g'\right),
\end{equation}\end{linenomath*}
where
\begin{linenomath*}\begin{equation}
g'=g-(\mathbf{h}+\frac{1}{2}\Lambda Q^T\mathbf{b})^T \,Q^T\mathbf{b}+(\mathbf{h}+\Lambda Q^T(\mathbf{b}-\frac{1}{2}FR'\mathbf{c}'))^TQ^TFR'\mathbf{c}'+\frac{1}{2}\log|Z|^2
\end{equation}\end{linenomath*}
since
\begin{linenomath*}\begin{align}
\frac{1}{2}SR'\mathbf{c}'+\mathbf{w}&=\frac{1}{2}F^TQ\Lambda Q^TFR'\mathbf{c}'+F^TQ(\mathbf{h}+\Lambda Q^T(\mathbf{b}-FR'\mathbf{c}')) \nonumber \\
&=F^TQ\left(\mathbf{h}+\Lambda Q^T\left(\mathbf{b}-\frac{1}{2}FR'\mathbf{c}'\right)\right).
\end{align}\end{linenomath*}
This concludes the derivation of Algorithm \ref{alg:conditional_factor}.

\subsection{Equivalent affine transformations}

To determine the parameters of the equivalent affine transformation in \eqref{eq:equiv_linear} as well as that of the degenerate factor representing the transform noise in \eqref{eq:noise_linear} as calculated according to Algorithm \ref{alg:equiv_transform}, we start by approximating the moments of the joint density $p(\mathbf{x},\mathbf{y})$ using the unscented transform. However, since the prior distribution in \eqref{eq:p_xw} is degenerate and the Cholesky decomposition for a positive semi-definite matrix is not defined, the $2n+1$ sigma points proposed by Thrun et al. \cite{thrun2005probabilistic} cannot be used directly. Instead, we propose adapting their result to draw $2(n-k)+1$ sigma points
\begin{linenomath*}\begin{equation} \label{eq:eps_sigma}
\mathcal{E}=\begin{bmatrix}
\mathbf{0}, & \gamma\sqrt{\Lambda^{-1}}, & {}-\gamma\sqrt{\Lambda^{-1}}
\end{bmatrix}+\Lambda^{-1}\mathbf{h}\mathbf{1}^T
\end{equation}\end{linenomath*}
from only the canonical component of the degenerate factor in \eqref{eq:degen_def}, where $k$ is the degree of degeneracy and where we have made use of the substitution
\begin{linenomath*}\begin{equation} \label{eq:eps_Q}
Q^T\begin{bmatrix}
\mathbf{x} \\
\mathbf{w}
\end{bmatrix}=\boldsymbol{\epsilon}.
\end{equation}\end{linenomath*}
Since all samples (with non-zero likelihood) from the prior distribution in \eqref{eq:p_xw} must satisfy
\begin{linenomath*}\begin{equation} \label{eq:c_R}
R^T\begin{bmatrix}
\mathbf{x} \\
\mathbf{w}
\end{bmatrix}=\mathbf{c},
\end{equation}\end{linenomath*}
the $2k$ sigma points corresponding to the Dirac delta component of the degenerate factor and the basis $R$ are not necessary. By combining \eqref{eq:eps_Q} and \eqref{eq:c_R}, and since $C(Q)=C(R)^\perp$, we can then write
\begin{linenomath*}\begin{equation}
QQ^T\begin{bmatrix}
\mathbf{x} \\
\mathbf{w}
\end{bmatrix}+RR^T\begin{bmatrix}
\mathbf{x} \\
\mathbf{w}
\end{bmatrix}=\begin{bmatrix}
\mathbf{x} \\
\mathbf{w}
\end{bmatrix}=Q\boldsymbol{\epsilon}+R\mathbf{c}.
\end{equation}\end{linenomath*}
This can in turn be used to transform the sigma points in \eqref{eq:eps_sigma} to correspond to the prior $p(\mathbf{x},\mathbf{w})$, i.e.,
\begin{linenomath*}\begin{equation}
\begin{bmatrix}
\mathcal{X} \\
\mathcal{W}
\end{bmatrix}=\begin{bmatrix}
\mathbf{0}, & \gamma Q\sqrt{\Lambda^{-1}}, & {}-\gamma Q\sqrt{\Lambda^{-1}}
\end{bmatrix}+(Q\Lambda^{-1}\mathbf{h}+R\mathbf{c})\mathbf{1}^T.
\end{equation}\end{linenomath*}
Each of the sigma point pairs $\mathbf{x}^{[i]}$ and $\mathbf{w}^{[i]}$ are then propagatded through the nonlinear transformation in \eqref{eq:non_linear} to yield $\mathbf{y}^{[i]}=\mathbf{f}\left(\mathbf{x}^{[i]},\mathbf{w}^{[i]}\right)$. As outlined by Thrun et al. \cite{thrun2005probabilistic}, the moments of the joint density $p(\mathbf{x},\mathbf{y})$ can be approximated as
\begin{linenomath*}\begin{align} \label{eq:unscented_degen}
\mathbb{E}\left[\begin{bmatrix}
\mathbf{x} \\
\mathbf{y}
\end{bmatrix}\right]&\approx\sum_{i=0}^{2(n-k)}w_m^{[i]}\begin{bmatrix}
\mathbf{x}^{[i]} \\
\mathbf{y}^{[i]}
\end{bmatrix} \nonumber \\
\text{Cov}\left[\begin{bmatrix}
\mathbf{x} \\
\mathbf{y}
\end{bmatrix}\right]&\approx\sum_{i=0}^{2(n-k)}w_c^{[i]}\,\left(\begin{bmatrix}
\mathbf{x}^{[i]} \\
\mathbf{y}^{[i]}
\end{bmatrix}-\boldsymbol{\mu}'\right)\left(\begin{bmatrix}
\mathbf{x}^{[i]} \\
\mathbf{y}^{[i]}
\end{bmatrix}-\boldsymbol{\mu}'\right)^T.
\end{align}\end{linenomath*}
For the equivalent affine transformation in \eqref{eq:equiv_linear}, the moments involving $\mathbf{y}$ are given by
\begin{linenomath*}\begin{align} \label{eq:three_moments}
\text{Cov}[\mathbf{x},\mathbf{y}]&=\text{Cov}\left[\mathbf{x}\right]\widetilde{A}^T \nonumber \\
\text{Cov}[\mathbf{y}]&=\widetilde{A}\,\text{Cov}\left[\mathbf{x}\right]\widetilde{A}^T+\text{Cov}\left[\widetilde{\mathbf{w}}\right] \nonumber \\
\mathbb{E}[\mathbf{y}]&=\widetilde{A}\,\mathbb{E}[\mathbf{x}]+\widetilde{\mathbf{b}}+\mathbb{E}[\widetilde{\mathbf{w}}],
\end{align}\end{linenomath*}
where $\text{Cov}\left[\mathbf{x}\right]$ and $\mathbb{E}[\mathbf{x}]$ are already specified according to \eqref{eq:p_xw}. Therefore, given the moments in \eqref{eq:unscented_degen}, the three equations in \eqref{eq:three_moments} can be used to determine $\widetilde{A}$, $\text{Cov}[\widetilde{\mathbf{w}}]$ and $\widetilde{\mathbf{b}}$, respectively. Specifically, if we let the marginal density $p(\mathbf{x})=\mathcal{D}\left(\mathbf{x};Q_{\mathbf{x}},R_{\mathbf{x}},\Lambda_{\mathbf{x}},\mathbf{h}_{\mathbf{x}},\mathbf{c}_{\mathbf{x}},g_{\mathbf{x}}\right)$, we can use the first part of \eqref{eq:three_moments} to write
\begin{linenomath*}\begin{equation} \label{eq:cov_xy}
\text{Cov}[\mathbf{x},\mathbf{y}]=Q_{\mathbf{x}}^{}\Lambda_{\mathbf{x}}^{-1}Q_{\mathbf{x}}^T\widetilde{A}^T\quad\Longrightarrow\quad Q_{\mathbf{x}}^T\widetilde{A}^T=\Lambda_{\mathbf{x}}^{}Q_{\mathbf{x}}^T\text{Cov}[\mathbf{x},\mathbf{y}].
\end{equation}\end{linenomath*}
Now defining the decomposition $\widetilde{A}^T=\widetilde{A}_Q^T+\widetilde{A}_R^T$ such that $C\left(\widetilde{A}_Q^T\right)\subseteq C\left(Q_{\mathbf{x}}\right)$ and $C\left(\widetilde{A}_R^T\right)\subseteq C\left(R_{\mathbf{x}}\right)$, multiplying on both sides of \eqref{eq:cov_xy} by $Q_{\mathbf{x}}$ yields
\begin{linenomath*}\begin{equation}
Q_{\mathbf{x}}^{}Q_{\mathbf{x}}^T\left(\widetilde{A}_Q^T+\widetilde{A}_R^T\right)=Q_{\mathbf{x}}^{}\Lambda_{\mathbf{x}}^{}Q_{\mathbf{x}}^T\text{Cov}[\mathbf{x},\mathbf{y}]\quad\Longrightarrow\quad\widetilde{A}_Q=\text{Cov}\left[\mathbf{x},\mathbf{y}\right]^T\text{Cov}\left[\mathbf{x}\right]^+,
\end{equation}\end{linenomath*}
where we have made use of the pseudo-inverse of the SVD for short. With this value of $\widetilde{A}_Q$ known, the second part of \eqref{eq:three_moments} can be used to write
\begin{linenomath*}\begin{align} \label{eq:cov_w}
\text{Cov}\left[\widetilde{\mathbf{w}}\right]&=\text{Cov}\left[\mathbf{y}\right]-\left(\widetilde{A}_Q^T+\widetilde{A}_R^T\right)\,Q_{\mathbf{x}}^{}\Lambda_{\mathbf{x}}^{-1}Q_{\mathbf{x}}^T\left(\widetilde{A}_Q^T+\widetilde{A}_R^T\right)^T \nonumber \\
&=\text{Cov}\left[\mathbf{y}\right]-\widetilde{A}_Q^{}\,\text{Cov}\left[\mathbf{x}\right]\widetilde{A}_Q^T,
\end{align}\end{linenomath*}
which clearly does not depend on $\widetilde{A}_R$. Finally, the third part of \eqref{eq:three_moments} can be used to write
\begin{linenomath*}\begin{equation}
\widetilde{\mathbf{b}}+\widetilde{A}_R\,\mathbb{E}[\mathbf{x}]+\mathbb{E}[\widetilde{\mathbf{w}}]=\mathbb{E}[\mathbf{y}]-\widetilde{A}_Q\,\mathbb{E}[\mathbf{x}].
\end{equation}\end{linenomath*}
Since this is an underdetermined system of equations, we are free to choose $\widetilde{A}_R=\textit{0}$ and $\mathbb{E}[\widetilde{\mathbf{w}}]=\mathbf{0}$, without loss of generality. The former implies that $\widetilde{A}=\widetilde{A}_Q$ and the latter that $\widetilde{\mathbf{h}}=\mathbf{0}$ and $\widetilde{\mathbf{c}}=\mathbf{0}$. The remainder of the parameters $\widetilde{Q}$, $\widetilde{R}$, $\widetilde{\Lambda}$ and $\widetilde{g}$ can then be obtained by computing the compact SVD of \eqref{eq:cov_w} and calculating the normalisation constant according to \eqref{eq:g_norm}. This concludes the derivation of Algorithm \ref{alg:equiv_transform}.

\subsection{Kullback-Leibler divergence}

To calculate the KL divergence in \eqref{eq:KL} for the two degenerate densities in \eqref{eq:KL_densities}, we need to calculate the expectation
\begin{linenomath*}\begin{equation} \label{eq:KL_1}
D_\text{KL}(P||Q)=\mathbb{E}_{p(\mathbf{x})}\left[\log \frac{p(\mathbf{x})}{q(\mathbf{x})}\right].
\end{equation}\end{linenomath*}
According to Algorithm \ref{alg:divide}, for the quotient of two degenerate densities
\begin{linenomath*}\begin{equation}
\frac{p(\mathbf{x})}{q(\mathbf{x})}=\mathcal{D}(\mathbf{x};Q',R',\Lambda',\mathbf{h}',\mathbf{c}',g')
\end{equation}\end{linenomath*}
where $C\left(R_1\right)=C\left(R_2\right)$, $C(R')=\{\mathbf{0}\}$. The expectation in \eqref{eq:KL_1} (with respect to the density $P$) therefore becomes
\begin{linenomath*}\begin{equation} \label{eq:KL_2}
D_\text{KL}(P||Q)=\mathbb{E}\left[\log \mathcal{C}(Q'^T\mathbf{x};\Lambda',\mathbf{h}',g')\right]=-\frac{1}{2}\mathbb{E}\left[\mathbf{x}^TQ'\Lambda'Q'^T\mathbf{x}\right]+\mathbb{E}\left[\mathbf{x}\right]^TQ'\mathbf{h}'+g'.
\end{equation}\end{linenomath*}
(Since the Dirac delta components of the two densities are equivalent, we are effectively only comparing their canonical components.) The next step is to calculate the three terms in \eqref{eq:KL_2}. By using the results in Algorithm \ref{alg:divide}, and since the matrix $Z$ is orthogonal and $C\left(Q_1\right)=C\left(Q_2\right)$, we can write
\begin{linenomath*}\begin{equation} \label{eq:QlamQ}
Q'\Lambda'Q'^T=Q_1^{}(\Lambda_1^{}-Q_1^TQ_2^{}\Lambda_2^{}Q_2^TQ_1^{})\,Q_1^T\quad\text{and}\quad Q'\mathbf{h}'=Q_1^{}\mathbf{h}_1^{}-Q_2^{}\mathbf{h}_2^{}.
\end{equation}\end{linenomath*}
Since
\begin{linenomath*}\begin{equation} \label{eq:ExTAx}
\mathbb{E}\left[\mathbf{x}^TA\mathbf{x}\right]=\mathbb{E}\left[\mathbf{x}\right]^TA\,\mathbb{E}\left[\mathbf{x}\right]+\text{tr}(A\,\text{Cov}[\mathbf{x}]),
\end{equation}\end{linenomath*}
we use the first part of \eqref{eq:QlamQ} and the results in \eqref{eq:mean_degen} and \eqref{eq:cov_degen} to calculate
\begin{linenomath*}\begin{align} \label{eq:sub2}
\mathbb{E}\left[\mathbf{x}\right]^TQ'\Lambda'Q'^T\mathbb{E}\left[\mathbf{x}\right]&=\mathbf{h}_1^T\Lambda_1^{-1}(\Lambda_1^{}-Q_1^TQ_2^{}\Lambda_2^{}Q_2^TQ_1^{})\,\Lambda_1^{-1}\mathbf{h}_1^{}
\end{align}\end{linenomath*}
and
\begin{linenomath*}\begin{align} \label{eq:sub1}
\text{tr}(Q'\Lambda'Q'^T\text{Cov}[\mathbf{x}])&=\text{tr}\left( Q_1^{}(\Lambda_1^{}-Q_1^TQ_2^{}\Lambda_2^{}Q_2^TQ_1^{})\,Q_1^TQ_1^{}\Lambda_1^{-1}Q_1^T\right) \nonumber \\
&=\text{tr}\left(Q_1^{}Q_1^T-Q_1^{}Q_1^TQ_2^{}\Lambda_2^{}Q_2^TQ_1^{}\Lambda_1^{-1}Q_1^T\right) \nonumber \\
&=\text{tr}\left(Q_1^{}Q_1^T\right)-\text{tr}\left(Q_2^{}\Lambda_2^{}Q_2^TQ_1^{}\Lambda_1^{-1}Q_1^TQ_1^{}Q_1^T\right) \nonumber \\
&=n-k-\text{tr}\left(Q_2^{}\Lambda_2^{}Q_2^TQ_1^{}\Lambda_1^{-1}Q_1^T\right).
\end{align}\end{linenomath*}
In the second-to-last line of \eqref{eq:sub1} we used the fact that $\text{tr}(AB)=\text{tr}(BA)$ and in the last line that the trace of an orthogonal projection matrix is equal to its rank. Next, by using the second part of \eqref{eq:QlamQ}, we can write
\begin{linenomath*}\begin{equation} \label{eq:sub3}
\mathbb{E}\left[\mathbf{x}\right]^TQ'\mathbf{h}'=\left(Q_1^{}\Lambda_1^{-1}\mathbf{h}_1^{}+R_1^{}\mathbf{c}_1^{}\right)^T\left(Q_1^{}\mathbf{h}_1^{}-Q_2^{}\mathbf{h}_2^{}\right)=\mathbf{h}_1^T\Lambda_1^{-1}(\mathbf{h}_1^{}-Q_1^TQ_2^{}\mathbf{h}_2^{}).
\end{equation}\end{linenomath*}
Finally, since $g'=g_1-g_2$, we can substitute \eqref{eq:sub2}, \eqref{eq:sub1} and \eqref{eq:sub3} into \eqref{eq:KL_2} to yield
\begin{linenomath*}\begin{align}
D_\text{KL}(P||Q)&=\frac{1}{2}\Big(\text{tr}\left(Q_2^{}\Lambda_2^{}Q_2^TQ_1^{}\Lambda_1^{-1}Q_1^T\right)+\mathbf{h}_1^T\Lambda_1^{-1}Q_1^TQ_2^{}\Lambda_2^{}Q_2^TQ_1^{}\Lambda_1^{-1}\mathbf{h}_1^{}+{} \nonumber \\
&\quad\qquad\mathbf{h}_1^T\Lambda_1^{-1}\mathbf{h}_1^{}-n+k\Big)-\mathbf{h}_1^T\Lambda_1^{-1}Q_1^TQ_2^{}\mathbf{h}_2^{}+g_1^{}-g_2^{}.
\end{align}\end{linenomath*}
This concludes the derivation of the KL divergence in \eqref{eq:KL_result}.

\section*{References}
\markright{}
\addcontentsline{toc}{section}{References}

\bibliography{references}

\end{document}